\documentclass[acmtog,authorversion]{acmart}
\usepackage{booktabs} 
\setcitestyle{square}

\usepackage[ruled]{algorithm2e} 
\usepackage{amsmath}
\usepackage{amssymb}
\usepackage{mathrsfs}
\usepackage{color}
\usepackage{subfigure}
\usepackage{array}
\usepackage{float}
\usepackage[british,UKenglish,USenglish,english,american]{babel}
\usepackage{csquotes}

\newcommand{\Tref}[1]{Table~\ref{#1}}
\newcommand{\Eref}[1]{Equation~(\ref{#1})}

\newcommand{\Sref}[1]{Section~\ref{#1}}

\newcommand{\fref}[1]{Fig.~\ref{#1}}

\newcommand{\ie}{\emph{i.e.}}
\newcommand{\eg}{\emph{e.g.}}
\definecolor{mycol}{RGB}{0, 51, 153}
\newcommand{\hmm}[1]{\textcolor{black}{#1}}

\SetAlFnt{\small}
\SetAlCapFnt{\small}
\SetAlCapNameFnt{\small}
\SetAlCapHSkip{0pt}
\IncMargin{-\parindent}

\setcopyright{none}

\hypersetup{draft} 

\begin{document}
	\title{Progressive Color Transfer with Dense Semantic Correspondences}
	\author{Mingming He}\authornote{This work was done when Mingming He and Dongdong Chen were interns at MSR Asia.}
	\affiliation{%
		\institution{Hong Kong UST}
		\city{Hong Kong}
	}
	\author{Jing Liao}\authornote{indicates corresponding author.}
	\affiliation{%
		\institution{City University of Hong Kong, Microsoft Research}
		\city{Hong Kong}
	}
	\author{Dongdong Chen}
	\affiliation{%
		\institution{University of Science and Technology of China}
		\city{Hefei}
	}
	\author{Lu Yuan}
	\affiliation{%
		\institution{Microsoft AI Perception and Mixed Reality}
		\city{Seattle}
	}
	\author{Pedro V. Sander} 
	\affiliation{%
		\institution{Hong Kong UST}
		\city{Hong Kong}
	}
	
	\renewcommand\shortauthors{He, M. et al.}
	
	\begin{abstract}
		We propose a new algorithm for color transfer between images that have perceptually similar semantic structures. We aim to achieve a more accurate color transfer that leverages semantically-meaningful dense correspondence between images. To accomplish this, our algorithm uses neural representations for matching. Additionally, the color transfer should be spatially variant and globally coherent. Therefore, our algorithm optimizes a local linear model for color transfer satisfying both local and global constraints. Our proposed approach jointly optimizes matching and color transfer, adopting a coarse-to-fine strategy. The proposed method can be successfully extended from \emph{\enquote{one-to-one}} to \emph{\enquote{one-to-many}} color transfer. The latter further addresses the problem of mismatching elements of the input image. We validate our proposed method by testing it on a large variety of image content.
	\end{abstract}
	
	%
	%

	\ccsdesc[500]{Computing methodologies~Image manipulation}
	\ccsdesc[300]{Computing methodologies~Computational photography}
	
	%
	%
	
	\keywords{color, transfer, deep matching}

	
	
	\begin{teaserfigure}
		\centering
		\footnotesize
		\setlength{\tabcolsep}{0.003\linewidth}
		{
			\begin{tabular}{ccccc}
				\includegraphics[height=0.135\linewidth]{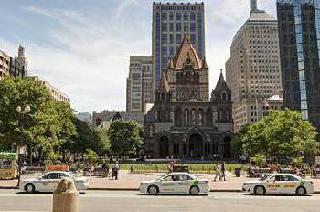}&
				\includegraphics[height=0.135\linewidth]{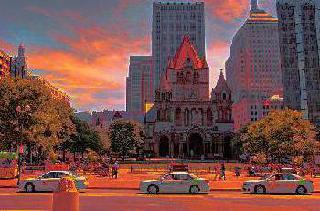}&
				\includegraphics[height=0.135\linewidth]{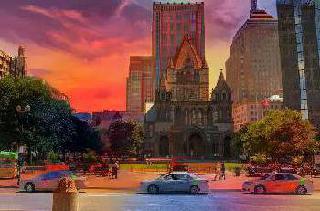}& \hspace{0.02in}
				\includegraphics[height=0.135\linewidth]{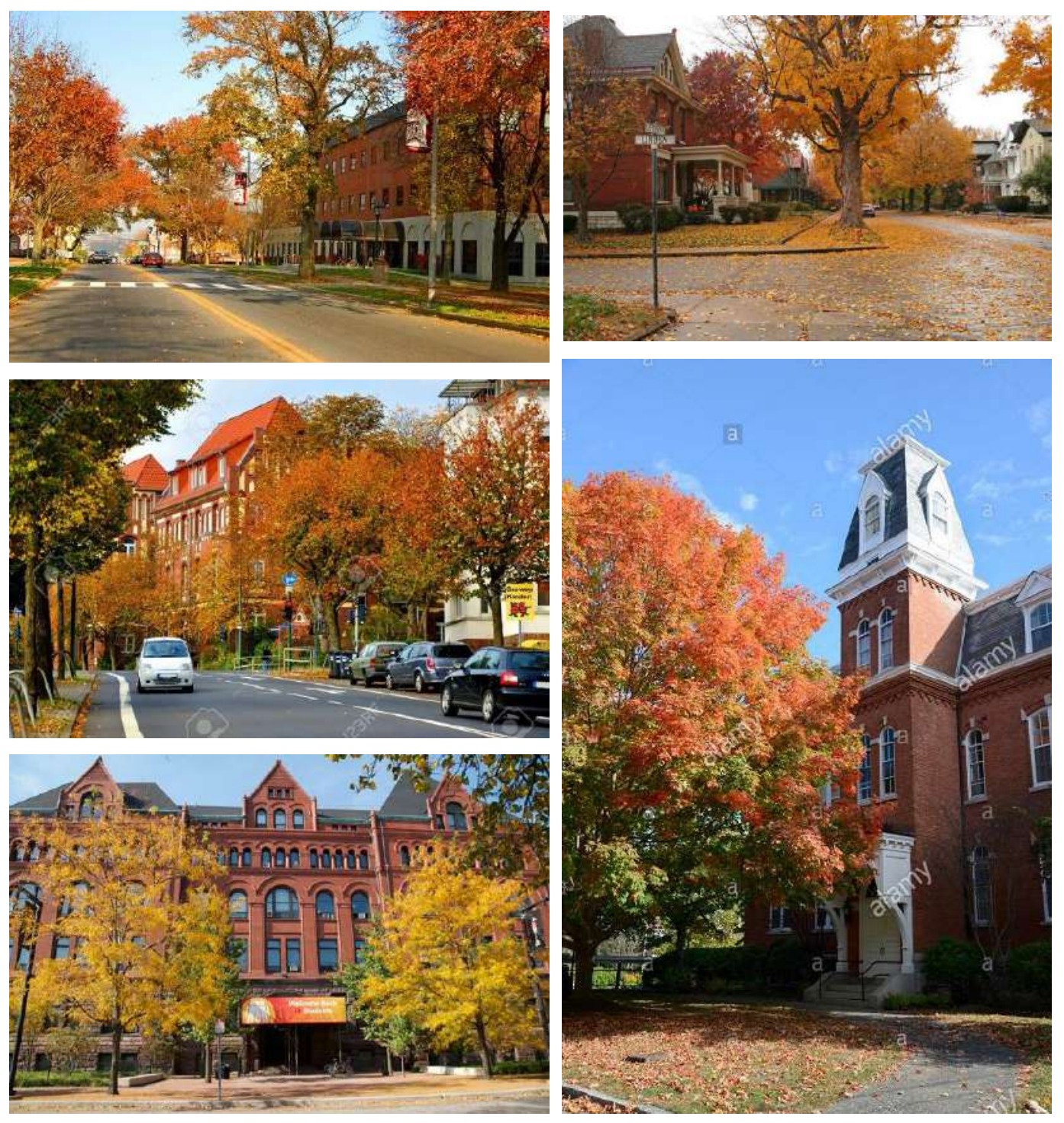}&
				\includegraphics[height=0.135\linewidth]{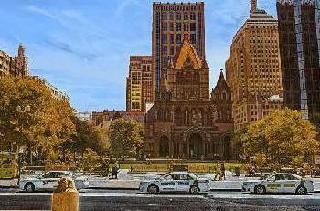}\\
				Source & \citet{pitie2005n} & \citet{luan2017deep} & \hspace{0.02in}Refs (\enquote{street autumn}) & Our result (\enquote{street autumn})\\
				\includegraphics[height=0.135\linewidth]{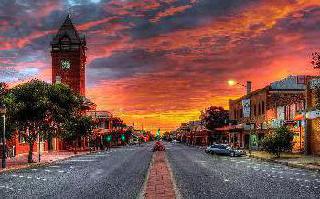}&
				\includegraphics[height=0.135\linewidth]{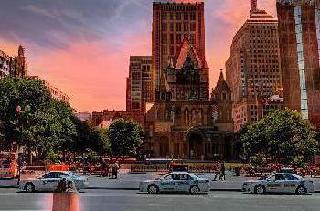}&
				\includegraphics[height=0.135\linewidth]{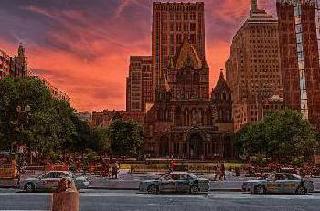}&
				\hspace{0.02in} \includegraphics[height=0.135\linewidth]{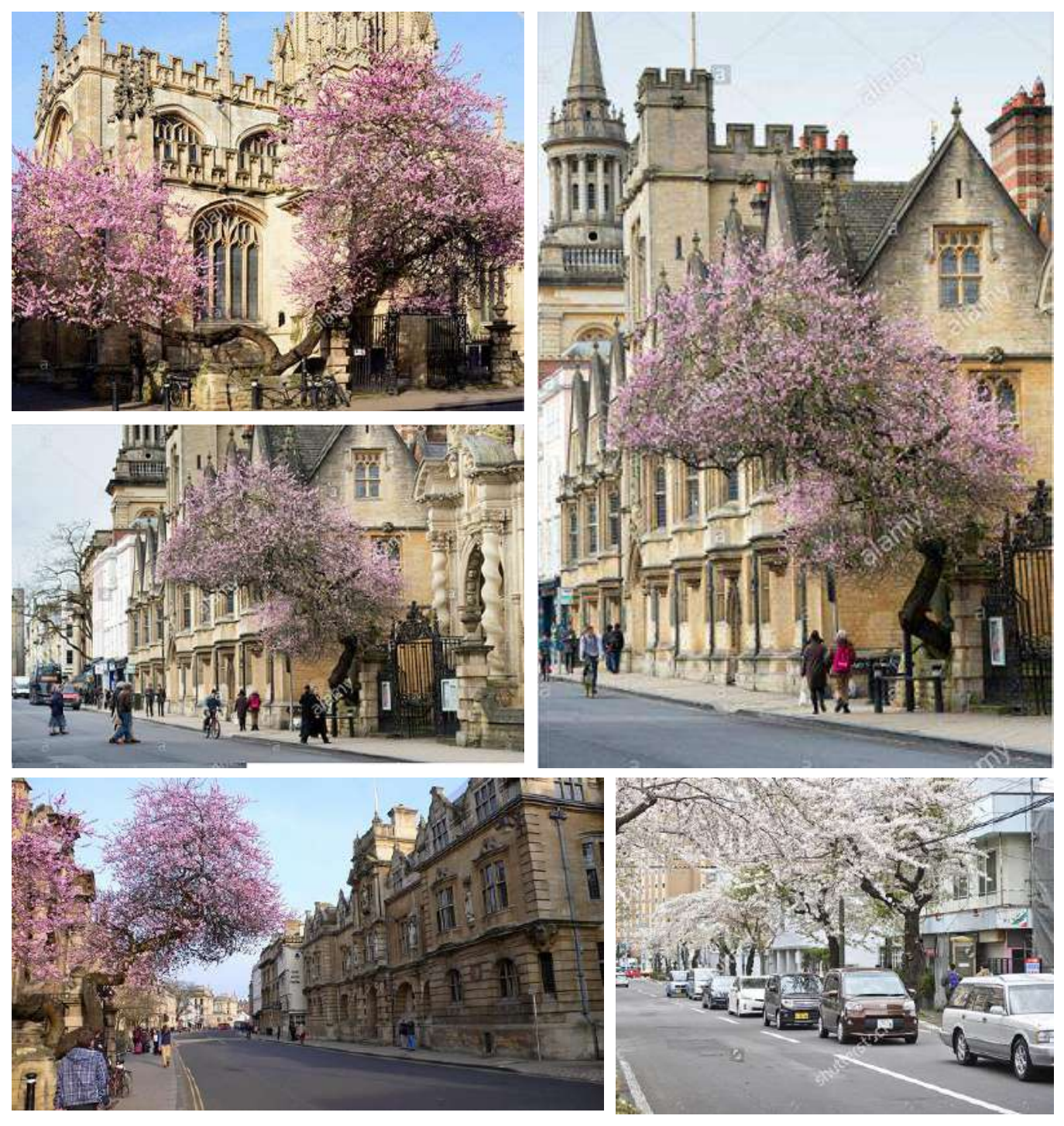}&
				\includegraphics[height=0.135\linewidth]{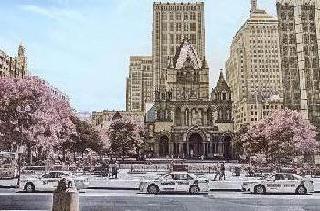}\\
				Reference& \citet{liao2017image} & Our result & \hspace{0.02in}Refs (\enquote{street sakura}) & Our result (\enquote{street sakura})
			\end{tabular}
		}
		\caption{Our method leverages semantically-meaningful dense correspondences between images, thus achieving a more accurate object-to-object color transfer than other methods (left). Moreover, our method can be successfully extended to multiple references (right). Input images: Bill Damon (Source) and PicsWalls.com (Reference). }
		\label{fig:teaser}
	\end{teaserfigure}
	
	\maketitle
	\section{Introduction}
	\hmm{Color transfer is a long-standing problem that seeks to transfer the color style of a reference image onto a source image. By using different references, one can alter the color style without changing the original image content in order to emulate different illumination, weather conditions, scene materials, or even artistic color effects.}
	
	\hmm{To achieve more accurate transfer, semantically meaningful correspondences are necessary to be established between input images. Due to large variations in appearance, matching methods based on hand-crafted features (\eg, intensity, Gabor wavelet, SIFT, or SSIM) may fail. Therefore, some methods require additional segmentation~\citep{dale2009image, luan2017deep}, or user specifications~\citep{an2008appprop} but these regional correspondences are not quite effective in their pixel-level accuracy. Recently,~\citet{liao2017image} leverage multi-level features of the deep neural network for dense correspondence and then conduct local color transfer during post-processing. This method is robust in finding high-level semantic correspondences between different objects, but may misalign some fine-scale image structures because low-level neural representations are still influenced by the color discrepancies. Thus, in their color transfer results, artifacts such as ghosting and a halo may appear, \eg, the halo around the pillar in~\fref{fig:teaser}.}
	
	\hmm{To refine correspondences and reduce these color transfer artifacts, we propose a novel progressive framework, which allows for the dense semantic correspondences estimation in the deep feature domain and local color transfer in the image domain to mutually contribute each other. This is implemented progressively by leveraging multi-level deep features extracted from a pre-trained VGG19~\citep{simonyan2014very}. At each level, the nearest-neighbor field (NNF~\citep{barnes2010generalized}) built on deep features is used to guide the local color transfer in the image domain. The local color transfer considers a linear transform at every pixel, enforcing both local smoothness and non-local constraints to avoid inconsistencies. Then the transferred result, whose appearance becomes much closer to the reference, helps the NNF to be refined at the next level. From coarse to fine,  dense correspondences between features ranging from high-level semantics to low-level details can be built as the differences between two input images are gradually reduced. Therefore, for the image pairs which share similar semantic content but demonstrate significant differences in appearance, our approach is able to achieve natural and consistent color transfer effects, which is challenging for the existing solutions.}
	
	\hmm{In addition to single reference color transfer, our approach can be easily extended to handle multiple references in a similar manner, which provide even richer reference content to help achieve stronger semantic matching. Our algorithm generalizes \emph{one-to-one} NNF search to \emph{one-to-many}, and enforces piecewise smoothness by placing it into a Markov Random Field (MRF) optimization framework.}
	
	\hmm{In brief, our major technical contributions are:}
	\begin{enumerate}
		\item \hmm{We present a novel progressive color transfer framework, which jointly optimizes dense semantic correspondences in the deep feature domain and the local color transfer in the image domain.}
		\item \hmm{We present a new local color transfer model, which is based on a pixel-granular linear function, avoiding local structural distortions and preserving global coherence by enforcing both local and global constraints.}
		\item \hmm{We extend our \emph{one-to-one} color transfer to \emph{one-to-many}, which further improves result quality and robustness through effectively avoiding content mismatching.}
	\end{enumerate}
	
	\hmm{We show how our local color transfer technique can be effectively applied to a variety of real scenarios, such as makeup transfer and time-lapse from images. Our technique can also be used to transfer colors to a gray image, known as the colorization problem.}
	
	\section{Related Work}
	\hmm{Color transfer can be applied to either grayscale or color source images. Transferring colors to a grayscale image, known as colorization, is a well-studied problem. Early approaches to address this issue rely on user scribbles and extend them via optimization across similar regions~\citep{levin2004colorization}. Recently, learning-based algorithms have been used for automatic image colorization~\citep{zhang2016colorful, iizuka2016let}, but these methods have to learn image statistics from large extensive datasets. Given one reference image instead of user input, some automatic methods transfer the chrominance between pixels containing similar statistics~\citep{arbelot2017local, welsh2002transferring}. \hmm{\citet{hechen2018deepcolor} integrate reference images into a learning-based method to achieve automatic examplar-based colorization but it is also limited to only transfer the chrominance.} Our method is applicable to colorization using reference images of the same class, but our focus is on \hmm{both luminance and chrominance} transfer between a color image pair.}
	
	\begin{figure*}[t]
		\footnotesize
		\setlength{\tabcolsep}{0.003\linewidth}
		\begin{tabular}{ccccccc}
			& & $L=5$ & $L=4$ & $L=3$ & $L=2$ & $L=1$
			\\
			\includegraphics[width=0.135\linewidth]{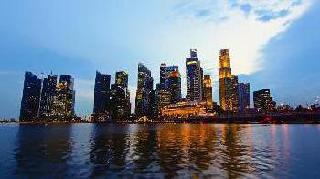}&
			\includegraphics[width=0.135\linewidth]{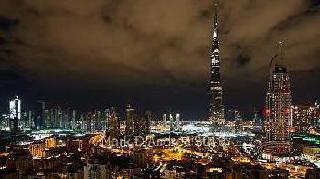}&
			\includegraphics[width=0.135\linewidth]{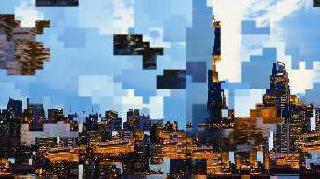}&
			\includegraphics[width=0.135\linewidth]{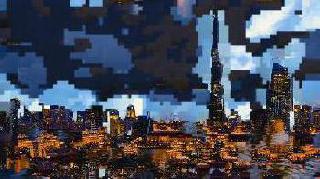}&
			\includegraphics[width=0.135\linewidth]{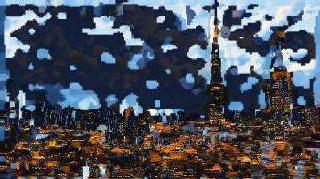}&
			\includegraphics[width=0.135\linewidth]{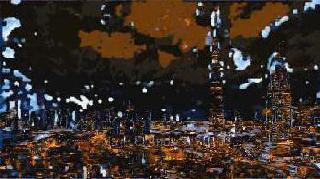}&
			\includegraphics[width=0.135\linewidth]{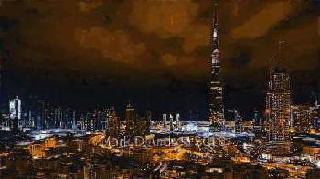}
			\\
			Reference & Source 1 & & & & &
			\\
			&
			\includegraphics[width=0.135\linewidth]{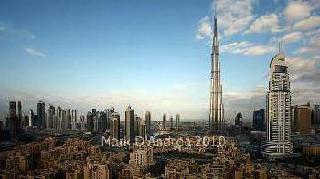}&
			\includegraphics[width=0.135\linewidth]{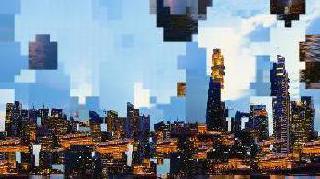}&
			\includegraphics[width=0.135\linewidth]{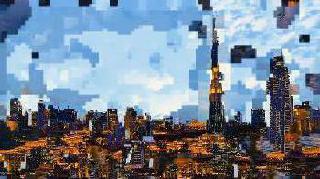}&
			\includegraphics[width=0.135\linewidth]{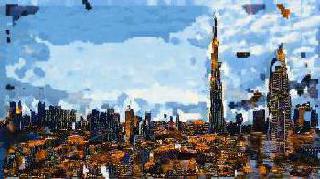}&
			\includegraphics[width=0.135\linewidth]{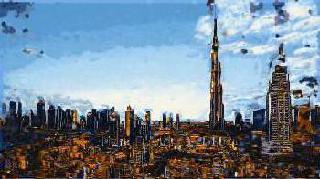}&
			\includegraphics[width=0.135\linewidth]{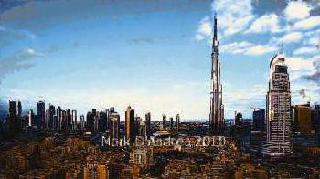}
			\\
			& Source 2 & & & & &
		\end{tabular}
		\caption{\hmm{Matching results from the NNF search using features from different layers individually ($reluL\_1, L=5,\dots,1$ in VGG19). We show two transfer pairs with sources of different colors (all from~\cite{shih2013data}): the first blue-to-dark image pair has more distinct colors (upper row) and the second blue-to-blue image pair has similar colors (lower row). It is clear that NNFs based at the coarsest level (\eg, $relu5\_1$) ignore color difference and achieve similar results. However, low-level features (\eg, $relu5\_1$) are sensitive to color appearance and fail to match objects with semantic similarity but different colors. Input images:~\citet{shih2013data}.}}
		\label{fig:match}
	\end{figure*}
	
	\subsection{Single-reference Color Transfer}
	\paragraph{\textbf{Traditional methods}} Global color transfer algorithms apply a spatially-invariant color transformation to an image \hmm{based on global information matching}, such as global color moves (\eg, sepia) and tone curves (\eg, high or low contrast). The seminal work by~\citet{reinhard2001color} matches the mean and standard deviation between the input and the reference in the $l\alpha\beta$ color space.~\citet{pitie2005n} transfer the full 3D color histogram using a series of 1D histograms.~\citet{freedman2010object} compute the transfer for each histogram bin with the mean and variance of pixel values in the bin, which strikes a compromise between mean-variance based methods and histogram based methods. These methods only consider global color statistics, ignoring the spatial layout.
	
	Local color transfer algorithms based on spatial color mappings are more expressive and can handle a broad class of applications~\citep{Bae2006twotone,shih2013data,Shih2014STH,Sunkavalli2010MIH,laffont2014transient}. \hmm{Having local correspondences is necessary for correct local transfers.} \hmm{Some methods identify regional correspondence and transfer color distributions between corresponding regions. They either require the user input to guide sparse correspondence~\citep{an2010user,welsh2002transferring} or rely on automatic image segmentation or clustering algorithms to estimate regional correspondence~\citep{yoo2013local,tai2005local,arbelot2017local,hristova2015style}. Such matches are not yet precise enough, causing some pixels to be transferred to inaccurate colors.}
	
	\hmm{To exploit pixel-level dense correspondences for more spatially complicated color transfer, some analogy-based methods~\citep{hertzmann2001image,shih2013data,laffont2014transient}) rely on an additional image which has similar colors to the source and similar structure to the reference. With this bridging image, building dense correspondences between two inputs gets easier and a locally linear color model is then estimated and applied. However, such a bridging image is not easy to obtain in practice.}
	
	\hmm{Without the bridging image, it is difficult to directly build dense correspondences between two inputs which are vastly different in color appearance. \citet{shen2016regional} propagate sparse correspondence with model fitting and optimization to build dense matching only inside foremost regions. ~\citet{hacohen2011non} introduce a coarse-to-fine scheme in which NNF computation is interleaved with fitting a global parametric color model to gradually narrow down the color gap for matching. We are inspired by this progressive idea, but our method is essentially different from theirs in both dense correspondence estimation and color transfer model fitting. One one side, dependence on the input image pairs is high in their method, for example, two photos of the same scene, because of the use of low-level features (\eg, image patch). Thanks to the integration of deep features, our method supports input pairs with vast differences in scene and appearance. One the other side, their color mapping, although locally refined, is a single global transformation model and thus cannot fit complicated spatial color variation. In contrast, ours is a pixel-level local color transfer model.}
	
	\paragraph{\textbf{Deep network-based methods}} \hmm{Traditional color transfer methods by matching low-level features are unable to reflect higher-level semantic relationships. Recently, deep neural networks have provided a good representation to establish semantically-meaningful correspondence between visually different image pairs, which can be used in style transfer~\citep{gatys2015neural,chen2017stylebank,chen2018stereoscopic}. The work of \enquote{deep photo style transfer}~\citep{luan2017deep} extends the global neural style transfer~\citep{gatys2015neural} to photo-realistic transfer by enforcing local color affine constraints in the loss function. Their regional correspondence relies on semantic segmentation~\citep{chen2016deeplab} of the image pairs.~\citet{luan2017deep} attempt to improve the photorealism of the stylized images via a post-processing step based on the Matting Laplacian of~\citet{levin2008closed}.~\citet{mechrez2017photorealistic} propose an approach based on the Screened Poisson Equation (SPE) to accelerate the post-processing step.}
	
	\hmm{To estimate the semantically dense correspondences between two images,~\citet{liao2017image} present \enquote{deep image analogy} to take advantage of multi-scale deep features. We use the same feature representation but our work has three key differences applicable for color transfer. First, our approach jointly optimizes the dense semantic correspondences and the local color transfer, while~\citet{liao2017image} achieve color transfer via a two-stage approach, starting with building dense correspondence and then post-processing to change the color.} \hmm{Second, to transfer color, our approach optimizes the linear transform model satisfying local and global coherence constraints while theirs directly applies the NNF to replace the low-frequency source color with the corresponding reference color.} Third, our approach can be easily extended to \emph{one-to-many} color transfer, which effectively avoids content-mismatching in \emph{one-to-one} transfer~\citep{liao2017image}.
	
	\subsection{Multi-reference Color Transfer}
	Choosing a proper reference for color transfer is crucial when using a single reference. To ease the burden of reference selection, some methods adopt multiple references, which can be searched and clustered with similar color styles from the Internet by providing a text query~\citep{liu2014autostyle,bonneel2016wasserstein}, or ranked and selected according to semantics and style~\citep{lee2016automatic}. These methods finally apply the global color transfer after getting multiple references. To achieve more precise local transfer,~\citet{Asad2016fastcolor} allow the user to manually give some correspondence guidance between input and multiple references, and then use the locally linear embedding (LLE) method~\citep{Roweis2000lle} to propagate the guidance.
	
	Deep networks have recently been introduced to the task of color transfer \hmm{among multiple images as well}.~\citet{yan2016automatic} learn a highly non-linear color mapping function for automatic photo adjustment by taking the bundled features as the input layer of a deep neural network.~\citet{isola2016image} train generative networks on a dataset of paired images for image appearance transfer, including colors.~\citet{zhu2017unpaired} loosen the constraints to unpaired images. These methods take several hours to train a single color style. The network-generated results are low resolution and often suffer from checkboard artifacts caused by deconvolution layers~\citep{odena2016deconvolution}. Instead, our method only uses features from pre-trained networks for matching. We can support the high-quality transfer of various color styles without training.
	
	\section{Method}  
	\hmm{Our goal is to apply precise local color transfer based on the established dense semantic correspondences between the source and reference images. In our scenario, the two input images share some semantically-related content, but may vary dramatically in appearance or structure. Building dense semantic correspondences between them is known to be a challenging problem. The hand-crafted features fail to reflect semantic information, so we resort to the deep features from an image classification Convolutional Neural Network (CNN) VGG19, which encodes the image gradually from low-level details to high-level semantics. We observe high-level deep features (\eg, $relu5\_1$ layer in VGG19) generally tend to be invariant to color differences, while low-level features (\eg, $relu1\_1$ layer in VGG19) are more sensitive, as demonstrated in~\fref{fig:match}. As their image colors get more similar, their features, especially at lower levels, get easier to match. This inspires our coarse-to-fine approach (Sec. 3.1) to alternately optimize the NNFs between deep features (Sec. 3.2) and perform local color transfer (Sec. 3.3). Thus, the two steps are mutually beneficial.}
	
	\subsection{Overview}\label{subsec:overview}
	\begin{figure}[t]
		\centering
		\includegraphics[width=1.0\linewidth]{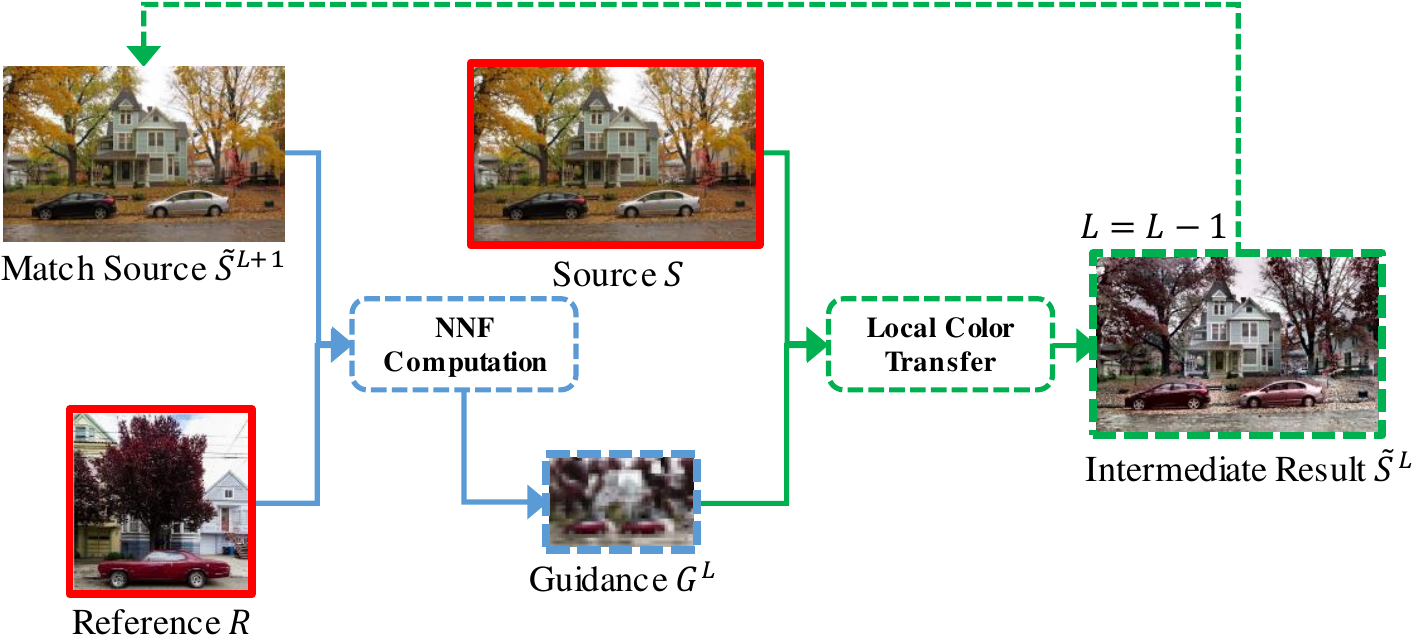}
		\\
		(a) Pipeline
		\\
		\includegraphics[width=1.0\linewidth]{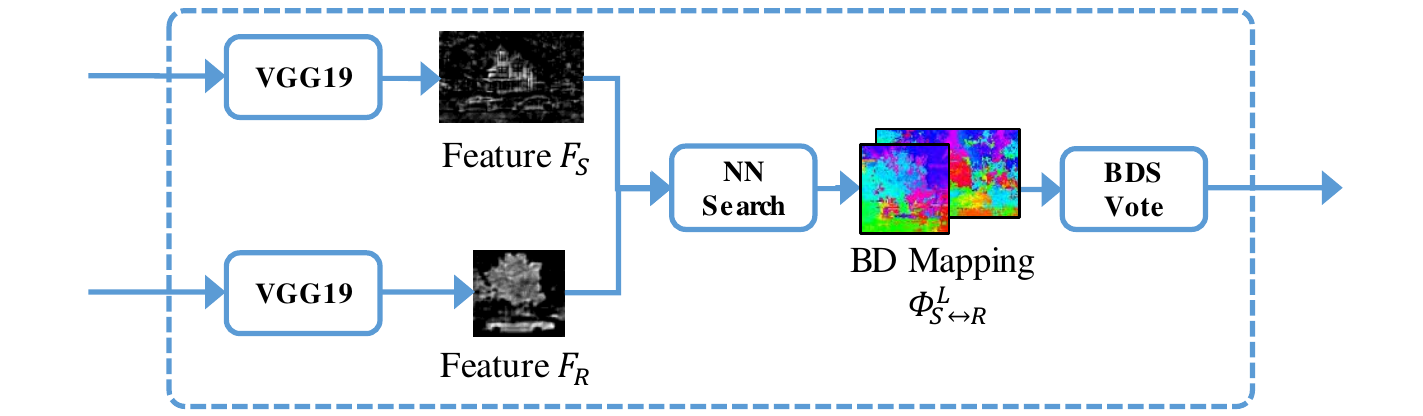}
		\\
		(b) NNF computation
		\includegraphics[width=1.0\linewidth]{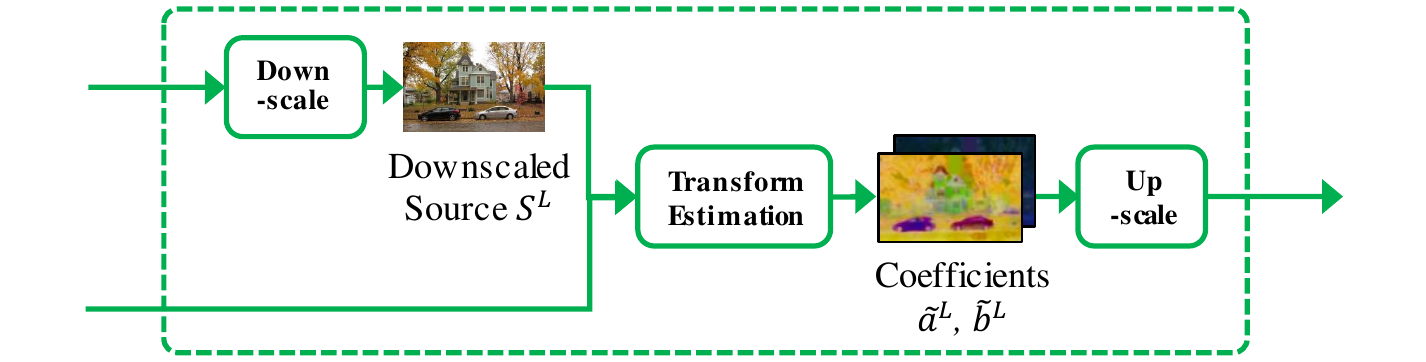}
		\\
		(c) Local color transfer
		\caption{System pipeline with two steps illustrated below. We perform NNF computation in the feature domain (in blue) and apply local color transfer in the image domain (in green). As shown in (a), the original input images are the source image $S$ and the reference image $R$ (in the red frames). At each level, the bidirectional NNFs $\phi_{S \rightarrow R}^L$ and $\phi_{S \rightarrow L}^R$ are computed between features from $\widetilde{S}^{L+1}$ and $R$, and are used to reconstruct $G^L$ as shown in (b). Then in (c), the color transform coefficients $\widetilde{a}^L$ and $\widetilde{b}^L$ are optimized between $G^L$ and the downscaled source image $S^L$, and then upscaled to the full resolution before being applied to $S$. The color transferred result $\widetilde{S}^L$ serves as the input for the matching of the next level $L-1$. The above process repeats from $L=5$ to $L=1$.}
		\label{fig:pipe}
	\end{figure}
	
	\hmm{Given a source image $S$ and a reference image $R$, our algorithm progressively estimates dense correspondence between them and applies accurate local color transfer on $S$, to generate output $S'$ preserving both the structure from $S$ and the color style from $R$.}
	
	Our system pipeline is shown in~\fref{fig:pipe}. At each level $L$, \hmm{there are two steps: NNF computation in the feature domain (\Sref{subsec:nnsearch}) and local color transfer in the image domain (\Sref{subsec:transfer}).} First, we match the reference $R$ to the intermediate source $\widetilde{S}^{L+1}$ using the $reluL\_1$ layer in VGG19 to get bidirectional NNFs in the feature domain and use the NNFs to reconstruct a color guidance $G^L$. Next, we estimate the local color transfer function between the downscaled version of source $S^L$ and $G^L$, upscale the transformation, and apply it to $S$ to get $\widetilde{S}^{L}$. The two steps alternate and mutually assist one another: the NNFs help obtain a more accurate local color transfer, while the color transferred result $\widetilde{S}^L$ serving as the source also helps refine the matching in the next level $L-1$, since $\widetilde{S}^L$ has much more similar colors to the reference than the original source $S$. Both intermediate results (NNFs and $\widetilde{S}^L$) serve as the bridge between both matching and color transfer which occur in different domains. Following this strategy, both steps are gradually refined.
	
	\subsection{Nearest-Neighbor Field Computation}\label{subsec:nnsearch}  
	\begin{figure*}[t]
		\footnotesize
		\setlength{\tabcolsep}{0.003\linewidth}
		\begin{tabular}{cccccc}
			\includegraphics[width=0.153\linewidth]{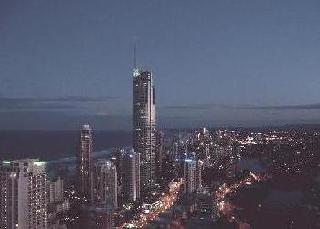}&
			\includegraphics[width=0.165\linewidth]{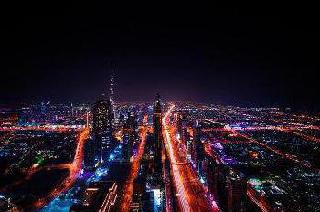}&
			\includegraphics[width=0.153\linewidth]{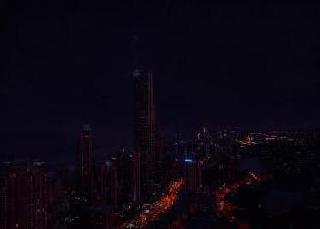} &
			\includegraphics[width=0.153\linewidth]{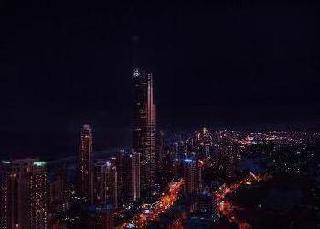} &
			\includegraphics[width=0.153\linewidth]{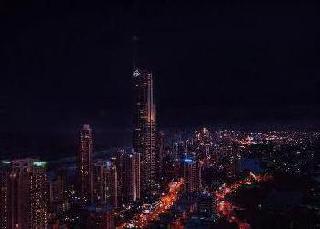} &
			\includegraphics[width=0.153\linewidth]{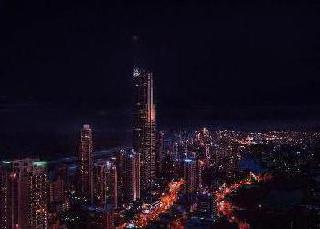}
			\\
			Source &  Reference & $w=0$ & $w=2$ & $w=4$ & $w=8$\\
		\end{tabular}
		\caption{\hmm{Final color transfer results by BDS voting with various completeness weights $w$. Compared to the result by average voting ($w=0$), more reference colors are used as $w$ increases. Input images: Anonymous/pxhere.com.}}  
		\label{fig:bds}
	\end{figure*}
	
	Given the intermediate source $\widetilde{S}^{L+1}$ and the reference $R$ at level $L (L=5,\dots,1)$, our NNF search step builds the bidirectional correspondences between them. Here, $\widetilde{S}^{L+1}$ is the color transferred result (with the same resolution to the source $S$) from the coarser level $L+1$ when $L<5$. At the coarsest level $L=5$, $\widetilde{S}^{L+1}$ is initialized as $S$.
	
	In view of the difficulty of building correct correspondences between $\widetilde{S}^{L+1}$ and $R$ \hmm{(potentially with large appearance variations)}, we perform NNF search in the deep feature domain. Since the CNN will almost always keep the spatial relationship of input images, the NNFs computed in the feature domain can be used in the image domain. To do so, we first feed the $\widetilde{S}^{L+1}$ and $R$ into the VGG19 network~\citep{simonyan2014very} pre-trained on the ImageNet database~\citep{russakovsky2015imagenet} for object recognition. We then extract their feature maps in $reluL\_1$ layer, labeled as $F_S^L$, $F_R^L$ respectively for $\widetilde{S}^{L+1}$, $R$. Each feature map is a 3D tensor with $width \times height \times channel$, and its spatial resolution is \hmm{$1/4^{L-1}$ of the input}.
	
	The mapping function $\phi_{S \rightarrow R}^L$ from $F_{S}^L$ to $F_{R}^L$ is computed by minimizing the following energy function:
	
	\begin{equation}\label{eqn:nnsearch}
	\phi_{S \rightarrow R}^L(p)= \mathop {\arg \min }\limits_q \sum\limits_{x \in N(p), y \in N(q)}\|{\overline F_{S}^L(x)}- {\overline F_{R}^L(y)}\|^2
	\end{equation}
	where $N(p)$ is the patch around $p$. We set the patch size to $3 \times 3$ at each level. For each patch around position $p$ in the source feature $F_S^L$, we find its nearest-neighbor patch around $q = \phi_{S \rightarrow R}^L(p)$ in the reference feature $F_R^L$. $F(x)$ in~\Eref{eqn:nnsearch} is a vector representing all channels of feature map $F$ at position $x$. We use normalized features $\overline F^L(x) = \frac{F^L(x)}{|F^L(x)|}$ in our patch similarity metric, because \hmm{the use of normalized features can achieve stronger invariance~\citep{chuanli2016mrf}.}
	
	The reverse mapping function $\phi_{R \rightarrow S}^L$ from $F_{R}^L$ to $F_{S}^L$ is computed in the same way as~\Eref{eqn:nnsearch} by exchanging $S$ and $R$. Both mappings $\phi_{S \rightarrow R}^L$ and $\phi_{R \rightarrow S}^L$ can be efficiently optimized with the PatchMatch algorithm~\cite{barnes2009patchmatch}, which also implicitly achieves smoothness through aggregation of overlapping patches.
	
	The bidirectional correspondences allow us to use Bidirectional Similarity (BDS) voting~\citep{simakov2008summarizing} to respectively reconstruct the guidance image $G_L$ and the feature map $F_G^L$. $G^L$ serves as the guidance for color transfer in the next step, while $F_G^L$ is used to measure matching errors:
	\begin{equation}\label{eqn:match_err}
	e^L(p)=\|{\overline F_{S}^L(p)}- {\overline F_G^L(p)}\|^2
	\end{equation}
	in~\Eref{eqn:affine_ed}. The BDS voting is performed to average the pixel colors \hmm{and} features from all overlapping nearest-neighbor patches in the reference $R^L$~\footnote{$R^L$ is the same resolution as $F_{R}^L$, downscaled from the reference $R$} \hmm{and} $F_{R}^L$ through the forward NNF $\phi_{S \rightarrow R}^L$ and the backward NNF $\phi_{R \rightarrow S}^L$. The forward NNF enforces coherence (\ie, each patch in the source can be found in the reference), while the backward NNF enforces completeness (\ie, each patch in the reference can be found in the source). By enforcing both coherence and completeness, \hmm{BDS voting can encourage more reference colors in $G^L$ than average voting with solely forward NNF.~\fref{fig:bds} shows a set of final results with various completeness weights (using 2 as the default).}
	
	We show the NNFs $\phi_{R \rightarrow S}^L$, and guidance image $G^L (L=5,\dots,1)$ gradually refined from the coarse layer to the fine layer in~\fref{fig:feature}.
	
	\subsection{Local Color Transfer}\label{subsec:transfer}
	Given the guidance image $G^L$ at the level $L$, we propose a new local color transfer algorithm, which changes the colors of the source $S$ to better match those of $G^L$. Then, we get the color transfer result $\widetilde{S}^L$. Since $S$ and $G^L$ have different resolutions at the coarse levels ($L>1$), it is impossible to build in-place correspondence between $S$ and $G^L$. Instead, we downscale $S$ to $S^L$ to match the resolution of $G^L$, estimate the color transfer function from $S^L$ to $G^L$, \hmm{and upscale the function parameters with an edge-preserving filter before applying it to the full-resolution $S$ to get $\widetilde{S}^L$. $\widetilde{S}^L$ is the intermediate transferred result used for the NNF search at the next level.}
	
	\hmm{Inspired by~\citet{reinhard2001color} which constructs a color transfer function by matching the global means and variances of pixel colors, we model the local color matching as a linear function of each channel in CIELAB color space for every pixel $p$ in $S^L$, denoted as:}
	\begin{equation}\label{eqn:linear}
	\mathcal{T}^L_{p}(S^L(p)) = a^L(p) S^L(p) + b^L(p).
	\end{equation}
	\hmm{(If we consider $b^L(p)$ only with $a^L(p)$ being set to zero, only the means are consistent.)}
	
	We aim to estimate linear coefficients $a^L(p)$ and $b^L(p)$ for each pixel $p$, making the transferred result $\mathcal{T}^L_{p}(S^L(p))$ visually similar to the guidance $G^L(p)$. We formulate the problem of estimating $\mathcal{T}^L$ by minimizing the following objective function consisting of three terms:
	\begin{equation}\label{eqn:affine_energy}
	E(\mathcal{T}^L)=\sum\limits_{p} E_{d}(p) + \lambda_{l}\sum\limits_{p} E_{l}(p) + \lambda_{nl}\sum\limits_{p} E_{nl}(p),
	\end{equation}
	where $\lambda_{l}$ and $\lambda_{nl}$ are trade-off weights (by default, $\lambda_{l} = 0.125$ and $\lambda_{nl} = 2.0$). 
	
	The first data term $E_{d}$ makes the color transfer result similar to the guidance $G^L$:
	\begin{equation}\label{eqn:affine_ed}
	E_d(p) = \hmm{\omega(L)}(1-\overline{e}^L(p))\|\mathcal{T}^L_{p}(S^L(p)) - G^L(p)\|^2,
	\end{equation}
	\hmm{where $\overline{e}^L$ is the normalized matching error in~\Eref{eqn:match_err}, used as the weight to give high confidence to well-matched points. $\omega(L)=4^{L-1}$ is the normalization factor to make this term resolution-independent at different levels.} 
	
	\begin{figure}[t]
		\footnotesize
		\setlength{\tabcolsep}{0.003\linewidth}
		\begin{tabular}{cc}
			\includegraphics[width=0.43\linewidth]{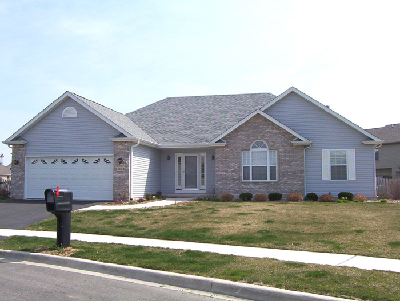}&
			\includegraphics[width=0.565\linewidth]{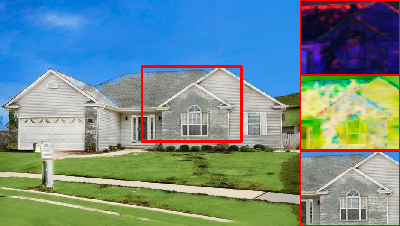}
			\\  
			Source &  $\widetilde{S}^1$ without non-local constraint ($\lambda_{nl}=0.0$)
			\\
			\includegraphics[width=0.43\linewidth]{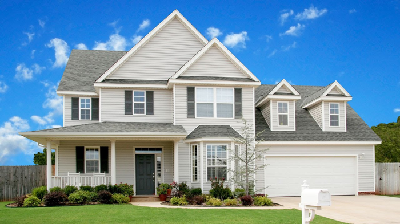}&
			\includegraphics[width=0.565\linewidth]{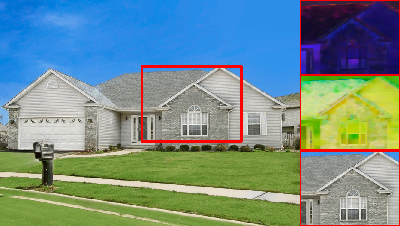}
			\\  
			Reference &  $\widetilde{S}^1$ with non-local constraint ($\lambda_{nl}=2.0$)
		\end{tabular}
		\caption{Comparison with and without non-local constraint \hmm{corresponding to $\lambda_{nl}=0.0$ and $\lambda_{nl}=2.0$ respectively}. Input images:~\citet{luan2017deep}.} 
		\label{fig:nonlocal}
	\end{figure}
	
	The second smoothness term $E_{l}$, \hmm{which is defined in the same manner as the smooth term in WLS-based filter~\citep{farbman2008edge}, encourages locally adjacent pixels to have similar linear transforms while preserving edges in the source $S^L$}:
	\begin{equation}\label{eqn:affine_el}
	E_l(p) = \sum\limits_{q \in N_4(p)} \omega_l(p, q) (\|a^L_{p} - a^L_{q}\|^2+\|b^L_{p} - b^L_{q}\|^2),
	\end{equation}
	\hmm{where $N_4(p)$ is the 4-connected neighborhood at $p$}. As for the smoothness weights, we define them in the same way as in~\citep{lischinski2006interactive}:
	\begin{equation}\label{eqn:weight_el}
	\hmm{\omega_l(p, q) = (\|\ell(p)-\ell(q)\|^\alpha + \epsilon)^{-1}}
	\end{equation}
	where $\ell$ is the luminance channel of $S^L$ and the exponent $\alpha = 1.2$ and the small constant $\epsilon=0.0001$.
	
	The last smoothness term $E_{nl}$ enforces the non-local constraint to penalize global inconsistency. It is based on the assumption that pixels with identical colors in the source should get similar transferred colors in the result. The constraint has been successfully applied in matting~\citep{chen2013image}, intrinsic image decomposition~\citep{zhao2012closed} and colorization~\citep{endo2016deepprop}. We consider the similarity of both color and semantics to compute the non-local term. \hmm{We first apply K-means to cluster all the pixels into $k$ groups according to their feature distance at the coarsest (most semantic) layer $relu5\_1$ (we set $k= 10$).} Inside each cluster, we find the $K$ nearest neighbors in the color space for each pixel $p$ of $S^L$, labeled as $K(p)$ (we set $K=8$). The non-local smoothness term is then defined as:
	\begin{equation}\label{eqn:affine_enl}
	E_{nl}(p) = \sum_{q \in K(p)} {\omega_{nl}(p, q)} \|\mathcal{T}^L_{p}(S^L(p)) - \mathcal{T}^L_{q}(S^L(q))\|^2,
	\end{equation}
	where $\omega_{nl}(p, q) = \frac{\exp(1 - SSD(S^L(p)-S^L(q)))}{K}$ is determined by the color similarity between $p$ and its non-local neighbor $q$ in the CIELAB color space. With the non-local constraints, artifacts are reduced and color transfer is thus more globally consistent as shown in~\fref{fig:nonlocal}.
	
	The closed-form solution of~\Eref{eqn:affine_energy} is very costly due to the irregular sparse matrix structure. Instead, our alternative solution first estimates a good initial guess and then performs a few conjugate gradient iterations, which achieves much faster convergence for $\mathcal{T}^L$. 
	
	We initialize $\mathcal{T}^L$ by applying the global color transformation method~\citep{reinhard2001color} on every local patch. Specifically, taking a patch $N(p)$ centered at pixel $p$ in $S^L$ and in $G^L$, we estimate the initialized $\mathcal{T}^L$ by matching the mean $\mu$ and standard deviation $\sigma$ of the patch pair in each color channel separately:
	\begin{equation}\label{eqn:transfer}
	\begin{split}
	a^L(p) & = \sigma_{G^L(N(p))}/ (\sigma_{S^L(N(p))} + \epsilon)\\
	b^L(p) & = \mu_{G^L(N(p))}-a^L(p) \mu_{S^L(N(p))},
	\end{split}
	\end{equation}
	where $\epsilon$ is used to avoid dividing zero ($\epsilon = 0.002$ for color range $[0, 1]$). We set the patch size to be $3$ for all layers.
	
	The above parameters $a^L(p)$ and $b^L(p)$ are estimated in a low resolution. \hmm{We also apply the WLS-based operator with the smoothness term matching that in~\Eref{eqn:affine_el} to upsample them to the full-resolution, which is guided by source image $S$, obtaining $a^L_\uparrow$ and $b^L_\uparrow$. The smoothness weight is set to $0.024$ by default.}
	
	\hmm{Next, we get}
	\begin{equation}\label{eqn:affine_transform}
	\widetilde{S}^L(p)= a^L_\uparrow(p) S(p) + b^L_\uparrow(p), ~~~\forall p \in \widetilde{S}^L.
	\end{equation}
	
\begin{figure*}[t]
		\footnotesize
		\setlength{\tabcolsep}{0.003\linewidth}
		\begin{tabular}{crccccc}
			Source $/$ Reference & & $L=5$ & $L=4$ & $L=3$ & $L=2$ & $L=1$\\
			\includegraphics[width=0.152\linewidth]{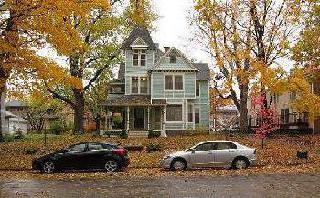} &
			$\phi^L_{S \rightarrow R}$&
			\includegraphics[width=0.152\linewidth]{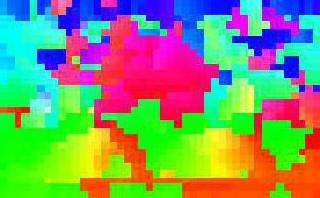}&
			\includegraphics[width=0.152\linewidth]{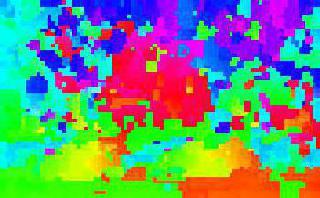}&
			\includegraphics[width=0.152\linewidth]{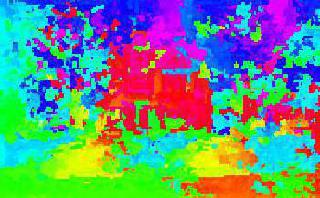}&
			\includegraphics[width=0.152\linewidth]{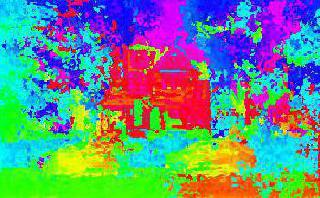}&
			\includegraphics[width=0.152\linewidth]{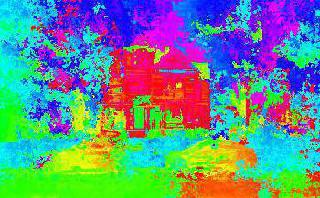}\\   
			\includegraphics[width=0.092\linewidth]{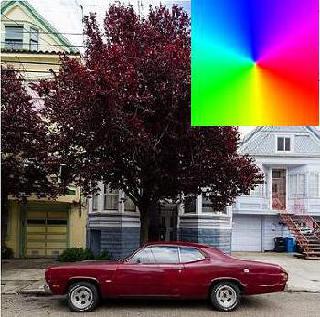} &
			$G^L$&
			\includegraphics[width=0.152\linewidth]{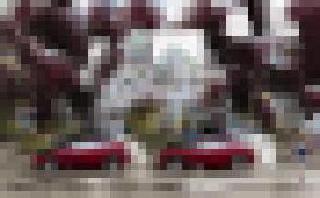}&
			\includegraphics[width=0.152\linewidth]{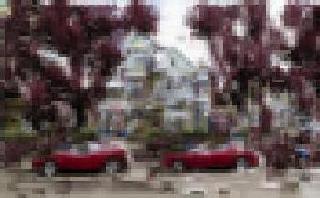}&
			\includegraphics[width=0.152\linewidth]{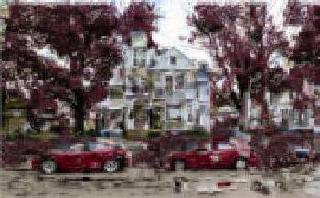}&
			\includegraphics[width=0.152\linewidth]{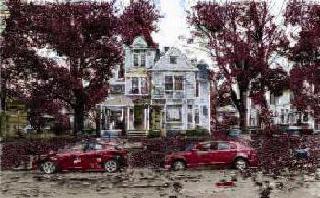}&
			\includegraphics[width=0.152\linewidth]{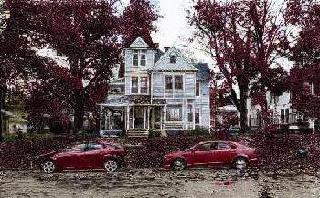}\\
			& $\widetilde{S}^L$&
			\includegraphics[width=0.152\linewidth]{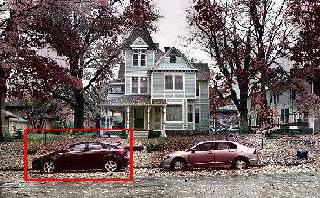}&
			\includegraphics[width=0.152\linewidth]{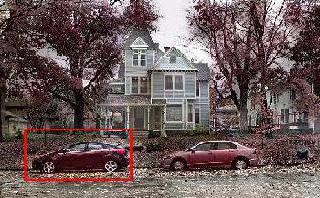}&
			\includegraphics[width=0.152\linewidth]{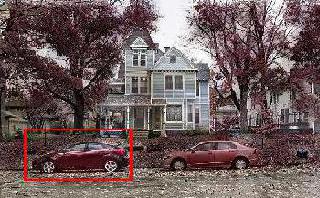}&
			\includegraphics[width=0.152\linewidth]{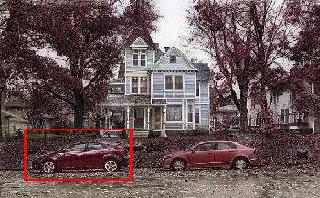}&
			\includegraphics[width=0.152\linewidth]{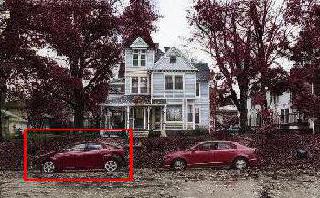}\\
			& close-up&
			\includegraphics[width=0.152\linewidth]{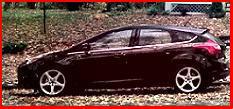}&
			\includegraphics[width=0.152\linewidth]{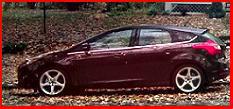}&
			\includegraphics[width=0.152\linewidth]{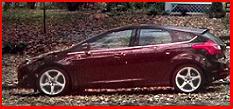}&
			\includegraphics[width=0.152\linewidth]{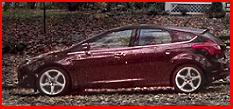}&
			\includegraphics[width=0.152\linewidth]{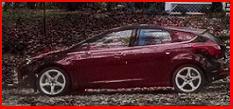}\\
		\end{tabular}
		\caption{\hmm{The NNF $\phi^L_{S \rightarrow R}$, reconstructed results $G^L$ and color transfer results $\widetilde{S}^L$ are gradually refined from high to low level. It can be observed that correspondence between semantically similar objects are built at the coarsest level and refined progressively, for example, both the black car and gray car mapped to the red car, the yellow trees mapped to the red trees, and the house mapped to the white house. As the level goes down, the color appearance of the matched objects gets closer.}}
		\label{fig:feature}
	\end{figure*}

	The result $\widetilde{S}^L$ (in~\fref{fig:feature}) is then used for the NNF search at the next level $L-1$ to update the correspondences. Once the finest layer $L=1$ is reached, $\widetilde{S}^1$ is our final output.
	
	The pseudo code of our implementation is listed in Algorithm~\ref{ag:dct}.
	\begin{algorithm}[t]
		\DontPrintSemicolon
		\SetAlgoLined
		\SetKwInOut{Input}{Input}\SetKwInOut{Output}{Output}
		\Input{Source image $S$ and reference image $R$.}
		
		\BlankLine
		\textbf{Initialization}:\\
		\quad $\widetilde{S}^6=S$.\\
		\For{$L =5$ to $1$}{
			\textbf{NNF search} (\Sref{subsec:nnsearch}):\\
			\quad $F_{S}^L$, $F_{R}^L$ $\leftarrow$ feed $\widetilde{S}^{L+1}$, $R$ to VGG19 and get features.\\
			\quad $\phi_{S\rightarrow R}^L\leftarrow$ map $F_R^L$ to $F_{S}^L$ by~\Eref{eqn:nnsearch}.\\
			\quad $\phi_{R\rightarrow S}^L\leftarrow$ map $F_S^L$ to $F_{R}^L$.\\
			\quad $G^L\leftarrow$ reconstruct $S^L$ with $R^L$ by BDS voting. \\
			\textbf{Local color transfer} (\Sref{subsec:transfer}):\\
			\quad $a^L, b^L \leftarrow$ optimize local linear transform from $S^L$ to $G^L$ by minimizing~\Eref{eqn:affine_energy}.\\
			\quad $a^L_\uparrow, b^L_\uparrow \leftarrow$ upscale and $a^L, b^L$ with WLS-based filter guided by $S$.\\
			\quad $\widetilde{S}^{L} \leftarrow$ transfer the color of $S$ by~\Eref{eqn:affine_transform}.\\
		}
		
		\Output{Color transferred result $\widetilde{S}^1$. }
		
		\caption{\hmm{Single-reference Color Transfer Algorithm}}
		\label{ag:dct}
	\end{algorithm}
	
	\hmm{\subsection{Extension to Multi-reference Color Transfer}}  
	Our algorithm is extendable to multi-reference color transfer. This avoids the difficulty of having to choose a single proper reference image that is suitable for all portions of the source image. The \emph{one-to-one} matching (described in \Sref{subsec:nnsearch}) can be extended to the \emph{one-to-many} matching as follows.  
	\begin{figure}[t]
		\centering
		\includegraphics[width=1.0\linewidth]{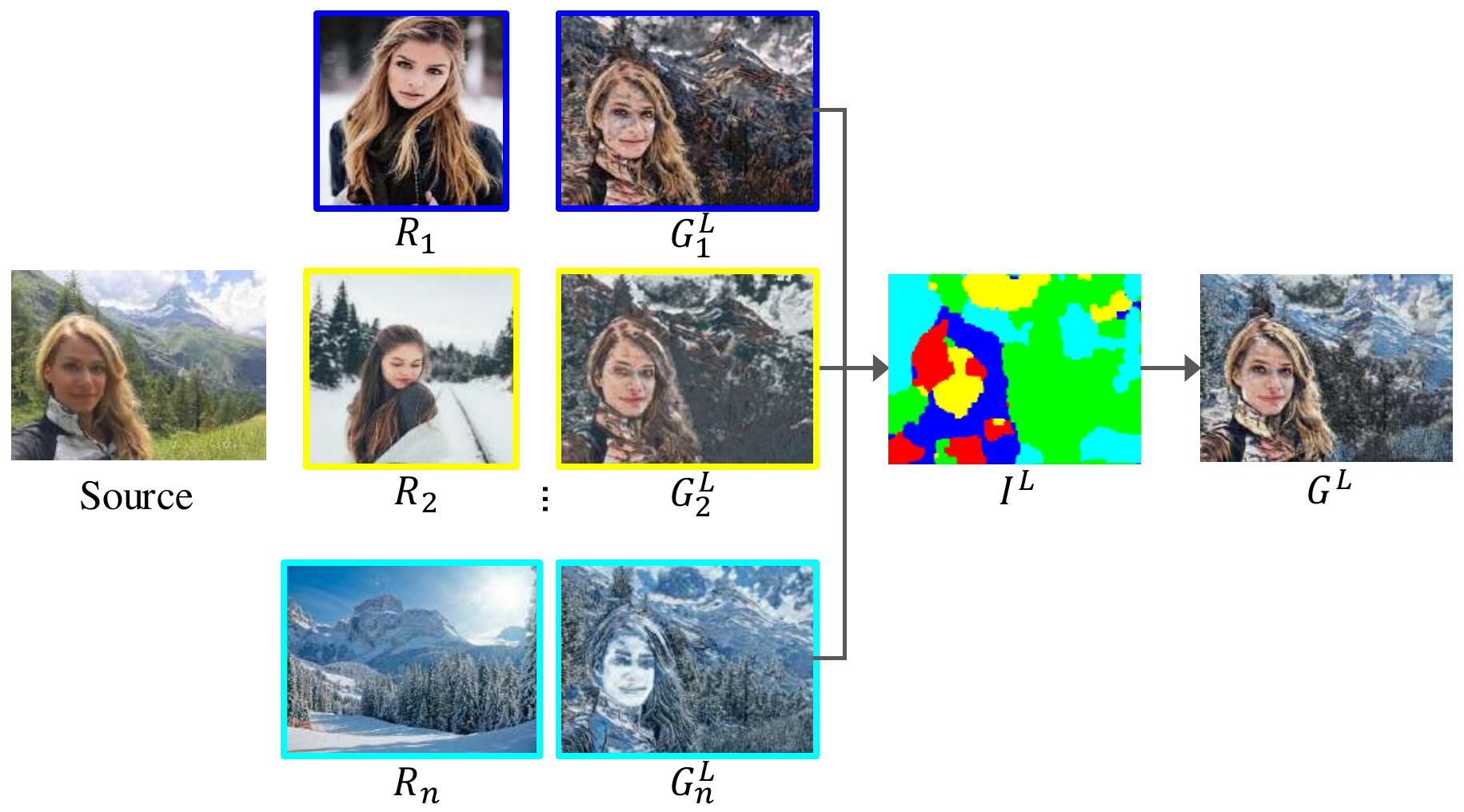} 
		\caption{In a multiple-reference scenario, the image $G^L$ used to guide color transfer is merged from multiple guidances ${G_i^L}. i=1,\dots,n$ reconstructed with bidirectional NNFs.}
		\label{fig:multiple_match}
	\end{figure}  
	Given multiple references ${R_i} (i=1,\dots,n)$, we compute the bidirectional NNFs between $\widetilde{S}^{L+1}$ and every reference $R_i$ at each level $L$, then reconstruct each guidance image $G_i^L$ (shown in~\fref{fig:multiple_match}) using obtained NNFs. Next, we combine these guidance images into a single $G^L$, which requires the selection of the best one from $n$ candidates $G_i^L, (i=1,\dots,n)$ at each pixel $p$. The selection criteria include: (1) how well the reference is matched to the source; (2) how similar the resulting pixel color is compared to the majority of guidance colors; (3) how consistently the indices are selected between the pixel and its neighborhoods. Based on these criteria, we compute the index selection map $I^L$ by minimizing the following objective function:
	\begin{equation}\label{eqn:E_i}
	\hmm{\mathcal{E}(I^L)=\sum\limits_{p}\mathcal{E}_{e}(p)+\beta_c \sum\limits_{p}\mathcal{E}_{c}(p)+\beta_l \sum_{p} \mathcal{E}_{l}(p)},
	\end{equation}
	where \hmm{we set the trade-off weights $\beta_c = 0.2$ and $\beta_l = 0.08$.}
	
	As in~\Eref{eqn:match_err}, $e_i^L$ be the feature error map of $i$-th NNFs. \hmm{The first term penalizes the feature dissimilarity between $S$ and $R_i$ for each pixel at the layer $reluL\_1$:}
	\begin{equation}\label{eqn:m_err}
	\mathcal{E}_{e}(p) = \omega_{e}(L)e_i^L(p)
	\end{equation}
	where \hmm{$\omega_{e}(L)=4^{L-5}$ is the normalization factor for different levels}.
	
	\hmm{The second term measures the difference between each guidance color and the guidance \enquote{majority} color at every pixel. To compute the majority color, we first build a per-pixel color histogram of guidance colors with $n$ bins of each channel (we set $n=8$), and take the mean of the colors that fall in the densest bin.}
	\begin{equation}\label{eqn:m_maj}
	\mathcal{E}_{c}(p) = \omega_{c}(L)\|G^L_i(p)-majority(G^L_i(p))\|^2
	\end{equation}
	where \hmm{$\omega_{c}(L)=\omega_{e}(L)$ is the normalization factor}.
	
	The third term measures the local smoothness, which encourages neighboring features in the combination result to be consistent:
	\begin{equation}
	\mathcal{E}_{l}(p) = \omega_{l}(L) \sum\limits_{q \in N_4(p)} \|{\overline F_{G_i}^L(p)}- {\overline F_{G_j}^L(p)}\|^2 + \|{\overline F_{G_i}^L(q)}- {\overline F_{G_j}^L(q)}\|^2,
	\label{eqn:smooth}
	\end{equation}
	where $i=I^L(p)$ and $j=I^L(q)$ and where \hmm{$\omega_{l}(L)=2^{L-5}$ is the normalization factor}.
	
	\Eref{eqn:E_i} formulates a Markov Random Field (MRF) problem over the 2D spatial domain, which can be efficiently solved by using multi-label graph cut~\citep{kolmogorov2004energy}. To obtain a good initialization for the optimization, $I^L(p)$ is initialized by minimizing only the data terms ($\mathcal{E}_{e}(p)$ and $\mathcal{E}_{c}(p)$). After solving $I^L$, we obtain a single guidance image $G^L$ by simply merging all results from multiple references, \ie, $G^L(p)=G^L_{I^L(p)}(p)$. Then, $G^L$ is used for the following step of local color transfer described in \Sref{subsec:transfer}. The right image of~\fref{fig:multiple_match} shows the optimal reference index map $I^L$ and the merged guidance $G^L$. Compared to single-reference matching, it can effectively solve the content-mismatch problem in situations where it is difficult to find a suitable single source.
	
	\vspace*{.15in}
	\section{Evaluation and Results}
	\subsection{Performance}
	\hmm{Our core algorithm was developed in CUDA. All of our experiments were conducted on a PC with an Intel E5 2.5GHz CPU and an NVIDIA Tesla K40c GPU. The runtime is approximately 60 seconds for single-reference color transfer with an approximate resolution of $700 \times 500$. There are two bottlenecks in the processing: the deep PatchMatch ($\sim$40 seconds), which needs to compute patch similarities on hundreds of feature channels, and the optimization of local color transform (approximately 10--20 seconds), which requires solving large sparse equations. For multi-reference color transfer, the total runtime involves the time of deep PatchMatch which is proportional to the number of references, the optimization of local color transform (same as single-reference) and the solution of MRF ($\sim$5 seconds).}
	\begin{figure*}[t]
		\footnotesize
		\setlength{\tabcolsep}{0.003\linewidth}
		\scalebox{0.96}
		{\begin{tabular}{ccccccc}
				Source & Reference & & $L=5$ & $L=3$ & $L=1$ & Result \\
				\includegraphics[width=0.12\linewidth]{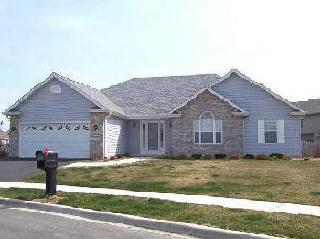} &
				\includegraphics[width=0.12\linewidth]{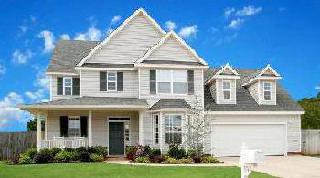} &
				(a) 
				&
				\includegraphics[width=0.185\linewidth]{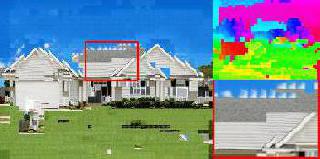} &
				\includegraphics[width=0.185\linewidth]{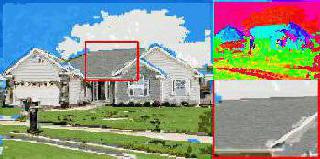} &
				\includegraphics[width=0.185\linewidth]{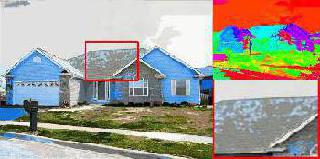} &
				\includegraphics[width=0.125\linewidth]{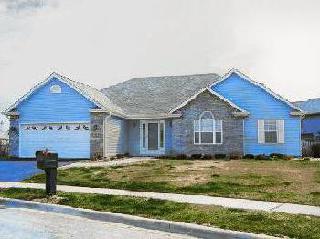}
				\\
				&
				&
				(b) &
				\includegraphics[width=0.185\linewidth]{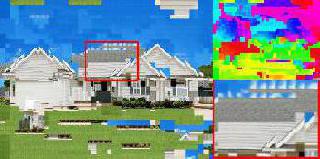} &
				\includegraphics[width=0.185\linewidth]{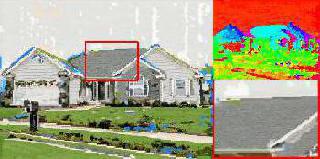} &
				\includegraphics[width=0.185\linewidth]{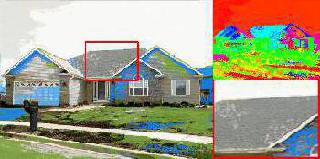} &
				\includegraphics[width=0.12\linewidth]{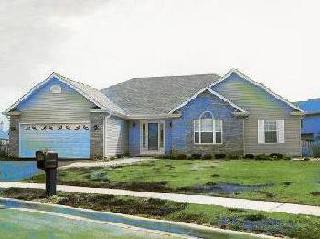}
				\\
				&
				&
				(c) &
				\includegraphics[width=0.185\linewidth]{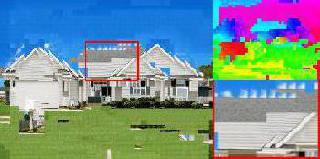} &
				\includegraphics[width=0.185\linewidth]{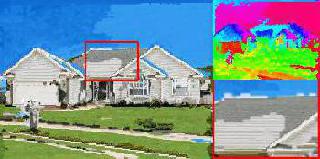} &
				\includegraphics[width=0.185\linewidth]{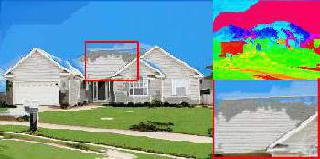} &
				\includegraphics[width=0.12\linewidth]{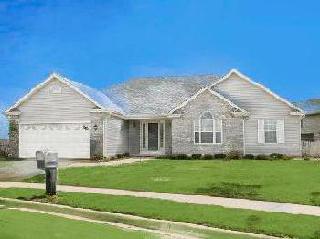}
				\\
				&
				&
				(d) &
				\includegraphics[width=0.185\linewidth]{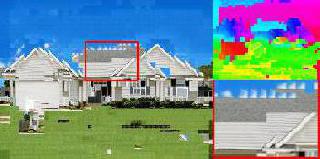} &
				\includegraphics[width=0.185\linewidth]{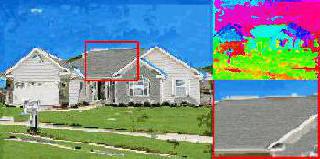} &
				\includegraphics[width=0.185\linewidth]{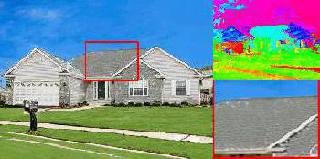} &
				\includegraphics[width=0.12\linewidth]{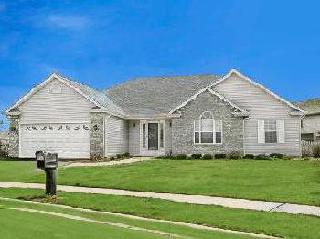}
				\\
				\includegraphics[width=0.12\linewidth]{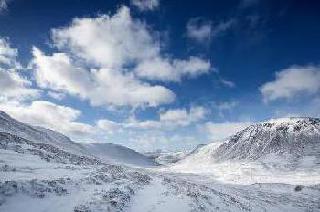} &
				\includegraphics[width=0.12\linewidth]{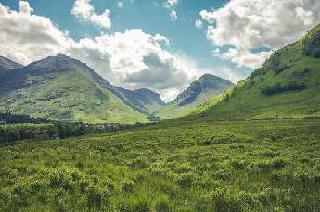} &
				(a) 
				&
				\includegraphics[width=0.185\linewidth]{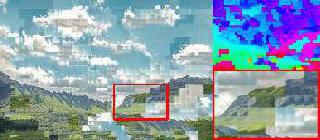} &
				\includegraphics[width=0.185\linewidth]{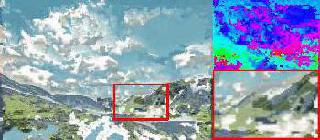} &
				\includegraphics[width=0.185\linewidth]{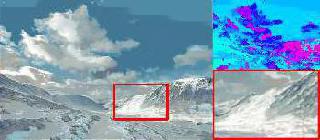} &
				\includegraphics[width=0.12\linewidth]{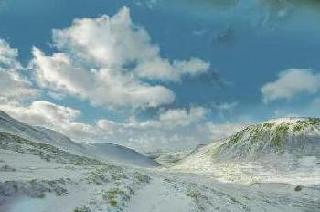}
				\\
				&
				&
				(b) &
				\includegraphics[width=0.185\linewidth]{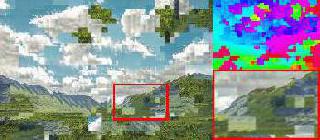} &
				\includegraphics[width=0.185\linewidth]{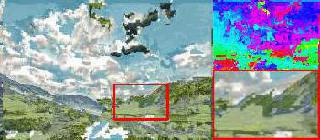} &
				\includegraphics[width=0.185\linewidth]{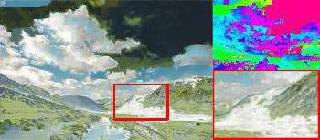} &
				\includegraphics[width=0.12\linewidth]{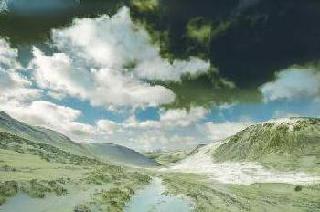}
				\\
				&
				&
				(c) &
				\includegraphics[width=0.185\linewidth]{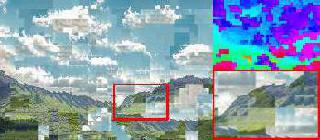} &
				\includegraphics[width=0.185\linewidth]{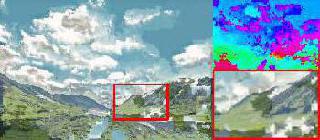} &
				\includegraphics[width=0.185\linewidth]{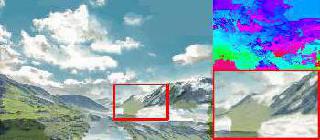} &
				\includegraphics[width=0.12\linewidth]{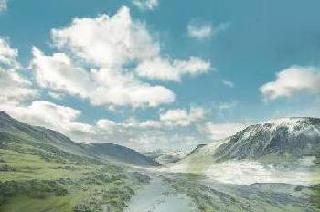}
				\\
				&
				&
				(d) &
				\includegraphics[width=0.185\linewidth]{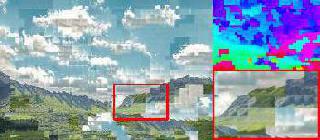} &
				\includegraphics[width=0.185\linewidth]{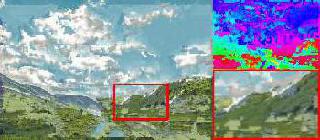} &
				\includegraphics[width=0.185\linewidth]{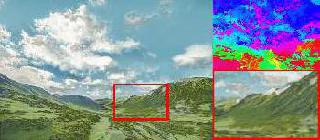} &
				\includegraphics[width=0.12\linewidth]{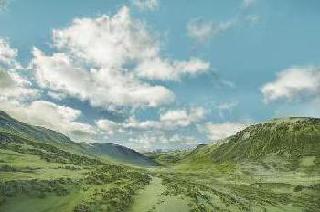}
				\\
				\includegraphics[width=0.12\linewidth]{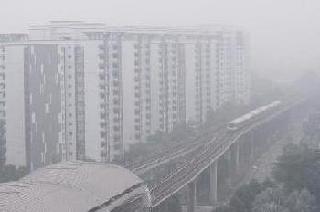} &
				\includegraphics[width=0.12\linewidth]{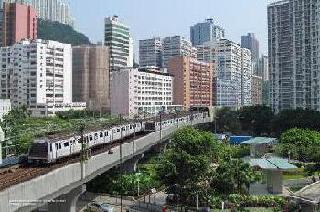} &
				(a) 
				&
				\includegraphics[width=0.185\linewidth]{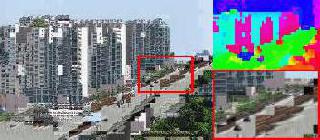} &
				\includegraphics[width=0.185\linewidth]{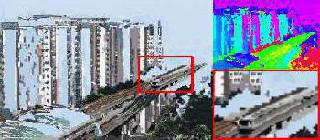} &
				\includegraphics[width=0.185\linewidth]{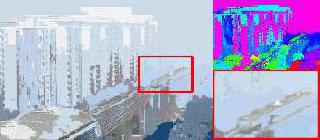} &
				\includegraphics[width=0.12\linewidth]{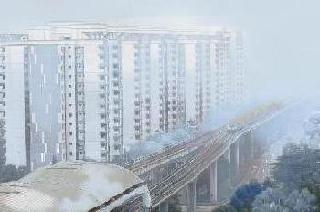}
				\\
				&
				&
				(b) &
				\includegraphics[width=0.185\linewidth]{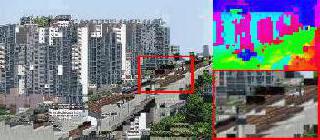} &
				\includegraphics[width=0.185\linewidth]{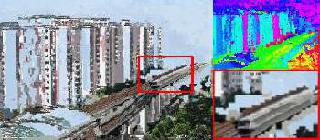} &
				\includegraphics[width=0.185\linewidth]{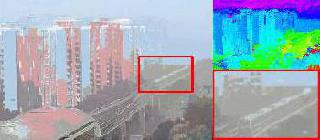} &
				\includegraphics[width=0.12\linewidth]{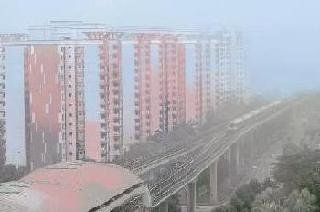}
				\\
				&
				&
				(c) &
				\includegraphics[width=0.185\linewidth]{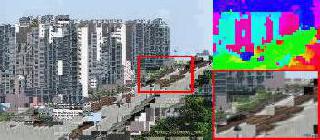} &
				\includegraphics[width=0.185\linewidth]{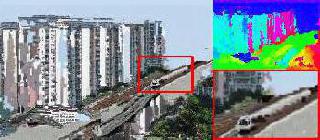} &
				\includegraphics[width=0.185\linewidth]{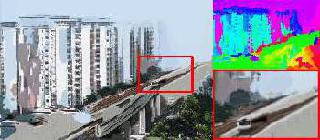} &
				\includegraphics[width=0.12\linewidth]{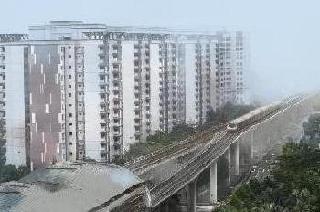}
				\\
				&
				&
				(d) &
				\includegraphics[width=0.185\linewidth]{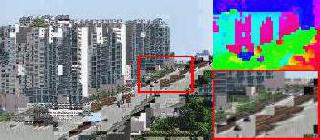} &
				\includegraphics[width=0.185\linewidth]{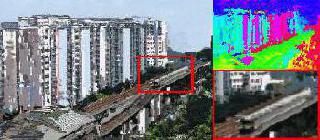} &
				\includegraphics[width=0.185\linewidth]{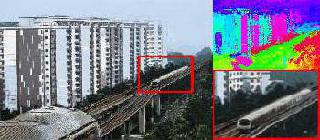} &
				\includegraphics[width=0.12\linewidth]{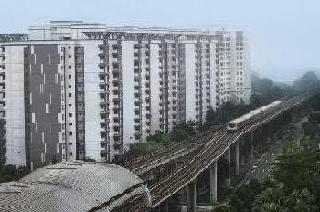}
				\\
			\end{tabular}
		}
		\caption{\hmm{Comparisons of results by separate methods and our joint method. Every example includes four rows with each showing the intermediate NNFs computed by one method and the final color transfer result using our local color transfer. (a) represents the NNF estimation of the color image pair, (b) shows NNF on the grayscale image pair, (c) represents NNF by~\citet{liao2017image} and (d) shows NNF by our joint optimization.} Input images:~\citet{luan2017deep}.} 
		\label{fig:ablation_post}
	\end{figure*}
	
	\begin{figure*}[t]
		\footnotesize
		\setlength{\tabcolsep}{0.003\linewidth}
		\begin{tabular}[t]{ccccccc}
			\includegraphics[width=0.162\linewidth]{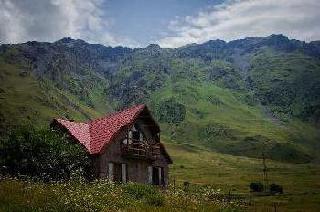} &
			\includegraphics[width=0.161\linewidth]{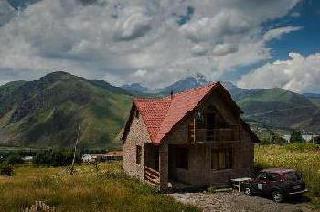} &
			\includegraphics[width=0.162\linewidth]{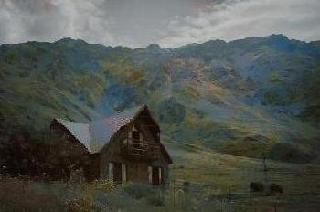} &
			\includegraphics[width=0.162\linewidth]{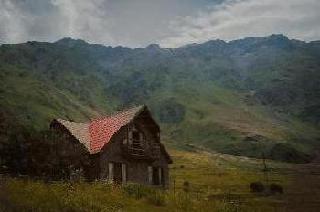} &
			\includegraphics[width=0.162\linewidth]{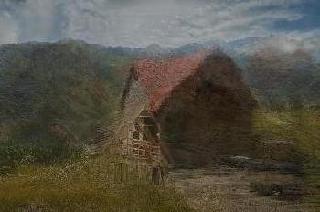} &
			\includegraphics[width=0.162\linewidth]{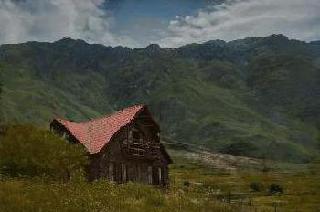}
			\\
			&
			\includegraphics[width=0.158\linewidth]{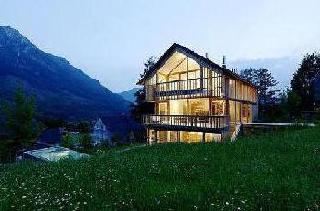} &
			\includegraphics[width=0.162\linewidth]{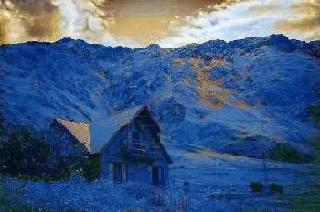} &
			\includegraphics[width=0.162\linewidth]{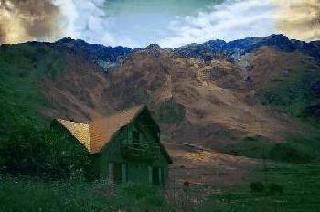} &
			\includegraphics[width=0.162\linewidth]{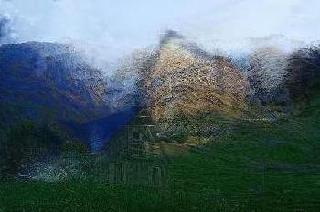} &
			\includegraphics[width=0.162\linewidth]{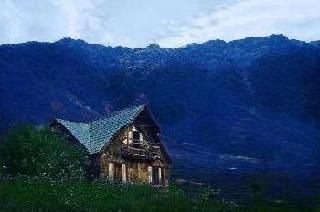}
			\\
			Source & Reference & Luminance features & Color features & SIFT & Deep features\\
		\end{tabular}
		\caption{\hmm{Comparisons between results by estimating correspondence with different features. Input images: Anonymous/tajawal.ae and Anonymous/winduprocketapps.com.}}  
		\label{fig:ablation_feature}
	\end{figure*}
	
	\begin{figure*}[t]
		\footnotesize
		\setlength{\tabcolsep}{0.003\linewidth}
		\begin{tabular}{ccccc}
			\includegraphics[width=0.196\linewidth]{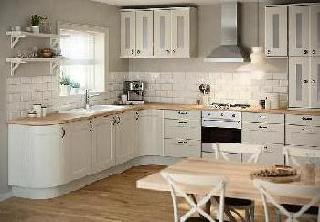} &
			\includegraphics[width=0.185\linewidth,height=0.138\linewidth]{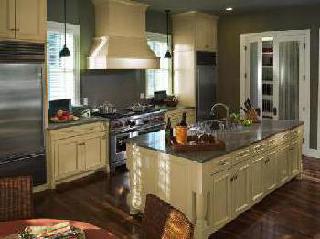} &
			\includegraphics[width=0.196\linewidth]{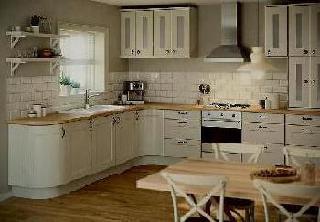} &
			\includegraphics[width=0.196\linewidth]{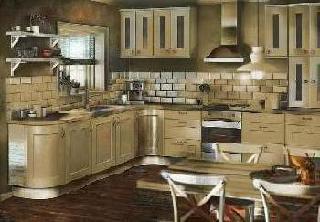} &
			\includegraphics[width=0.196\linewidth]{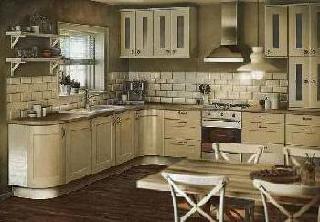}
			\\
			\includegraphics[width=0.196\linewidth]{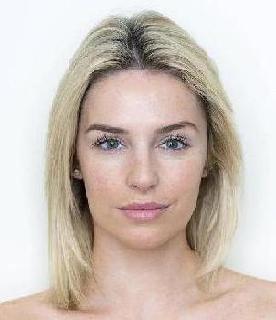} &
			\includegraphics[height=0.225\linewidth]{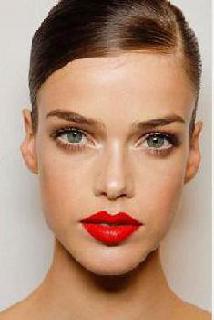} &
			\includegraphics[width=0.196\linewidth]{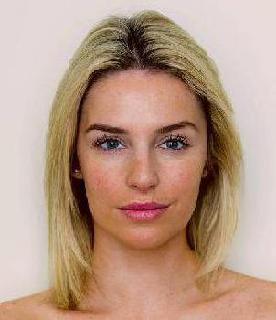} &
			\includegraphics[width=0.196\linewidth]{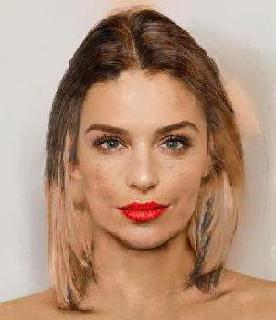} &
			\includegraphics[width=0.196\linewidth]{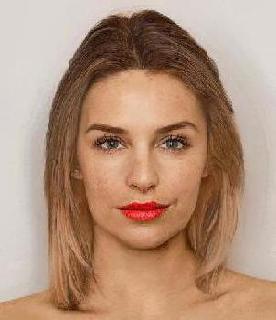}
			\\
			\includegraphics[width=0.196\linewidth]{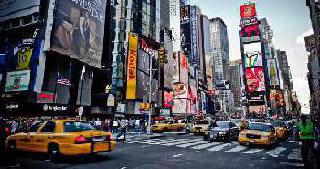} &
			\includegraphics[width=0.185\linewidth]{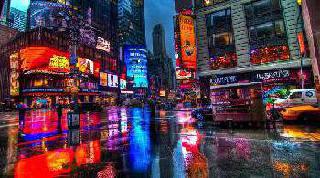} &
			\includegraphics[width=0.196\linewidth]{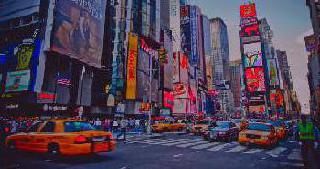} &
			\includegraphics[width=0.196\linewidth]{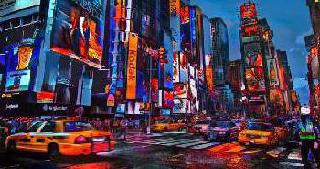} &
			\includegraphics[width=0.196\linewidth]{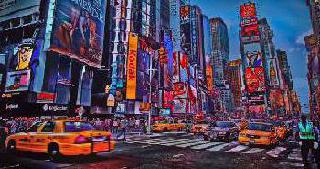}
			\\
			Source & Reference & Global~\citep{reinhard2001color} & Local & Our color transfer
		\end{tabular}
		\caption{\hmm{Comparisons between results by different color transfer methods applied to transfer the color of the aligned reference to the source. Input images: Anonymous/diy.com and~\citet{luan2017deep}}.} 
		\label{fig:ablation_color}
	\end{figure*}
	
	\begin{figure*}[t]
		\footnotesize
		\setlength{\tabcolsep}{0.003\linewidth}
		\begin{tabular}{cccccc}
			Source & Reference & (a)/(d) & (b)/(e) & (c)/(f)
			\\
			\includegraphics[width=0.135\linewidth]{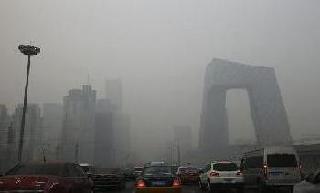} &
			\includegraphics[width=0.135\linewidth, height=0.0835\linewidth]{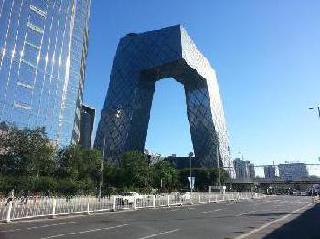} &
			\includegraphics[width=0.225\linewidth]{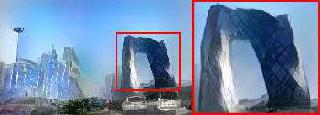} &
			\includegraphics[width=0.225\linewidth]{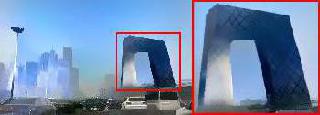} &
			\includegraphics[width=0.225\linewidth]{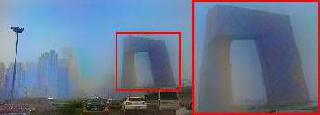}
			\\
			& &
			\includegraphics[width=0.225\linewidth]{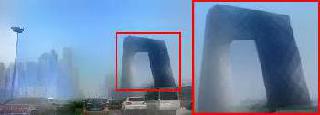} &
			\includegraphics[width=0.225\linewidth]{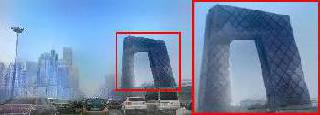} &
			\includegraphics[width=0.225\linewidth]{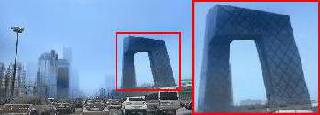} 
			\\
			\includegraphics[width=0.135\linewidth]{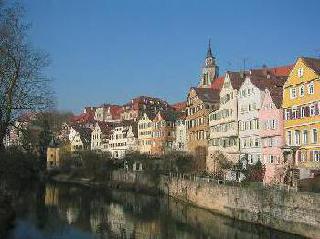} &
			\includegraphics[width=0.135\linewidth]{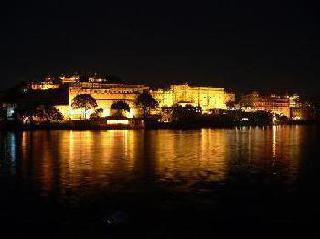} &
			\includegraphics[width=0.225\linewidth]{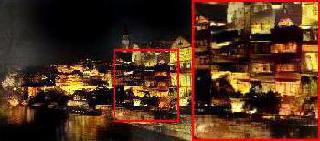} &
			\includegraphics[width=0.225\linewidth]{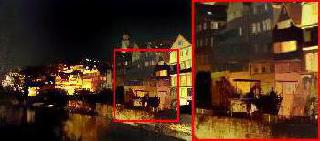} &
			\includegraphics[width=0.225\linewidth]{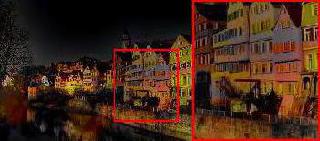} 
			\\
			& &
			\includegraphics[width=0.225\linewidth]{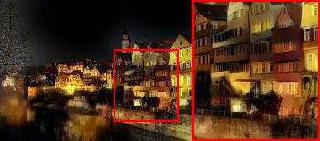} &
			\includegraphics[width=0.225\linewidth]{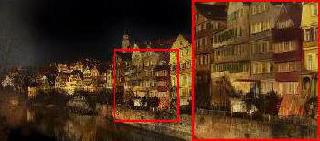} &
			\includegraphics[width=0.225\linewidth]{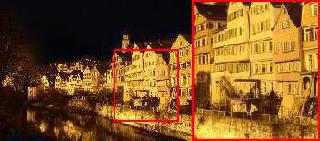} 
			\\  
			\includegraphics[width=0.135\linewidth]{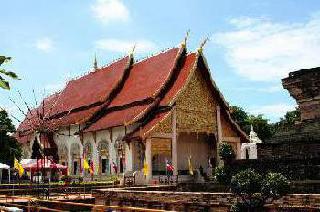} &
			\includegraphics[width=0.135\linewidth]{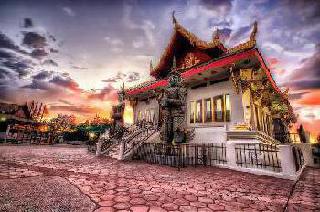} &
			\includegraphics[width=0.225\linewidth]{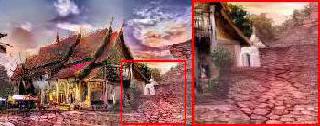} &
			\includegraphics[width=0.225\linewidth]{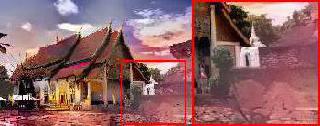} &
			\includegraphics[width=0.225\linewidth]{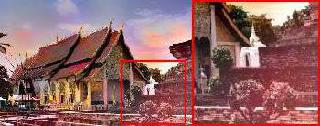} 
			\\
			& &
			\includegraphics[width=0.225\linewidth]{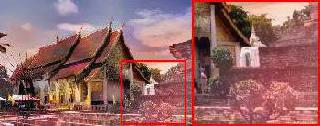} &
			\includegraphics[width=0.225\linewidth]{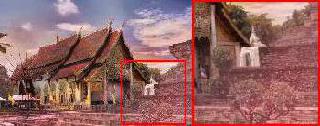} &
			\includegraphics[width=0.225\linewidth]{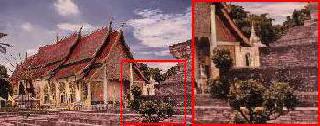} 
			\\
		\end{tabular}
		\caption{\hmm{Comparisons between results by different color transfer methods from the aligned reference to the source. (a) represents the aligned reference by~\citet{gatys2015neural} with a segmentation mask, (b)--(e) are the results refined by~\citet{luan2017deep},~\citet{mechrez2017photorealistic},~\citet{liao2017image} and our local color transfer method respectively, and (f) is the result of our joint optimization approach.} Input images:~\citet{luan2017deep}.}
		\label{fig:ablation_joint}
	\end{figure*}
	\subsection{Evaluation}
	\hmm{We analyze and evaluate the different components of our algorithm through three studies.}
	\hmm{\paragraph{\textbf{Joint optimization}}To verify that correspondence estimation and local color transfer benefit from our interleaved joint optimization process, we conduct an ablation where both steps are separated as two individual steps. In the first step, the NNF computation is conducted down to the finest level using the way described in~\Sref{subsec:nnsearch} or the more advanced method~\citep{liao2017image}. Additionally, in the second step, our color transform method (\Sref{subsec:transfer}) is applied based on the obtained correspondences. To reduce the influence brought by color, we test it on grayscale image pairs as well as color image pairs, and their results are respectively shown in the $1st$ and $2nd$ rows of each example in~\fref{fig:ablation_post}. Besides our NNF search method in~\Sref{subsec:nnsearch}, we also use the approach by~\citet{liao2017image} to compute the dense correspondence shown in the $3rd$ row in~\fref{fig:ablation_post}.}
	\hmm{Compared to the joint optimization (last column in~\fref{fig:ablation_post}), we find that in our method, without local color transfer as the bridge to connect the correspondence estimations of two consecutive levels, the semantic matching of the higher level barely influences the matching of the lower levels. Thus, the final correspondences are dominated by low-level features (\eg, luminance and chrominance for color images, or luminance for gray images) and the subsequent color transfer results are semantically incorrect.~\citet{liao2017image} build a connection between two consecutive levels by partially blending higher-level reconstructed features for lower-level NNF computation, so their result can preserve the semantic correspondence. However, as the unchanged lower-level features are blended with the reconstructed features, the neural representations are still influenced by the color discrepancy and may misalign at fine-scale image structures. This method is suitable for style transfer but may result in distortion artifacts for color transfer, for example, the mottled roof in the $1st$ example and the unclear building boundaries in the $3rd$ example. In contrast, our method can preserve the fine image details and the semantic relationship because of the joint optimization.}
	\hmm{\paragraph{\textbf{Deep features for correspondences}}In recent works, deep features have been verified to be robust enough to match semantically similar objects despite large appearance differences~\citep{chuanli2016mrf,liao2017image}. We designed a study to validate their importance in our progressive method. We replace deep features with low-level features (luminance, color and SIFT). For the image pair with very similar appearance like the first row in~\fref{fig:ablation_feature}, color features are sufficient to build correspondence. However, for the pair with the larger differences, only deep features can match semantically similar objects, such as the houses in~\fref{fig:ablation_feature}.}
	\hmm{\paragraph{\textbf{Local color transfer}}To verify the robustness of our local color transfer method for consistent and faithful color effects, we first evaluate the effectiveness of our optimization in the progressive algorithm, and then apply our local color transfer as a post-processing step to compare with other color transfer approaches.}  
	
	\begin{figure*}[t]
		\footnotesize
		\setlength{\tabcolsep}{0.003\linewidth}
		\begin{tabular}{ccccc}
			\includegraphics[width=0.192\linewidth]{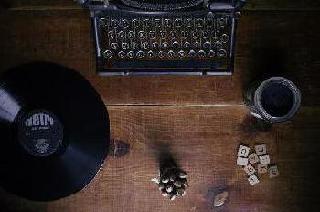}&
			\includegraphics[width=0.192\linewidth]{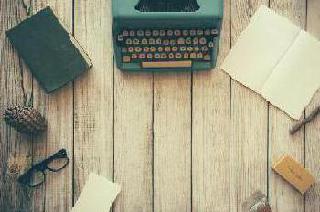}&
			\includegraphics[width=0.192\linewidth]{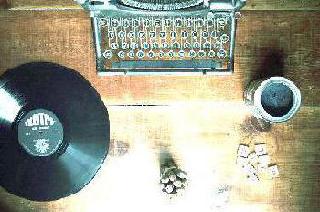}&
			\includegraphics[width=0.192\linewidth]{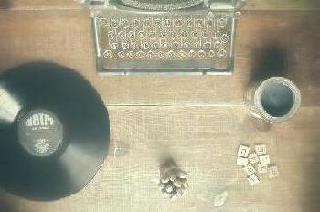}&
			\includegraphics[width=0.192\linewidth]{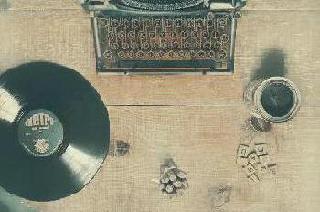}
			\\
			\includegraphics[width=0.192\linewidth]{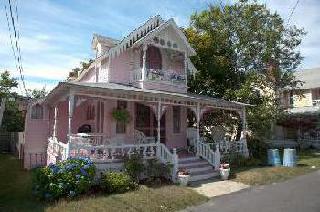}&
			\includegraphics[width=0.192\linewidth]{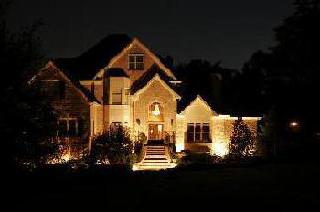}&
			\includegraphics[width=0.192\linewidth]{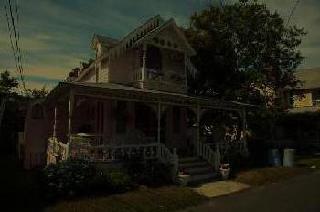}&
			\includegraphics[width=0.192\linewidth]{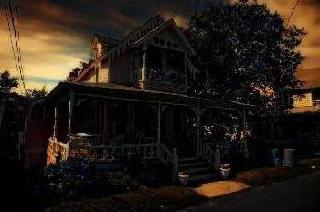}&
			\includegraphics[width=0.192\linewidth]{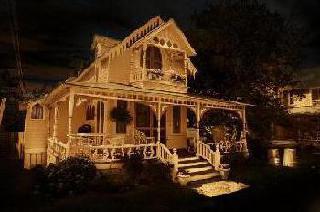}
			\\
			\includegraphics[width=0.192\linewidth]{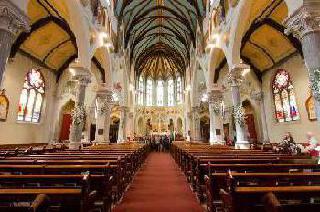}&
			\includegraphics[width=0.192\linewidth,height=0.128\linewidth]{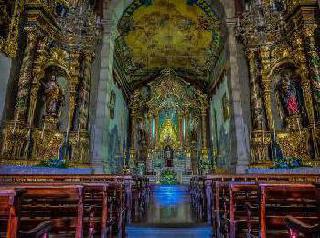}&
			\includegraphics[width=0.192\linewidth]{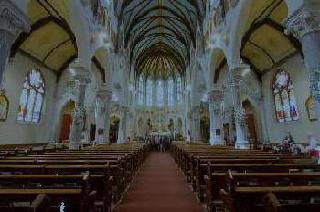}&
			\includegraphics[width=0.192\linewidth]{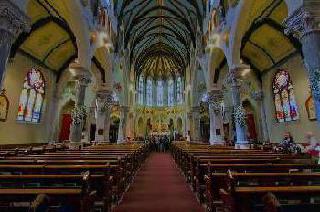}&
			\includegraphics[width=0.192\linewidth]{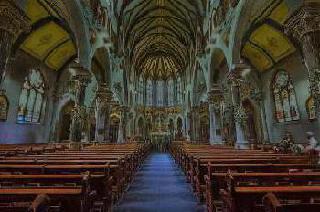}
			\\
			Source & Reference & \citet{reinhard2001color} & \citet{pitie2005n} & Ours\\
		\end{tabular}
		\caption{Comparison with global color transfer methods~\citep{reinhard2001color} and~\citep{pitie2005n}. \hmm{The test images} are from~\citet{luan2017deep}. Input images:~\citet{luan2017deep}.}
		\label{fig:comp_set1}
	\end{figure*}
	
	\begin{figure*}[t]
		\footnotesize
		\setlength{\tabcolsep}{0.003\linewidth}
		\begin{tabular}{cccccccc}
			\includegraphics[height=0.095\linewidth]{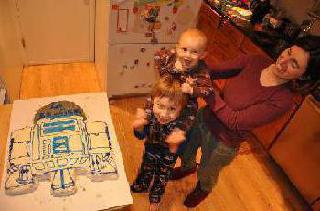}&
			\includegraphics[height=0.095\linewidth]{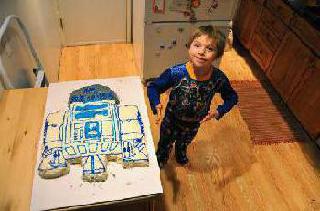}&
			\includegraphics[height=0.095\linewidth]{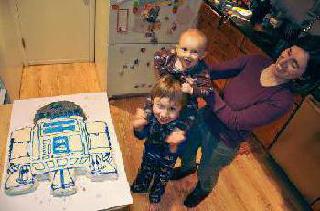}&
			\includegraphics[height=0.095\linewidth]{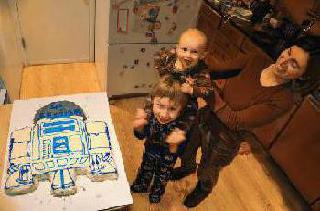} &
			\hspace{0.02in}
			\includegraphics[height=0.095\linewidth]{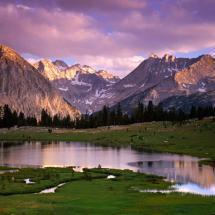}&
			\includegraphics[height=0.095\linewidth]{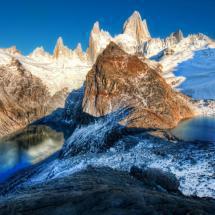}&
			\includegraphics[height=0.095\linewidth]{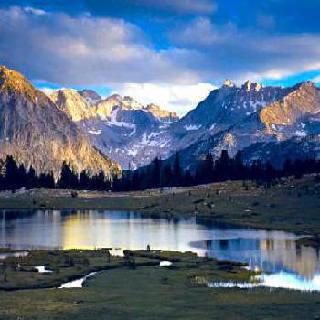}&
			\includegraphics[height=0.095\linewidth]{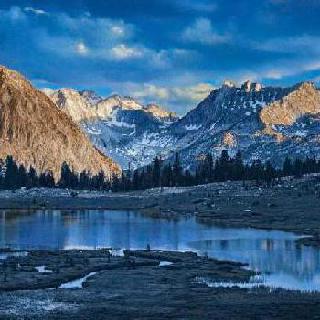}
			\\
			\includegraphics[height=0.095\linewidth]{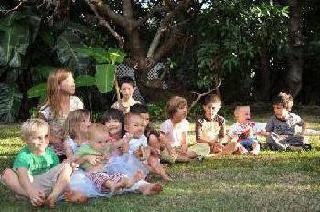}&
			\includegraphics[height=0.095\linewidth]{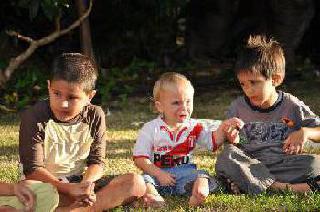}&
			\includegraphics[height=0.095\linewidth]{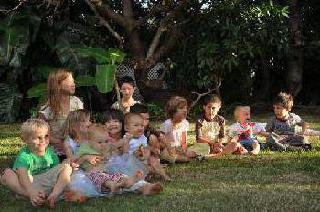}&
			\includegraphics[height=0.095\linewidth]{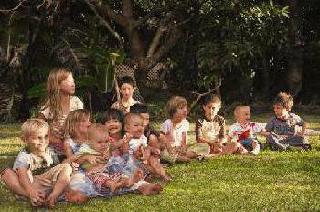}&
			\hspace{0.02in}
			\includegraphics[height=0.095\linewidth]{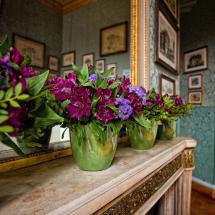}&
			\includegraphics[height=0.095\linewidth]{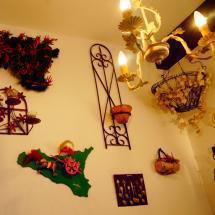}&
			\includegraphics[height=0.095\linewidth]{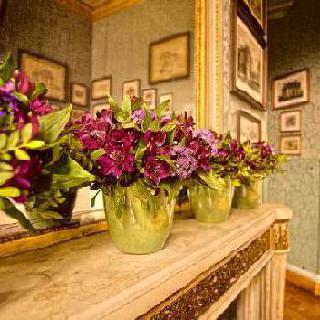}&
			\includegraphics[height=0.095\linewidth]{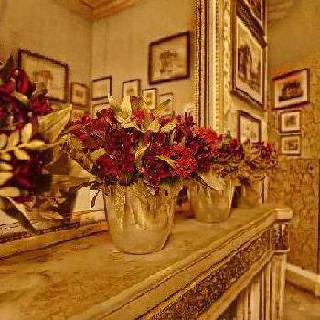}
			\\
			\includegraphics[height=0.095\linewidth]{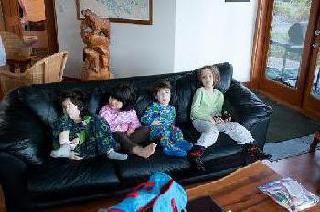}&
			\includegraphics[height=0.095\linewidth]{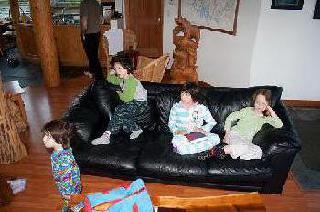}&
			\includegraphics[height=0.095\linewidth]{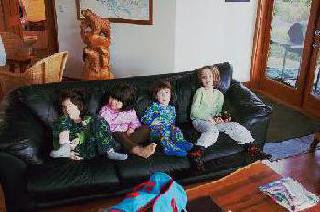}&
			\includegraphics[height=0.095\linewidth]{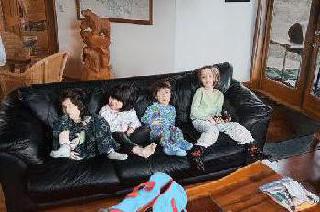}&
			\hspace{0.02in}
			\includegraphics[height=0.095\linewidth]{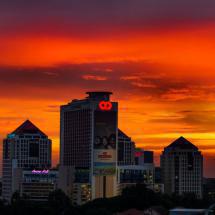}&
			\includegraphics[height=0.095\linewidth]{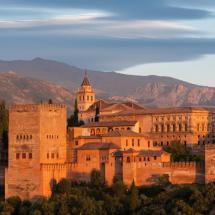}&
			\includegraphics[height=0.095\linewidth]{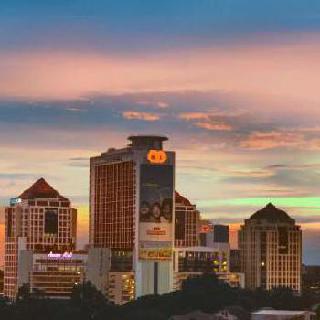}&
			\includegraphics[height=0.095\linewidth]{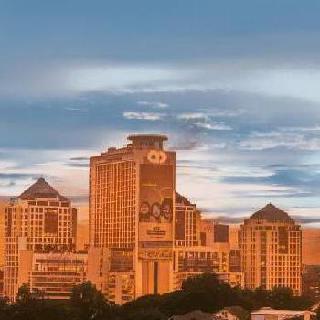}
			\\
			Source & Reference & \citet{hacohen2011non} & Ours & \hspace{0.02in} Source & Reference & Arbelot et al. & Ours\\
			&  &  &  & \hspace{0.02in} &  & [2017] & Ours
		\end{tabular}
		\caption{ Comparison with traditional local color transfer methods. \hmm{The test images} are from~\citet{hacohen2011non} (left) and~\citet{arbelot2017local} (right). Input images:~\citet{hacohen2011non} and~\citet{arbelot2017local}.}
		\label{fig:comp_local}
	\end{figure*} 
	
	\hmm{We run an ablation of the color transform by replacing it with the global color transform~\citep{reinhard2001color}, and the local color transform adopted from~\citep{reinhard2001color} which is also the initialization of our optimization, as shown in~\fref{fig:ablation_color}. The global color transform can only adjust the global tone but fails to reflect spatially varying effects. When the global method is used to match the means and variances of local patches, spatial color features are preserved but globally inconsistent artifacts appear (like the hair in the $2nd$ row and the ground in the $3rd$ row). Our color transfer enforces both local and global consistency and effectively avoids such artifacts as ghosting, halos, and inconsistencies.}
	
	\hmm{Next we show how effective our local color transfer is when combined to the region-to-region correspondences obtained by the neural style transfer~\citep{gatys2015neural} with a segmentation mask during the post-processing stage. The intermediate results with this correspondence method are shown as (a) of each example in~\fref{fig:ablation_joint}. Beside ours, there are several methods proposed to transfer colors based on these intermediate results.~\citet{luan2017deep} constrain the color transformation to be a locally affine in color space, shown as (b).~\citet{mechrez2017photorealistic} use the Screened Poisson Equation to improve photorealism, shown as (c).~\citet{liao2017image} replace only the low-frequency color bands of the source with those of the aligned reference, shown as (d). Compared to the above, our method is better at preserving fine details and image boundaries as (e) in the $1st$ example, and is more faithful to the reference with similar chrominance and contrast and does not introduce any new colors as do~\citet{mechrez2017photorealistic} ($2nd$ example). However, serving as a post-processing step, our color transfer algorithm still can not repair large correspondence errors (\eg, the black windows in the $2nd$ example and the red trees in $3rd$ example). That also shows the necessity of our joint optimization scheme.}
	
	\begin{figure*}[t]
		\footnotesize
		\setlength{\tabcolsep}{0.003\linewidth}
		\begin{tabular}{cccccc} 
			\includegraphics[width=0.16\linewidth]{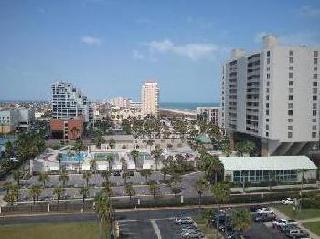}&
			\includegraphics[width=0.16\linewidth]{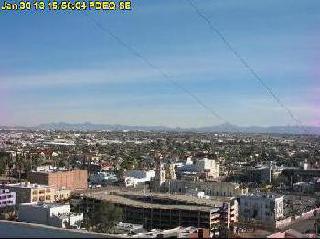}&
			\includegraphics[width=0.16\linewidth]{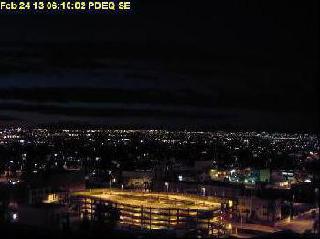}&
			\includegraphics[width=0.16\linewidth]{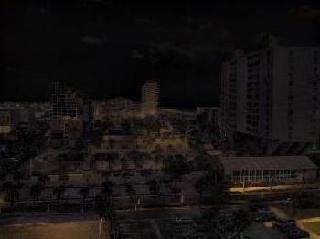}&
			\includegraphics[width=0.16\linewidth]{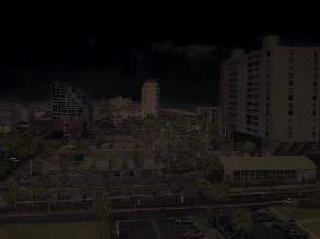}&
			\includegraphics[width=0.16\linewidth]{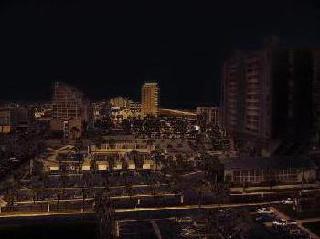}
			\\
			\includegraphics[width=0.16\linewidth]{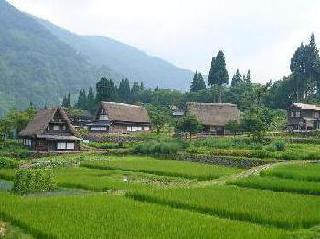}&
			\includegraphics[width=0.16\linewidth]{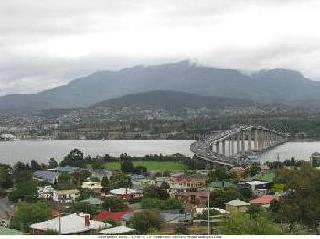}&
			\includegraphics[width=0.16\linewidth]{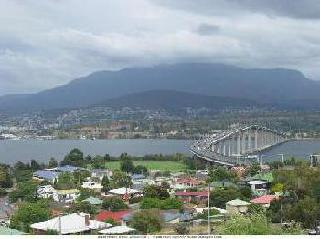}&
			\includegraphics[width=0.16\linewidth]{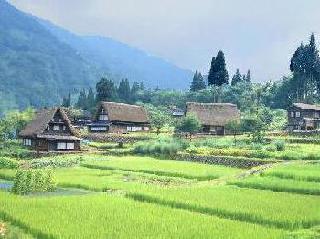}&
			\includegraphics[width=0.16\linewidth]{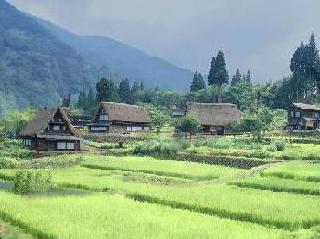}&
			\includegraphics[width=0.16\linewidth]{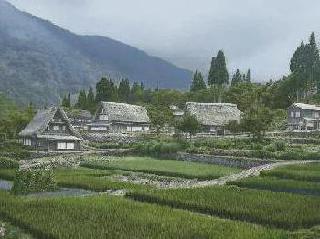}
			\\
			\includegraphics[width=0.16\linewidth]{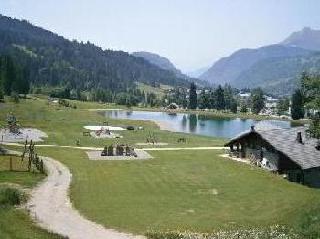}&
			\includegraphics[width=0.16\linewidth]{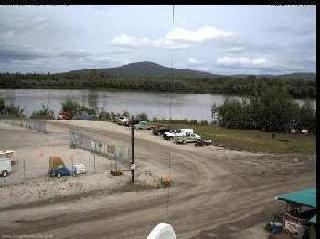}&
			\includegraphics[width=0.16\linewidth]{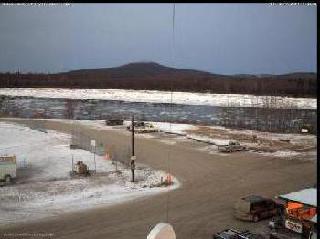}&
			\includegraphics[width=0.16\linewidth]{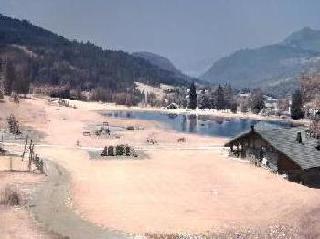}&
			\includegraphics[width=0.16\linewidth]{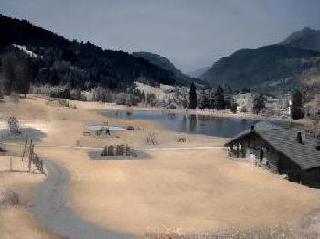}&
			\includegraphics[width=0.16\linewidth]{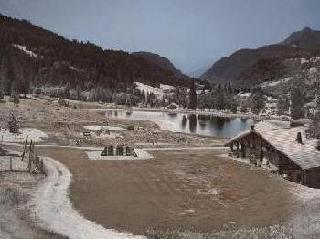}
			\\
			\includegraphics[width=0.16\linewidth]{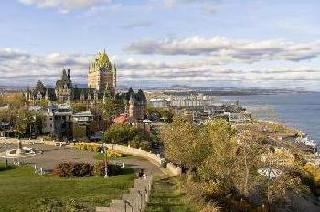}&
			\includegraphics[width=0.16\linewidth,height=0.108\linewidth]{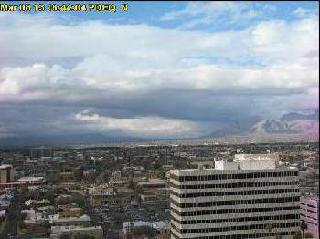}&
			\includegraphics[width=0.16\linewidth,height=0.108\linewidth]{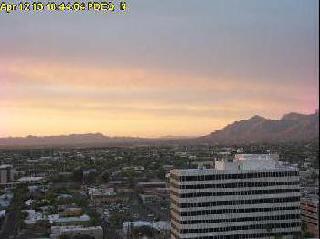}&
			\includegraphics[width=0.16\linewidth]{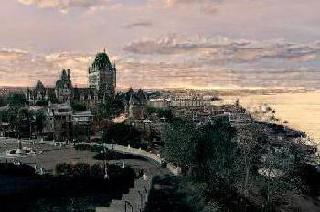}&
			\includegraphics[width=0.16\linewidth]{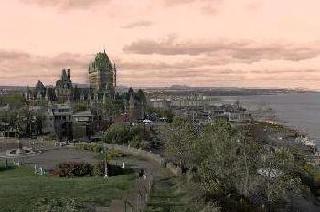}&
			\includegraphics[width=0.16\linewidth]{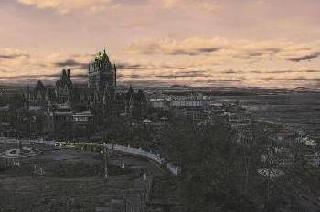}
			\\
			\includegraphics[width=0.16\linewidth]{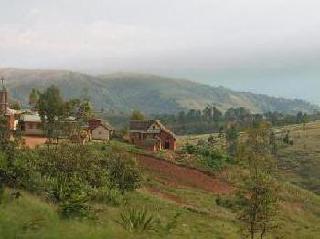}&
			\includegraphics[width=0.16\linewidth]{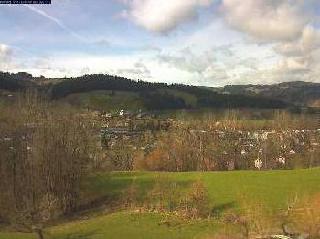}&
			\includegraphics[width=0.16\linewidth]{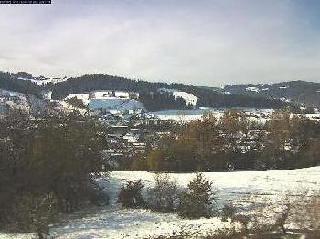}&
			\includegraphics[width=0.16\linewidth]{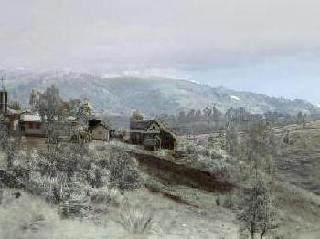}&
			\includegraphics[width=0.16\linewidth]{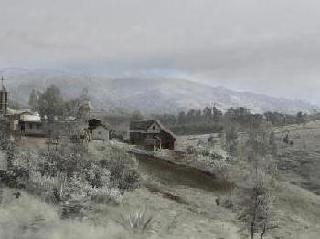}&
			\includegraphics[width=0.16\linewidth]{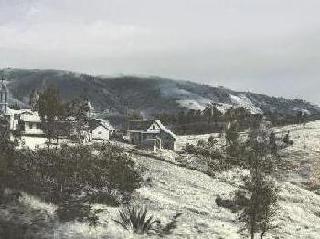}
			\\
			\includegraphics[width=0.16\linewidth]{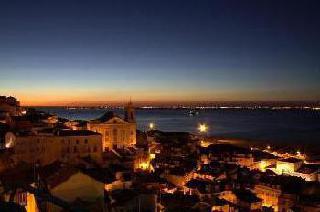}&
			\includegraphics[width=0.16\linewidth,height=0.107\linewidth]{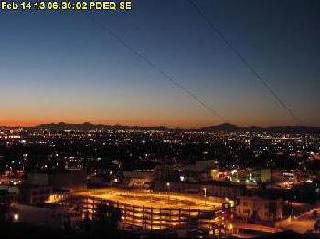}&
			\includegraphics[width=0.16\linewidth,height=0.107\linewidth]{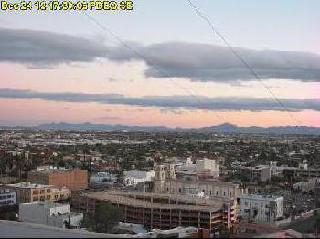}&
			\includegraphics[width=0.16\linewidth]{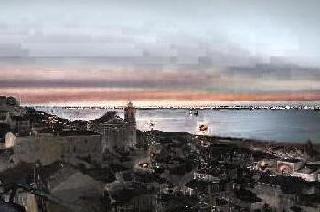}&
			\includegraphics[width=0.16\linewidth]{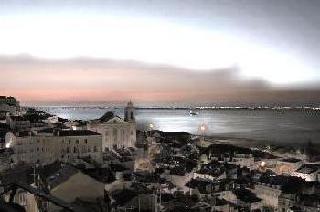}&
			\includegraphics[width=0.16\linewidth]{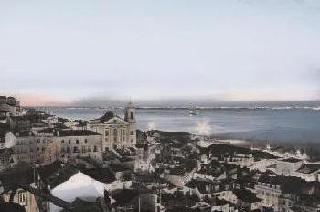}\\
			Source & Reference 1 & Reference 2& \citet{shih2013data} & \citet{laffont2014transient} & Ours
		\end{tabular}   
		\caption{Comparison to the \hmm{analogy-based} local color transfer methods on the data from~\citet{laffont2014transient}.~\citet{shih2013data} and~\citet{laffont2014transient} take both Reference 1 and Reference 2 as references while ours only takes Reference 2. \hmm{Input images:}~\citet{laffont2014transient}.}
		\label{fig:comp_set2}
	\end{figure*}  
	
	\subsection{Single-Reference Color Transfer}
	To validate our approach on color transfer, we first discuss visual comparisons with previous works in \hmm{traditional} and deep color transfer, and then report the statistics of our conducted \hmm{perceptual} study.
	
	In~\fref{fig:comp_set1}, we compare our method with the \hmm{traditional} global color transfer methods.~\citet{reinhard2001color} and~\citet{pitie2005n} only match the global color statistics between the source image and the reference image for color transfer, thus limiting their ability to conduct more sophisticated color transformations. For example, in the $2nd$ result, the house is rendered in black matching the color of the sky. In contrast, our transfer is local and capable of handling semantic object-to-object color transfer.
	
	\hmm{Next, we compare our method with the traditional color transfer methods based on local correspondence~\citep{hacohen2011non,arbelot2017local} in~\fref{fig:comp_local}. The NRDC method~\citep{hacohen2011non} is based on a small number of reliable matches to estimate the global color mapping function, so it achieves a more spatially varying result. NRDC are suitable for the image pair of the common scenes (\eg, the left half in~\fref{fig:comp_local}). In such scenes, our method builds much denser correspondence and applies local instead of global transformation, so our color transfer produces more accurate results with fewer artifacts than NRDC in local regions, like the children who are absent in the reference in the $2nd$ example and the bag in the $3rd$ example. Moreover, NRDC fails to match two different scenes, for example, there was no matching found in all the image pairs on the right half in~\fref{fig:comp_local}.~\citet{arbelot2017local} develop edge-aware descriptors to match similar textual content from different scenes but their local color transfer between similar regions is not faithful to the reference (\eg, the sky color in the $3rd$ example) and the proposed descriptors are unable to detect higher-level semantic information.}  
	
	We compare our method with the \hmm{analogy-based} local color transfer methods~\citep{shih2013data, laffont2014transient} in~\fref{fig:comp_set2}. These two algorithms depend on an additional pair of examples (\eg, Reference 1 and Reference 2) which are aligned but have different color styles for the transfer. In contrast, ours learns the color directly from the reference image (\eg, Reference 2). Therefore, our results look more faithful to the reference colors than theirs. Moreover, our method is more flexible in practice since it does not require an additional aligned pair for the transfer.
	
	\begin{figure*}[t]
		\footnotesize
		\setlength{\tabcolsep}{0.003\linewidth}
		\scalebox{0.98}
		{\begin{tabular}{cccccc}  
				\includegraphics[width=0.16\linewidth]{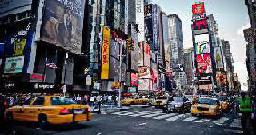}&
				\includegraphics[width=0.16\linewidth,height=0.085\linewidth]{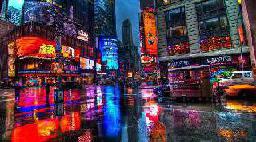}&
				\includegraphics[width=0.16\linewidth]{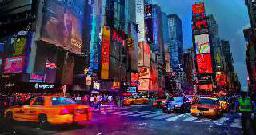}&
				\includegraphics[width=0.16\linewidth]{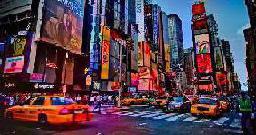}&
				\includegraphics[width=0.16\linewidth]{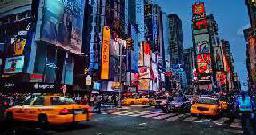}&
				\includegraphics[width=0.16\linewidth]{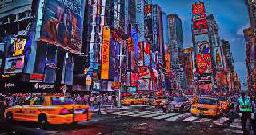}
				\\
				\includegraphics[width=0.16\linewidth]{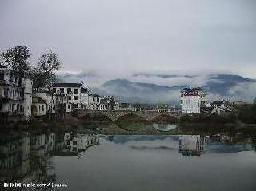}&
				\includegraphics[width=0.16\linewidth,height=0.12\linewidth]{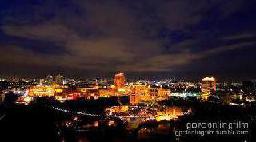}&
				\includegraphics[width=0.16\linewidth]{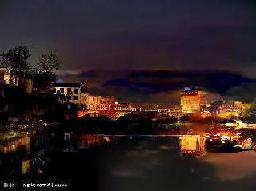}&
				\includegraphics[width=0.16\linewidth]{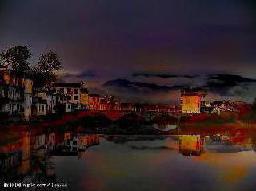}&
				\includegraphics[width=0.16\linewidth]{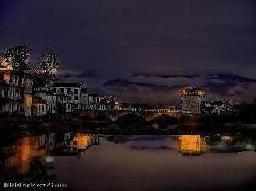}&
				\includegraphics[width=0.16\linewidth]{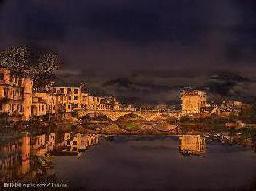}
				\\  
				\includegraphics[width=0.16\linewidth]{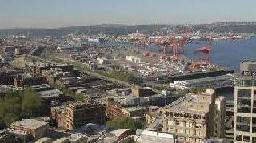}&
				\includegraphics[width=0.16\linewidth]{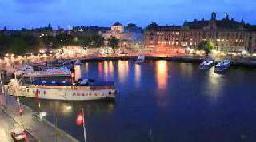}&
				\includegraphics[width=0.16\linewidth]{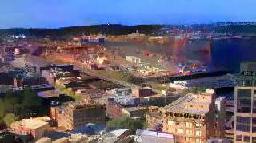}&
				\includegraphics[width=0.16\linewidth]{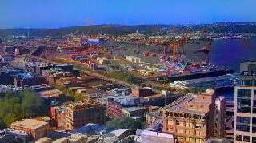}&
				\includegraphics[width=0.16\linewidth]{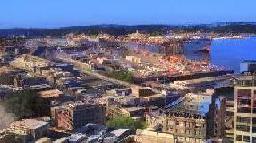}&
				\includegraphics[width=0.16\linewidth]{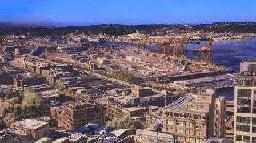}
				\\
				\includegraphics[width=0.16\linewidth]{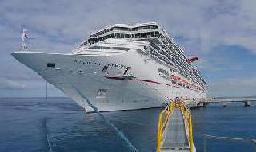}& \includegraphics[width=0.16\linewidth,height=0.097\linewidth]{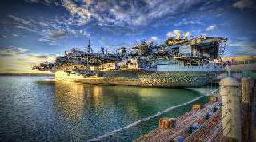}&
				\includegraphics[width=0.16\linewidth]{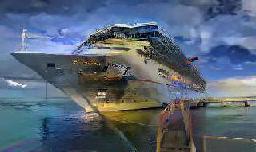}&
				\includegraphics[width=0.16\linewidth]{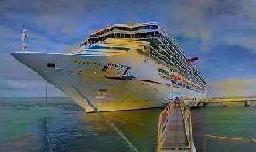}&
				\includegraphics[width=0.16\linewidth]{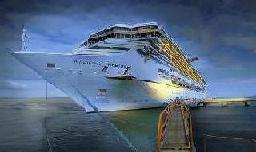}&
				\includegraphics[width=0.16\linewidth]{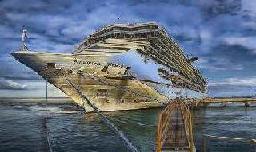}
				\\
				\includegraphics[width=0.16\linewidth]{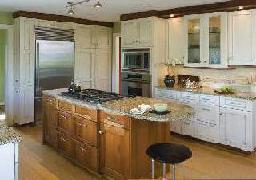}&   \includegraphics[width=0.16\linewidth,height=0.114\linewidth]{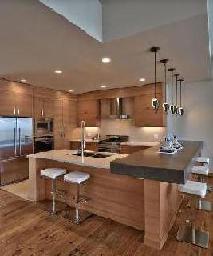}&
				\includegraphics[width=0.16\linewidth]{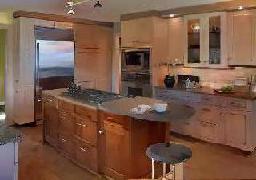}&
				\includegraphics[width=0.16\linewidth]{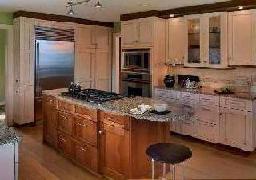}&
				\includegraphics[width=0.16\linewidth]{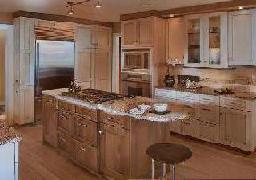}&
				\includegraphics[width=0.16\linewidth]{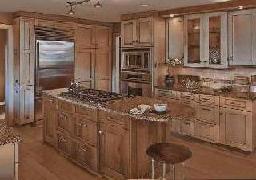}
				\\      
				\includegraphics[width=0.16\linewidth]{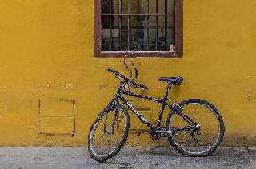}&
				\includegraphics[width=0.16\linewidth,height=0.11\linewidth]{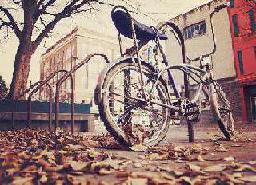}&
				\includegraphics[width=0.16\linewidth]{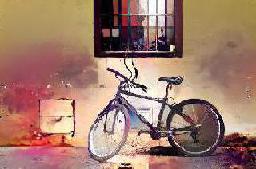}&
				\includegraphics[width=0.16\linewidth]{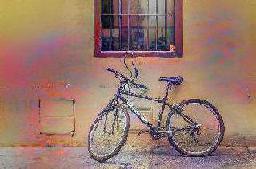}&
				\includegraphics[width=0.16\linewidth]{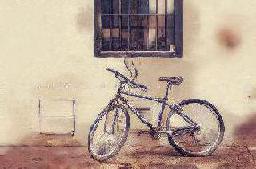}&
				\includegraphics[width=0.16\linewidth]{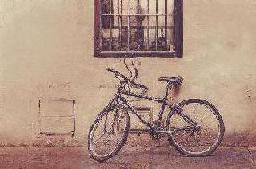}
				\\
				\includegraphics[width=0.16\linewidth]{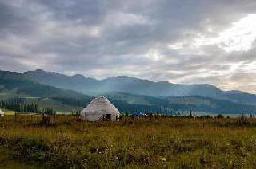}&
				\includegraphics[width=0.16\linewidth]{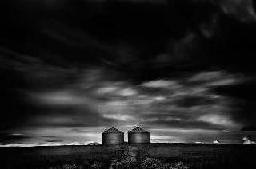}&
				\includegraphics[width=0.16\linewidth]{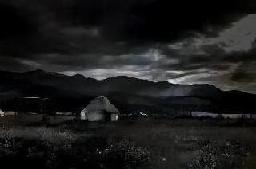}&
				\includegraphics[width=0.16\linewidth]{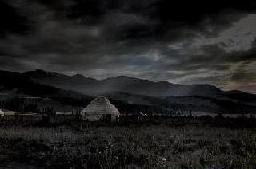}&
				\includegraphics[width=0.16\linewidth]{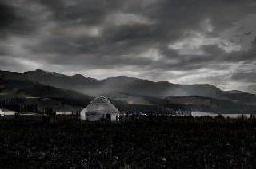}&
				\includegraphics[width=0.16\linewidth]{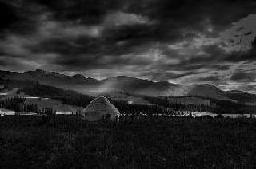}
				\\
				\includegraphics[width=0.16\linewidth]{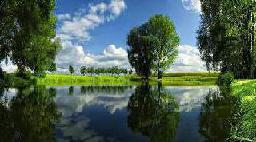}&
				\includegraphics[width=0.16\linewidth,height=0.09\linewidth]{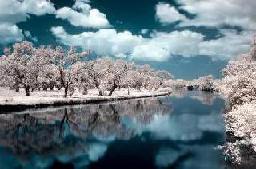}&
				\includegraphics[width=0.16\linewidth]{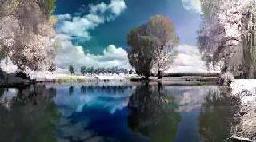}&
				\includegraphics[width=0.16\linewidth]{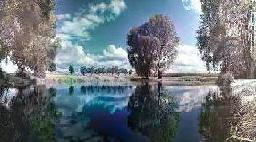}&
				\includegraphics[width=0.16\linewidth]{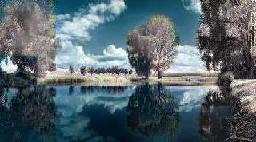}&
				\includegraphics[width=0.16\linewidth]{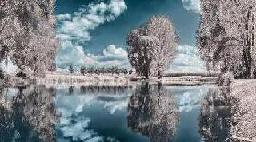}
				\\
				\includegraphics[width=0.16\linewidth]{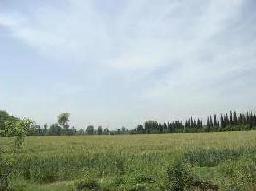}&
				\includegraphics[width=0.16\linewidth,height=0.12\linewidth]{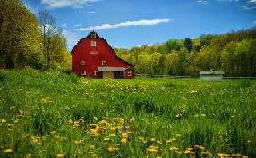}&
				\includegraphics[width=0.16\linewidth]{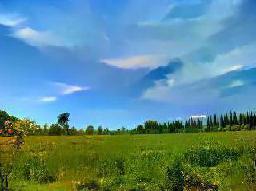}&
				\includegraphics[width=0.16\linewidth]{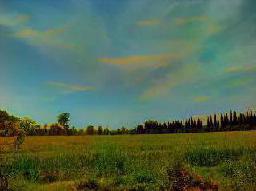}&
				\includegraphics[width=0.16\linewidth]{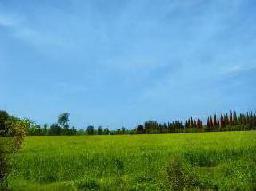}&
				\includegraphics[width=0.16\linewidth]{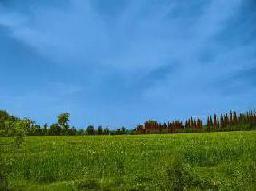}
				\\
				\includegraphics[width=0.16\linewidth]{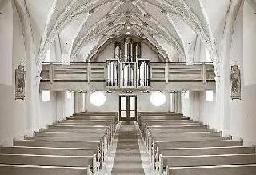}&
				\includegraphics[width=0.16\linewidth,height=0.11\linewidth]{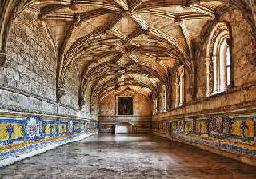}&
				\includegraphics[width=0.16\linewidth]{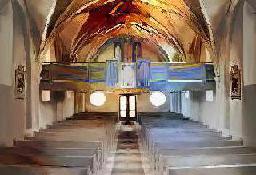}&
				\includegraphics[width=0.16\linewidth]{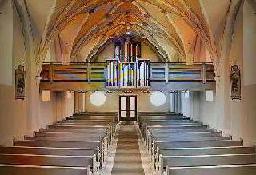}&
				\includegraphics[width=0.16\linewidth]{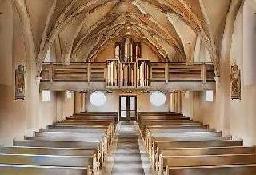}&
				\includegraphics[width=0.16\linewidth]{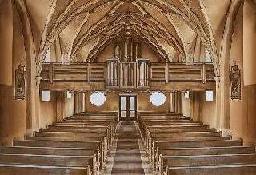}
				\\
				Source & Reference & \citet{luan2017deep} & \citet{mechrez2017photorealistic} & \citet{liao2017image} & Ours
			\end{tabular}
		}
		\caption{Comparison to recent color transfer methods based on deep features. \hmm{Input images:}~\cite{luan2017deep}.}
		\label{fig:comp_set3}
	\end{figure*}
	
	\begin{figure*}[t]
		\footnotesize
		\setlength{\tabcolsep}{0.003\linewidth}
		\begin{tabular}{cccccccc}
			&
			\includegraphics[height=0.13\linewidth]{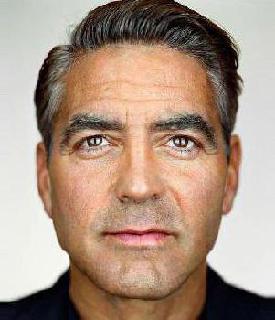}&
			\includegraphics[height=0.13\linewidth]{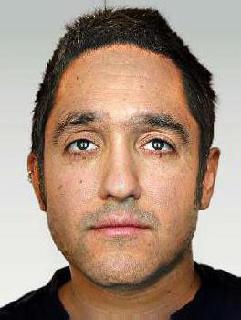}&
			\includegraphics[height=0.13\linewidth]{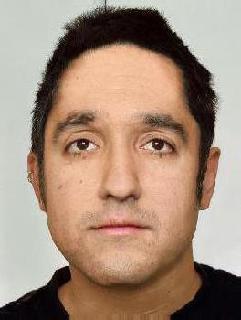}
			&      \hspace{0.02in}&
			\includegraphics[width=0.128\linewidth,height=0.13\linewidth]{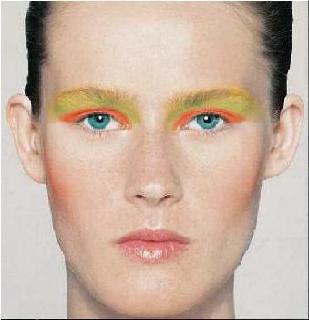}&
			\includegraphics[height=0.13\linewidth]{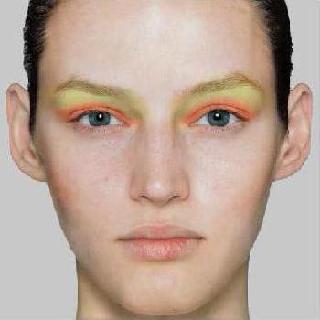}&
			\includegraphics[height=0.13\linewidth]{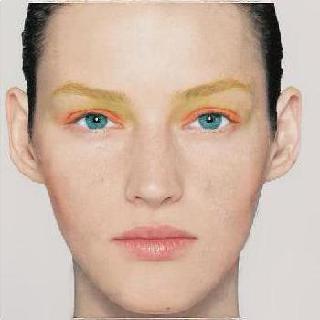}
			\\
			
			\includegraphics[height=0.13\linewidth]{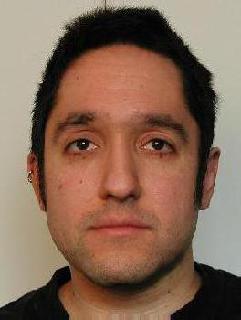}&
			\includegraphics[height=0.13\linewidth]{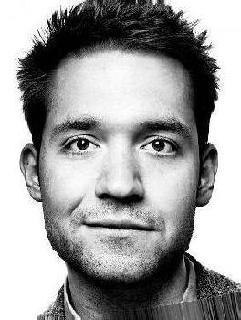}&
			\includegraphics[height=0.13\linewidth]{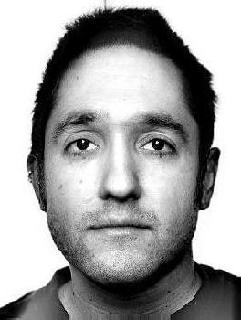}&
			\includegraphics[height=0.13\linewidth]{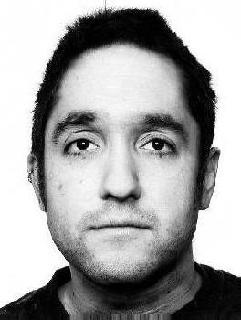}&
			\hspace{0.02in}
			\includegraphics[height=0.13\linewidth]{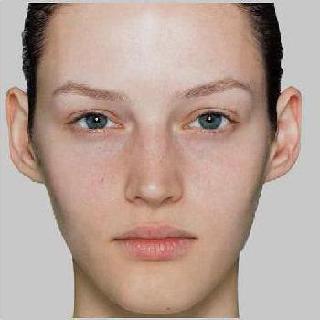}&
			\includegraphics[height=0.13\linewidth]{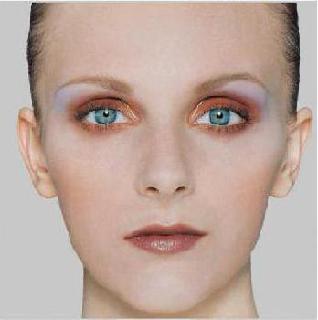}&
			\includegraphics[height=0.13\linewidth]{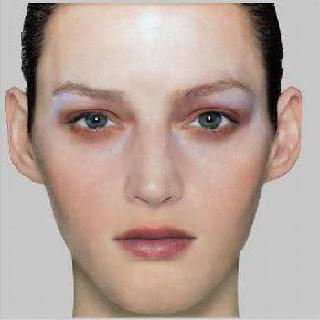}&
			\includegraphics[height=0.13\linewidth]{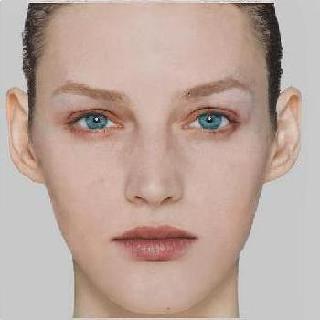}
			\\
			Source & Reference & \citet{Shih2014STH} &\hspace{0.02in} Ours & Source & Reference & \citet{tong2007example} & Ours
		\end{tabular}
		\caption{ Comparison of portrait style transfer (left) and cosmetic transfer (right). \hmm{Input images:}~\citet{Shih2014STH} and~\citet{tong2007example}.} 
		\label{fig:comp_portrait}
	\end{figure*}
	
	\hmm{In~\fref{fig:comp_set3}, we compare with three recent color transfer methods based on CNN features~\citep{luan2017deep,mechrez2017photorealistic,liao2017image}. The methods of~\citet{luan2017deep} and~\citet{mechrez2017photorealistic} match the global statistics of deep features (\ie, Gram Matrix) and guarantee region-to-region transfer via segmentation masks. One type of noticeable artifact in the results of~\citet{luan2017deep} is posterization, which is visible in the bicycle in the $6th$ row and the cloud in the $9th$ row of~\fref{fig:comp_set3}.~\citet{mechrez2017photorealistic} post-process the stylized image based on the Screened Poisson Equation to constrain the gradients to those of the original source image. They make the stylized results more photorealistic but may generate unnatural colors (\eg, the yellow sky in the $4th$ and $9th$ rows). Independent of segmentation masks,~\citet{liao2017image} find dense correspondence between two images using deep features, yielding the results that are more similar to ours. However, unlike their approach which separates correspondence estimation and color transfer, our method performs joint optimization, which can align two images better for the color transfer and generate results with fewer ghosting artifacts. This is clearly shown from the buildings in the $2nd$ and $3rd$ rows. Moreover, their color transfer only replaces low-frequency color bands; while our local color transfer considers all-frequency color bands. Thus, ours can generate results more faithful to the reference. For example, in the $8th$ row, the original green color can be observed in the result while ours is better at preserving the contrast and the chrominance of the reference.}
	
	Moreover, our method is effective in transferring effects like makeup or photographer styles from one portrait to another. Compared to the methods  specifically focusing on portraits and very specific kinds of effects, ours can generate comparable results as shown in~\fref{fig:comp_portrait}, but without requiring extra inputs of the pair before and after makeup (\citet{tong2007example}), face landmarks (\citet{Shih2014STH} and~\citet{tong2007example}) or matting (\citet{Shih2014STH}).
	
	\begin{figure}[h]
		\footnotesize
		\setlength{\tabcolsep}{0.003\linewidth}
		\begin{tabular}{cc} 
			\includegraphics[width=0.49\linewidth]{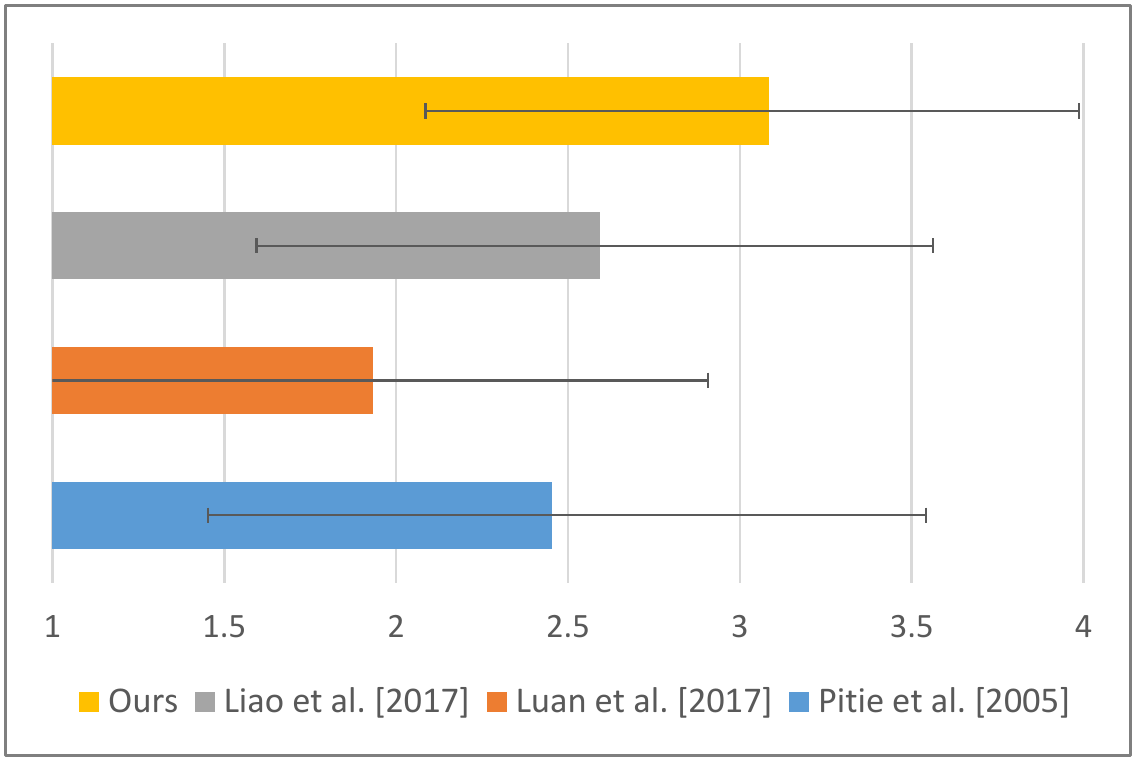} &
			\includegraphics[width=0.49\linewidth]{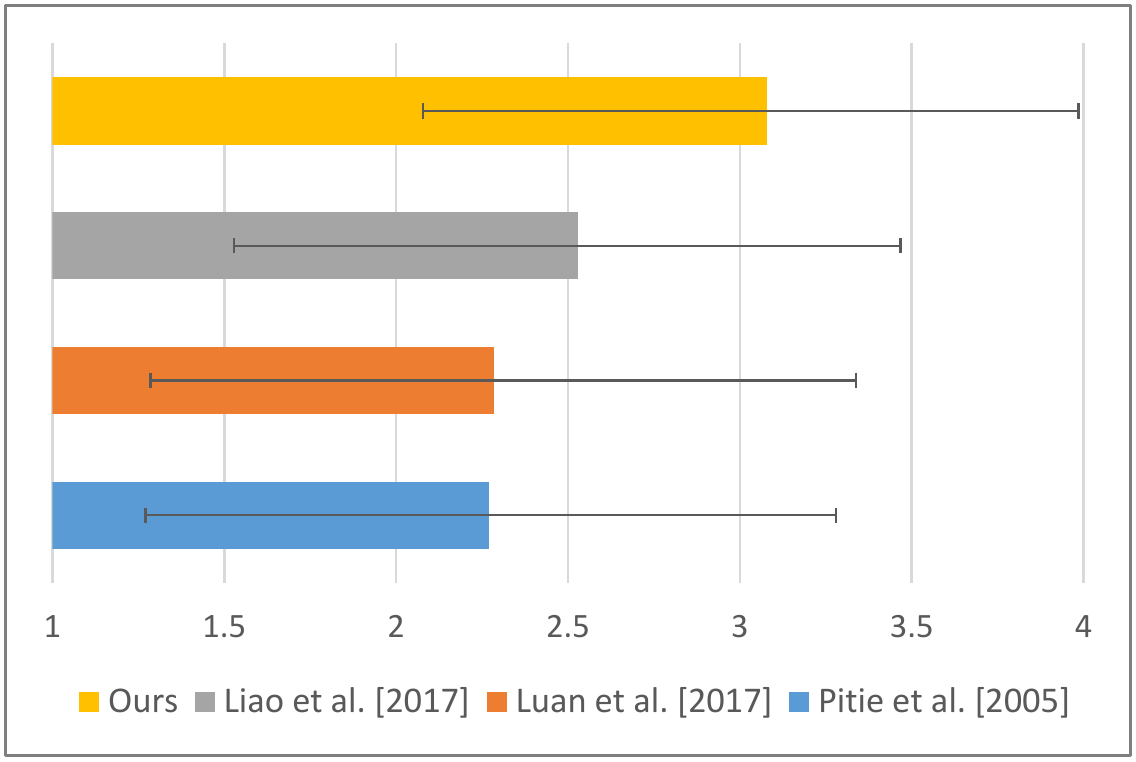}
			\\
			(a) Photorealistic (avg score and std) & (b) Faithful to style (avg score and std)  
			\\
			\\
			\includegraphics[width=0.485\linewidth]{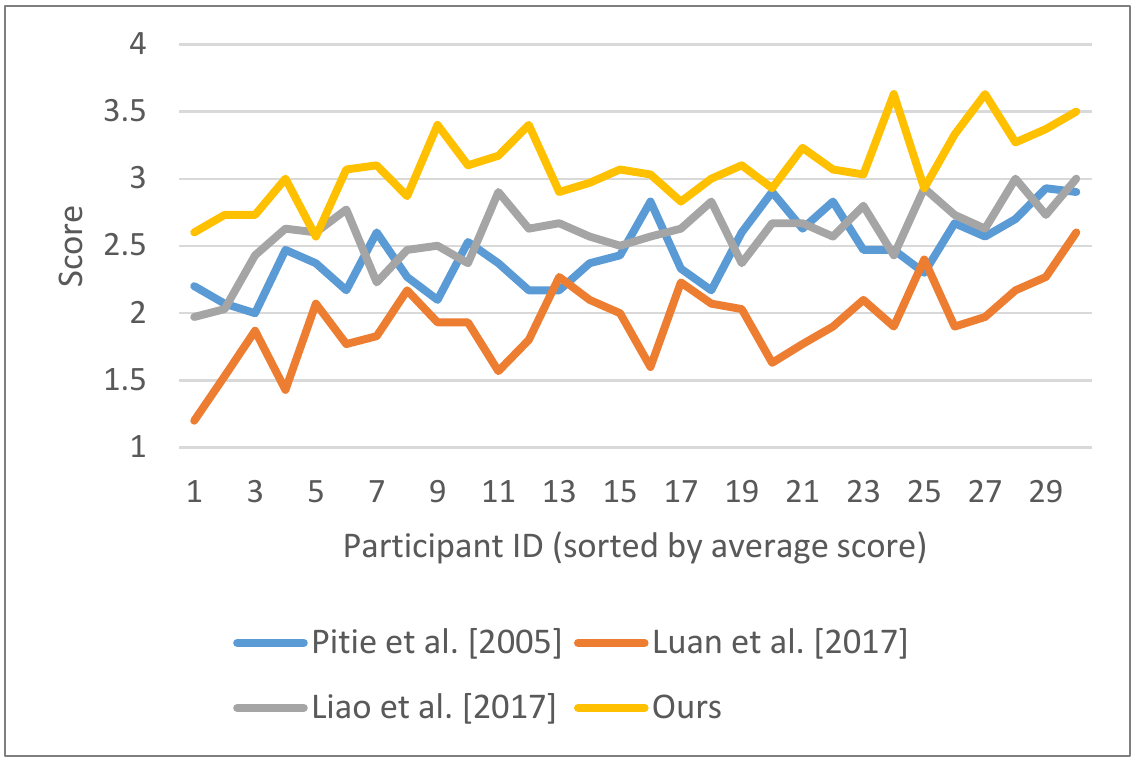} &
			\includegraphics[width=0.49\linewidth]{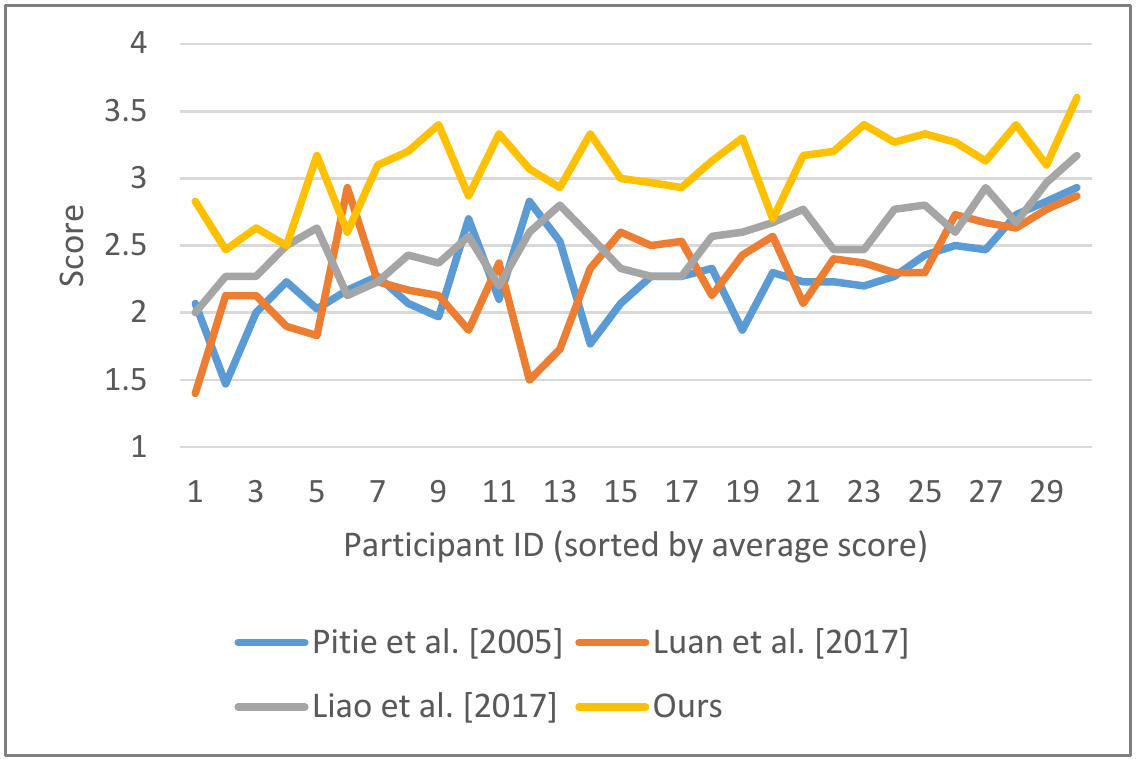}
			\\
			(c) Photorealistic (avg score per user) & (d) Faithful to style (avg score per user)  
			\\
			\\
			\includegraphics[width=0.49\linewidth]{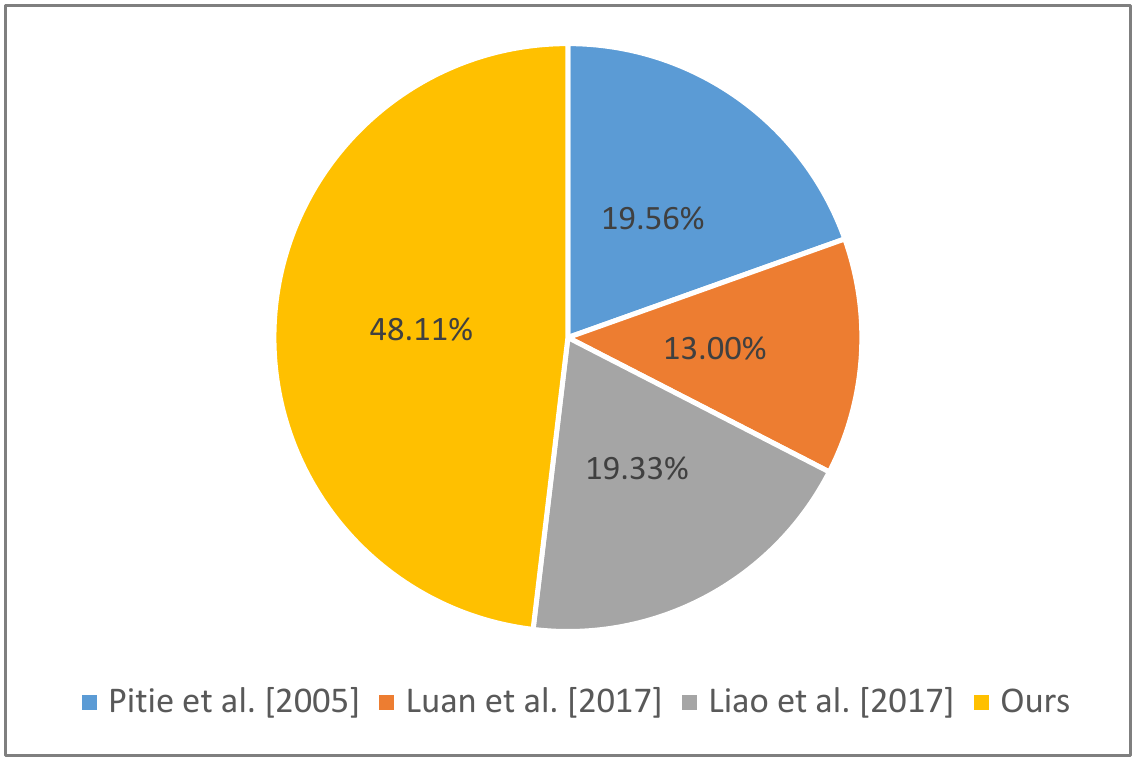} &
			\includegraphics[width=0.49\linewidth]{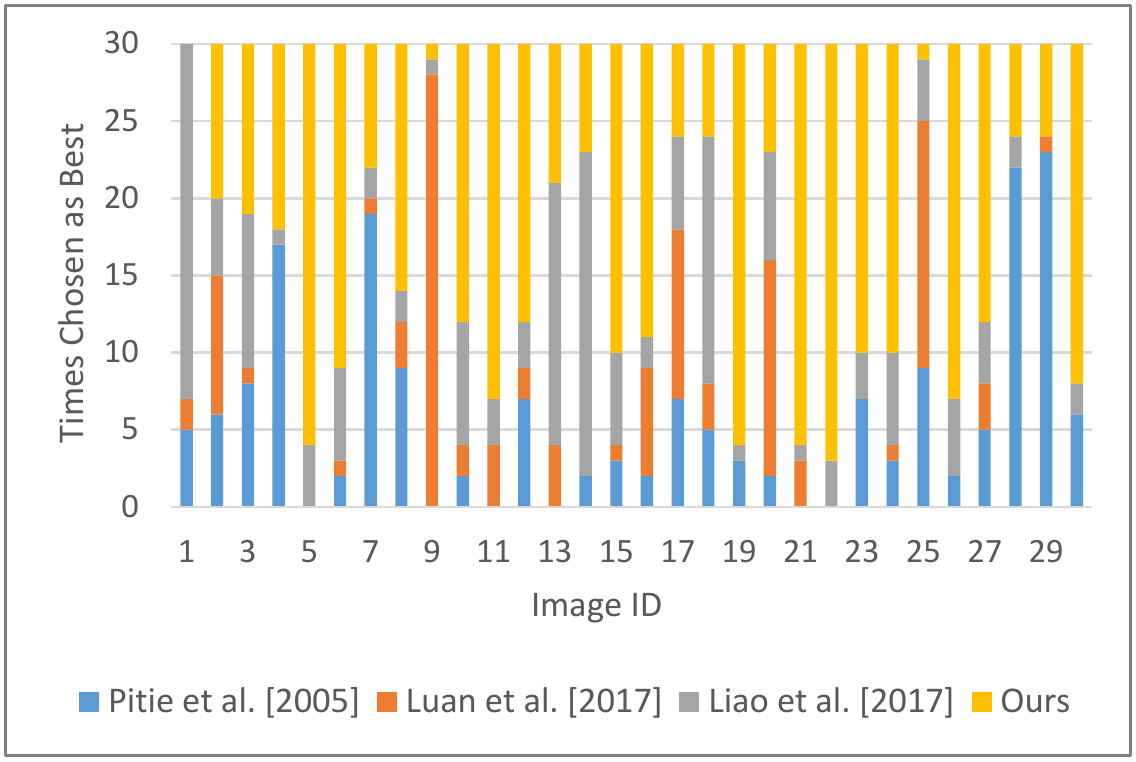}
			\\
			(e) Choice of the best (percentage) & (f) Choice of the best (times per scene)
		\end{tabular}
		\caption{Perceptual study results. \hmm{(a) and (b) demonstrate the average scores (as data bars) and the standard deviations (as error bars) of the four methods:~\citet{pitie2005n},~\citet{luan2017deep},~\citet{liao2017image} and ours in photorealism and faithfulness. (c) and (d) show the average scores given by each participant in the two aspects. (e) and (f) illustrates the percentage and the times of each method voted as the best in total and in every example based on photorealism and faithfulness.}}
		\label{fig:user}
	\end{figure}
	
	\begin{table}
		\centering
		\caption{Runtime of different color transfer methods.}
		\footnotesize
		\smallskip
		\label{tab:runtime}
		\begin{tabular}{lcccc}
			\toprule
			Method & \citet{pitie2005n} & \citet{liao2017image} & \citet{luan2017deep} & Ours\\
			\toprule
			Runtime (sec) & 7 & 300 & 600 & 60 \\
			\toprule
		\end{tabular}
	\end{table}
	
	\paragraph{\textbf{\hmm{Perceptual} Study.}} We conduct a \hmm{perceptual} study to evaluate our color transfer work in terms of photorealism and faithfulness to reference style. We compared the following techniques:~\citet{pitie2005n},~\citet{luan2017deep},~\citet{liao2017image} and ours in the study. We present the results of the four methods to participants in a random order and ask them to score images in a 1-to-4 scale from \enquote{definitely not photorealistic} to \enquote{definitely photorealistic} for question 1, from \enquote{definitely not faithful to the reference} to \enquote{definitely faithful to the reference} for question 2, and select the best one considering both criteria for question 3. \hmm{The metric \enquote{photorealism} is defined as no ghosting, no halo, and no unnatural colors, while the metric \enquote{faithfulness} is \hmm{defined} to measure the similarity in chrominance, luminance and contrast between semantically corresponding regions in the reference and the result.} We use \hmm{30} different \hmm{scenes} for each of the four methods and collect the responses from \hmm{30 participants}. The examples are randomly selected from the test images of~\citet{luan2017deep,hacohen2011non,laffont2014transient}.
	
	\begin{figure*}[t]
		\footnotesize
		\setlength{\tabcolsep}{0.003\linewidth}
		\includegraphics[width=0.98\linewidth]{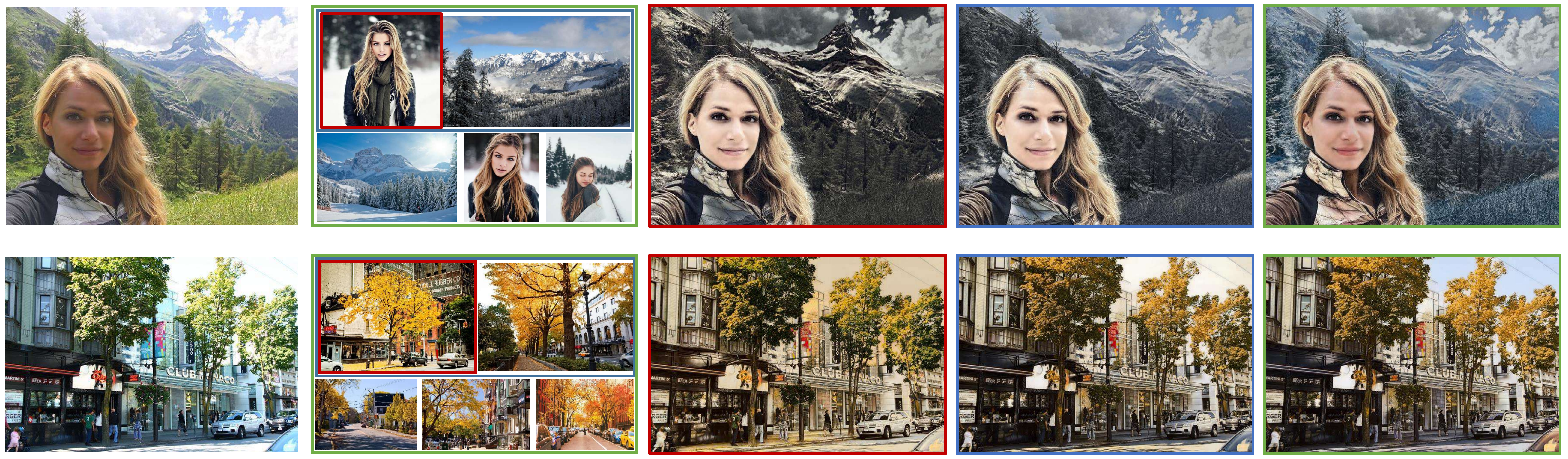}\\
		\begin{tabular}{ccccc}  
			\qquad \qquad Source \qquad \quad & \qquad \qquad \qquad \qquad References \qquad \qquad \qquad & \qquad \quad Result (single reference) \qquad \quad & \qquad Result (two references) \qquad & \quad \qquad Result (five references) \quad \\
		\end{tabular}   
		\caption{Comparison of our method with single and multiple references. \hmm{Please note that each result is generated respectively using the references in the same border color as it.} Input source images: and Anonymous/Flickr.com.}
		\label{fig:comp_set41}
	\end{figure*}

	\hmm{\fref{fig:user}(a)(b) demonstrate the average scores and standard deviations of each method. For photorealism and faithfulness, ours and~\citet{liao2017image} are ranked 1st ($3.09\pm0.90$ and $3.08\pm0.91$) and 2nd ($2.59\pm0.97$ and $2.54\pm0.93$) respectively, followed by~\citet{pitie2005n} ($2.45\pm1.09$ and $2.27\pm1.01$) and~\citet{luan2017deep} ($1.93\pm0.97$ and $2.28\pm1.05$). The method by~\citet{luan2017deep} performs worst in photorealism, since it often produces posterization (cartoon-like) effects and introduces unnatural colors; while~\citet{pitie2005n} perform the worst in faithfulness to style, since global transfer limits the spatial variety of styles.~\fref{fig:user}(c)(d) show the average scores given by every participant. Ours is consistently better than others in both photorealism and faithfulness among all participants. We further conduct repeated-measures ANOVAs on the collected data, and it shows the differences between the four methods from these two aspects are all significant ($p < 0.005$). We also use simple contrasts to compare each method against our method among all 900 scores in both photorealism and faithfulness (30 participants by 30 scenes). For photorealism, participants prefer ours over~\citet{pitie2005n} ($56.22\%$ better, $14.89\%$ equal),~\citet{luan2017deep} ($75.89\%$ better, $10.78\%$ equal) and~\citet{liao2017image} ($56.22\%$ better, $19.00\%$ equal). For faithfulness, \hmm{participants} prefer ours over~\citet{pitie2005n} ($62.11\%$ better, $14.11\%$ equal),~\citet{luan2017deep} ($67.11\%$ better, $9.11\%$ equal) and~\citet{liao2017image} ($58.33\%$ better, $19.00\%$ equal).}
	
	Since color transfer quality depends on both photorealism and style fidelity, we examine the subject's preferred result considering both criteria. The pie chart on~\fref{fig:user}(e) shows the percentage of each method selected as the best. Our algorithm is the top overall selection over the other three methods at \hmm{$48.11\%$} of the time.~\fref{fig:user} (f) gives the detailed numbers of how many times each method is selected as the best in each scene. \hmm{It shows that more users prefer ours over~\citet{pitie2005n} ($23$ vs. $7$),~\citet{luan2017deep} ($25$ vs. $5$), and~\citet{liao2017image} ($22$ vs. $5$ with $3$ equal).} 
	\hmm{Moreover, our method is more efficient than the other two deep network-based methods. The runtime of the four competing algorithms is shown in~\Tref{tab:runtime}. }
	
	\subsection{Multi-Reference Color Transfer}   
\begin{figure*}[t]
		\footnotesize
		\setlength{\tabcolsep}{0.003\linewidth}
		\begin{tabular}{ccccc}        \includegraphics[width=0.24\linewidth]{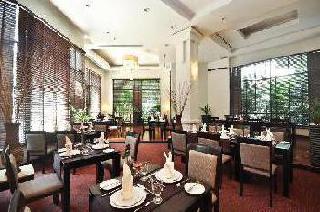}&      \includegraphics[width=0.24\linewidth]{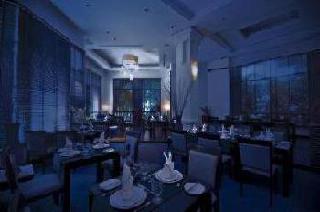}&     \includegraphics[width=0.23\linewidth, height=0.161\linewidth]{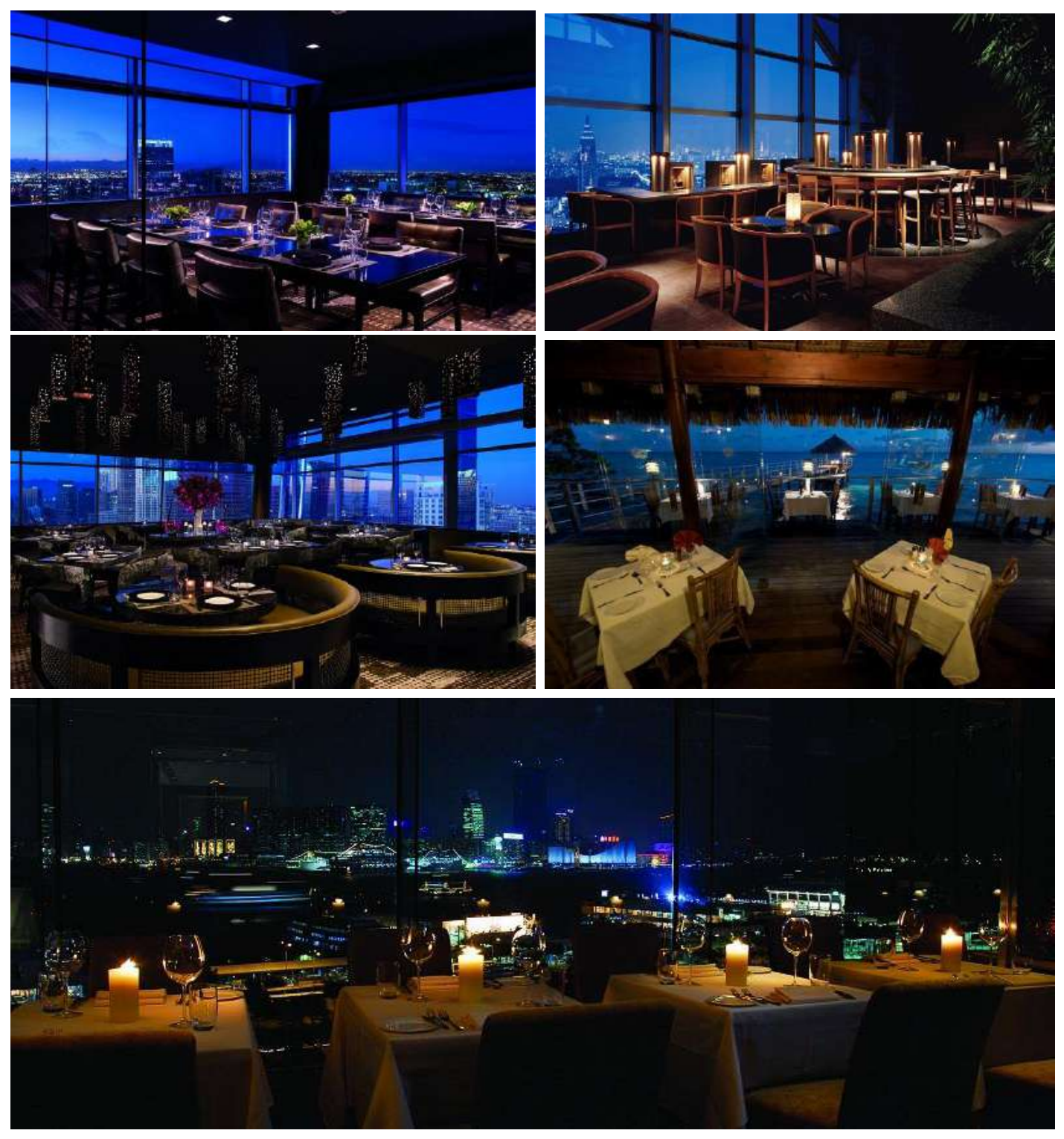}&
			\includegraphics[width=0.24\linewidth]{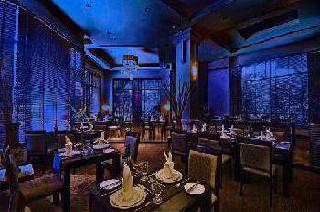}
			\\
			Source& \citet{liu2014autostyle} (\enquote{night}) & Refs (\enquote{restaurant night}) & Ours    
			\\
			\\      \includegraphics[width=0.24\linewidth]{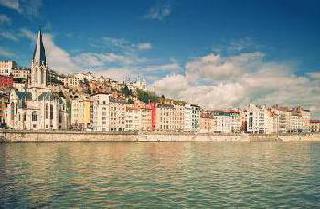}&      \includegraphics[width=0.24\linewidth]{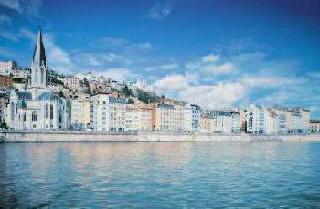}&
			\includegraphics[width=0.23\linewidth,height=0.158\linewidth]{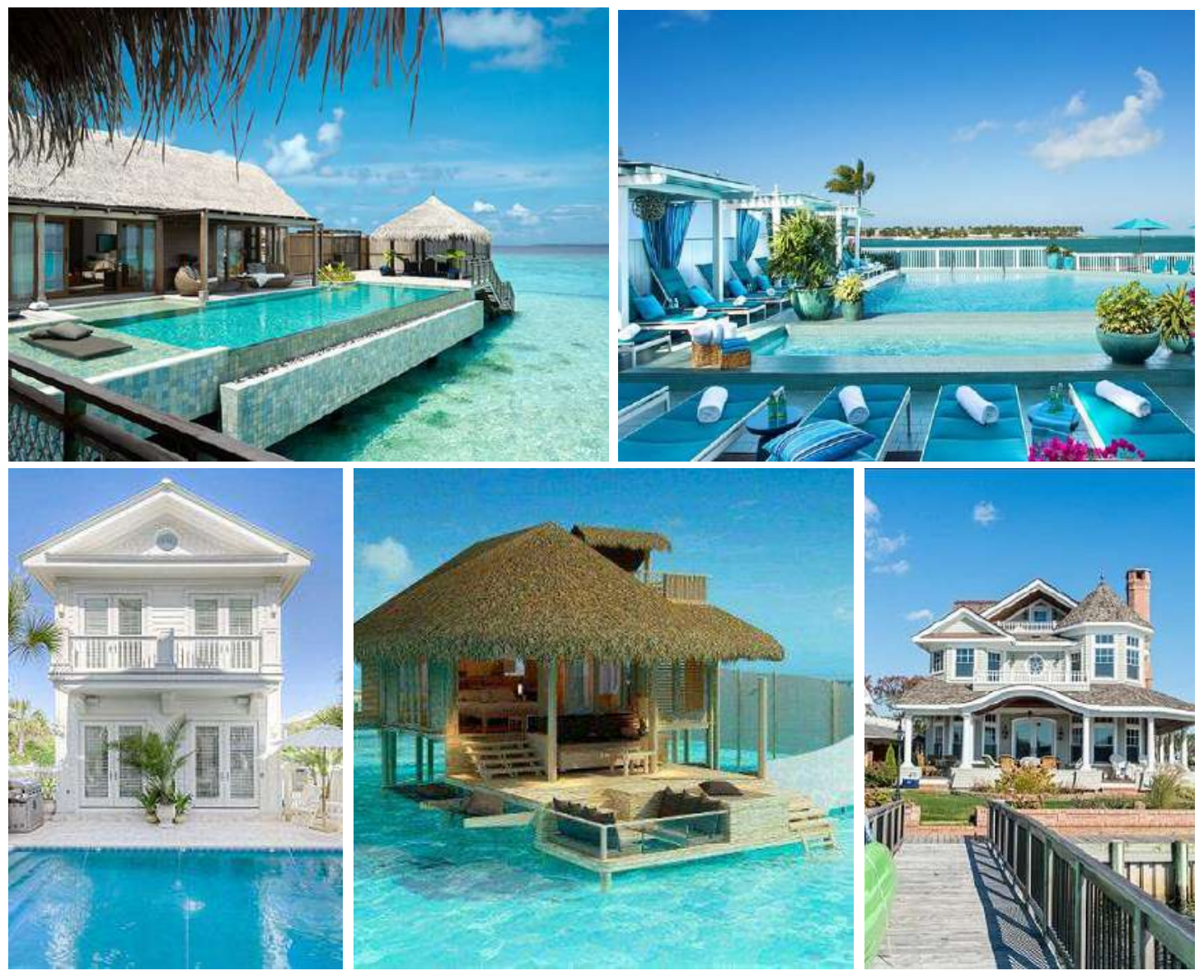}&
			\includegraphics[width=0.24\linewidth]{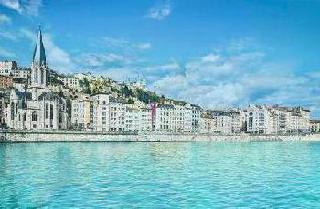}
			\\
			Source& \citet{liu2014autostyle} (\enquote{beach}) & Refs (\enquote{building beach}) & Ours
			\\ 
			\\      \includegraphics[width=0.24\linewidth]{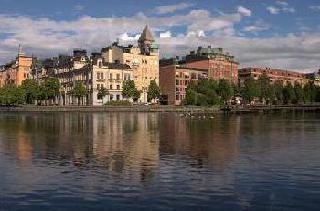}&     \includegraphics[width=0.24\linewidth]{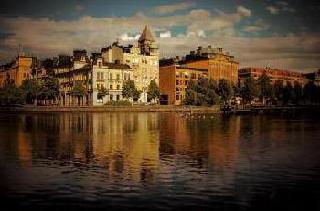}&    \includegraphics[width=0.23\linewidth,height=0.16\linewidth]{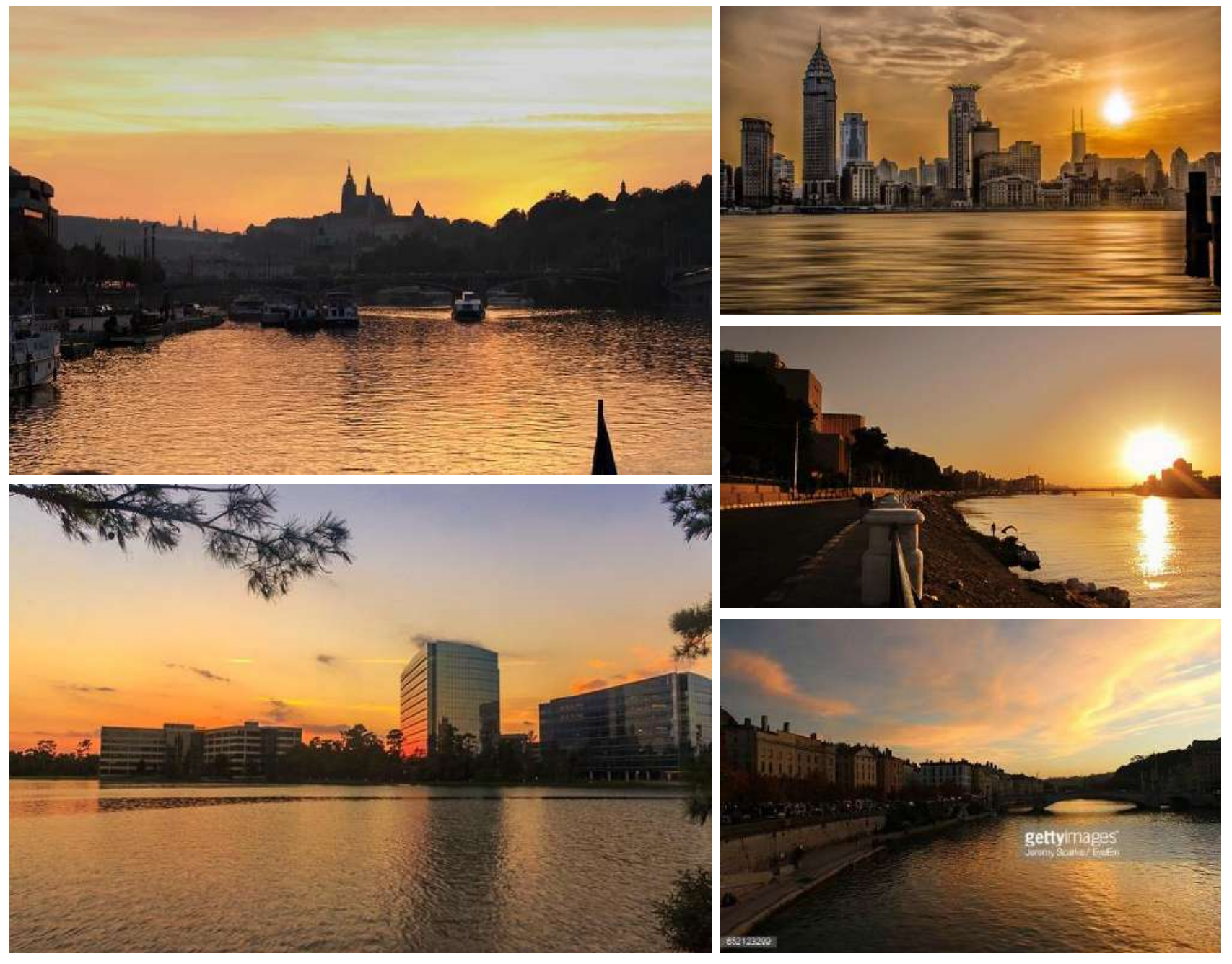}&      \includegraphics[width=0.24\linewidth]{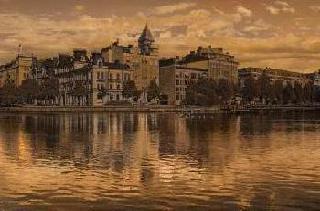}
			\\
			Source& \citet{liu2014autostyle} (\enquote{sunset}) & Refs (\enquote{building river sunset}) & Ours
			\\ 
			\\
			\\
			\\      \includegraphics[width=0.24\linewidth]{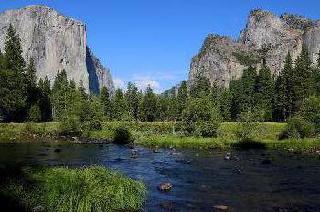}&      \includegraphics[width=0.24\linewidth]{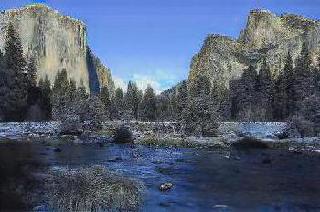}&      \includegraphics[width=0.23\linewidth,height=0.162\linewidth]{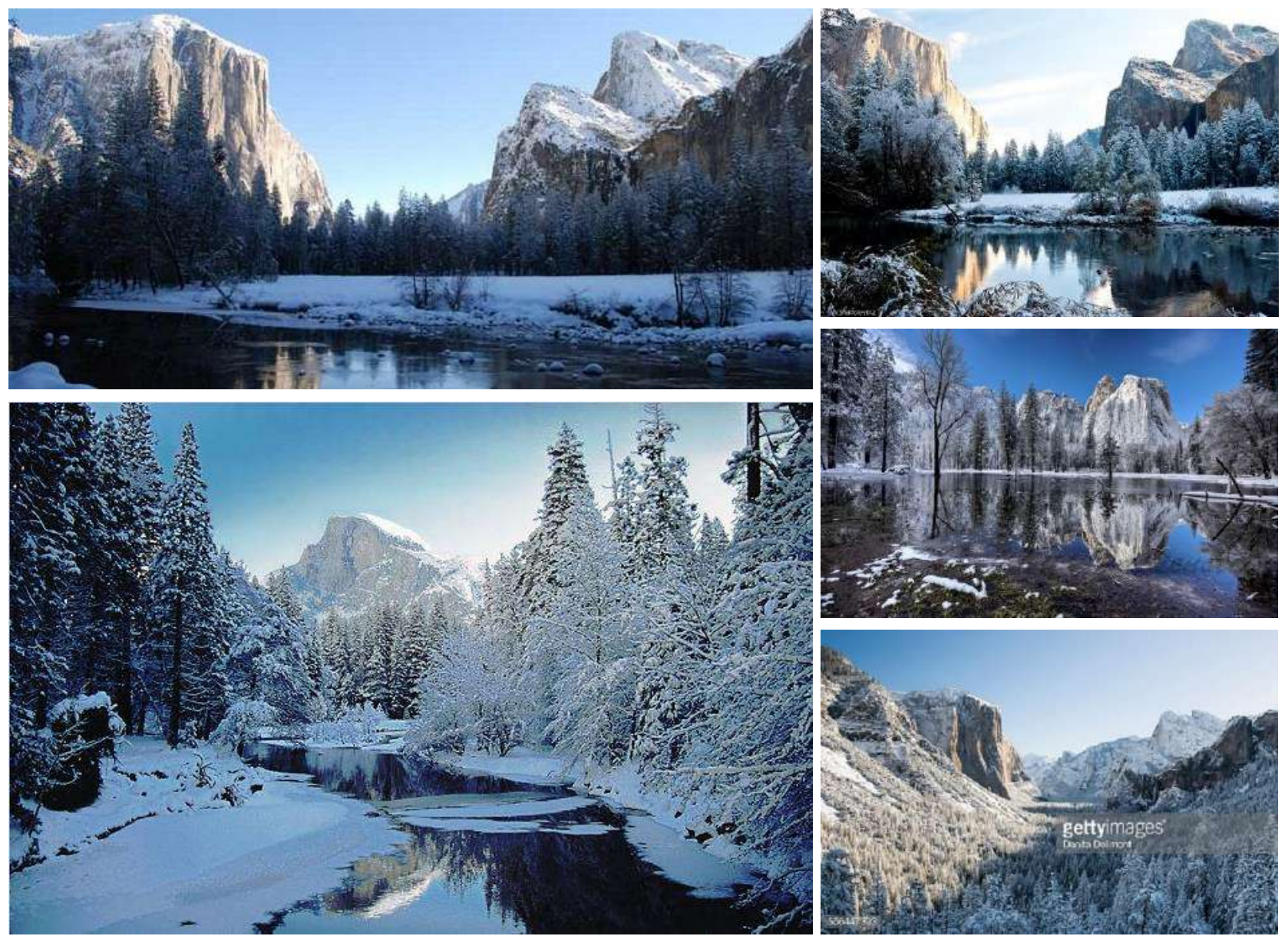}&     \includegraphics[width=0.24\linewidth]{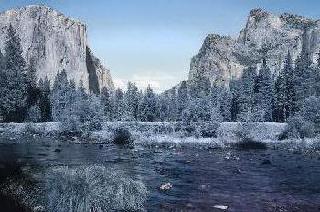}
			\\
			Source& \citet{zhu2017unpaired} (\enquote{winter Yosemite}) & Refs (\enquote{Yosemite winter}) &Ours 
			\\      \includegraphics[width=0.24\linewidth]{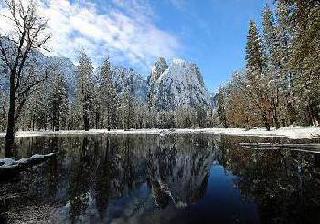}&
			\includegraphics[width=0.24\linewidth]{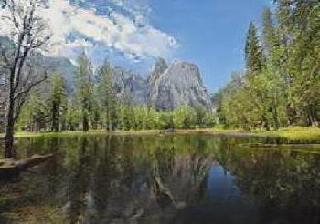}&      \includegraphics[width=0.23\linewidth,height=0.173\linewidth]{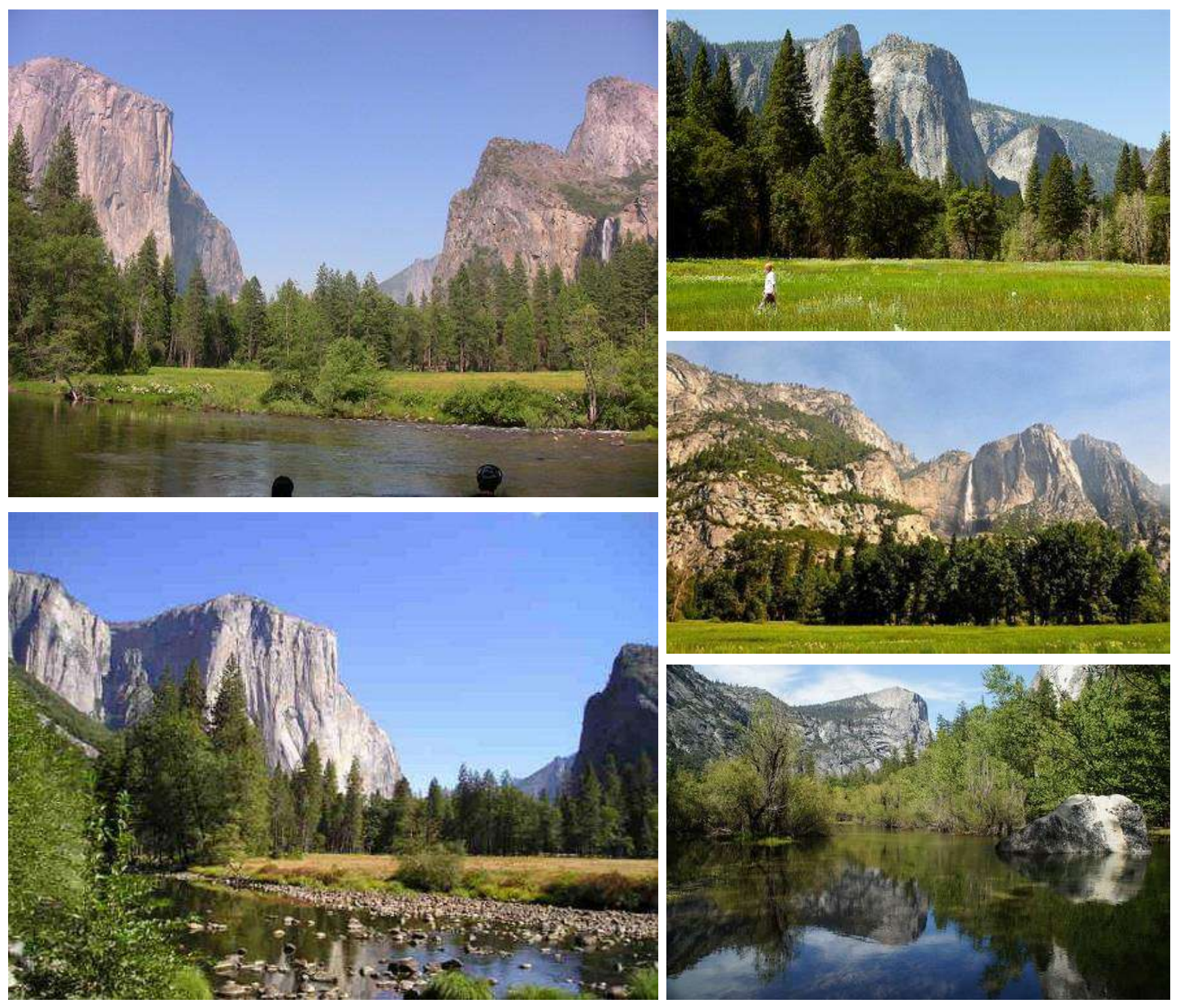}&
			\includegraphics[width=0.24\linewidth]{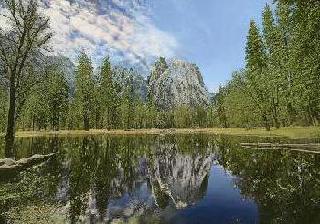}
			\\
			Source& \citet{zhu2017unpaired} (\enquote{summer Yosemite}) & Refs (\enquote{Yosemite summer})  &Our      
		\end{tabular}
		\caption{Comparison with multiple color transfer methods on their source images and our own references automatically retrieved from the Internet based on the keywords. Input images:~\citet{liu2014autostyle} (top three) and~\citet{zhu2017unpaired} (bottom two).}
		\label{fig:comp_set4}
	\end{figure*}

	In all single-reference color transfer methods, reference selection is crucial to achieving satisfactory results. Our method fails to transfer correct colors and yields an unnatural appearance in the regions where no proper color guidance can be found in the reference. This problem can be addressed by introducing multiple references.  For example, in the bottom row of~\fref{fig:comp_set41}, the sky color is incorrect because the single reference does not contain the sky, but it is correct with multiple references, some of which contains the sky. Our multi-reference approach allows the user to provide keywords for controlling the color transfer, for example, \enquote{restaurant night} in the first row of~\ref{fig:comp_set4}. \enquote{Restaurant} describes the content in the source images, and \enquote{night} defines the desired style. To automatically obtain multiple references for the transfer, these keywords are used to retrieve images through a search engine (\eg, Google Images). We collect the top 50 search results as candidates. Naturally, these images may have a wide variety of color styles as well as outliers. To select a set of images with consistent colors, for each candidate image, we compute its normalized histogram in the HSV color space and select the most representative one with a minimum $L_2$ histogram distance to all others. Then we choose the closest subset of $5$ images to the representative one as the references for transfer. 
	
	Compared with Autostyle~\citep{liu2014autostyle} which uses a global method to transfer vignetting, color, and local contrast from a collection of images searched by a particular keyword for one specific style (\eg, \enquote{night}, \enquote{beach}, \enquote{sunset}), our approach performs a local transfer based on semantic correspondence and thus requires more keywords in order to describe both content and style (\eg, \enquote{restaurant night}, \enquote{building beach}, \enquote{building river sunset}). However, it allows us to produce a more precise color transfer than Autostyle~\citep{liu2014autostyle}. Cycle-GAN~\citep{zhu2017unpaired} also allows the transfer of a specific style (\eg, \enquote{winter Yosemite}) to a given source without selecting the reference by leveraging network training on large datasets. Compared to theirs, our method is more flexible when testing different styles without retraining. Our results have fewer checkerboard artifacts, as shown on the bottom two rows of~\fref{fig:comp_set4}.
	
	\subsection{Colorization}
\begin{figure*}[t]
		\footnotesize
		\setlength{\tabcolsep}{0.001
			\linewidth}
		\begin{tabular}{cccccc}       \includegraphics[height=0.1555\linewidth]{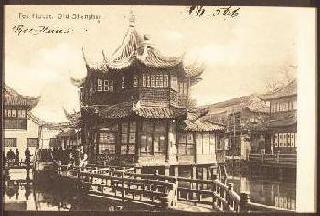}&
			\includegraphics[height=0.1555\linewidth]{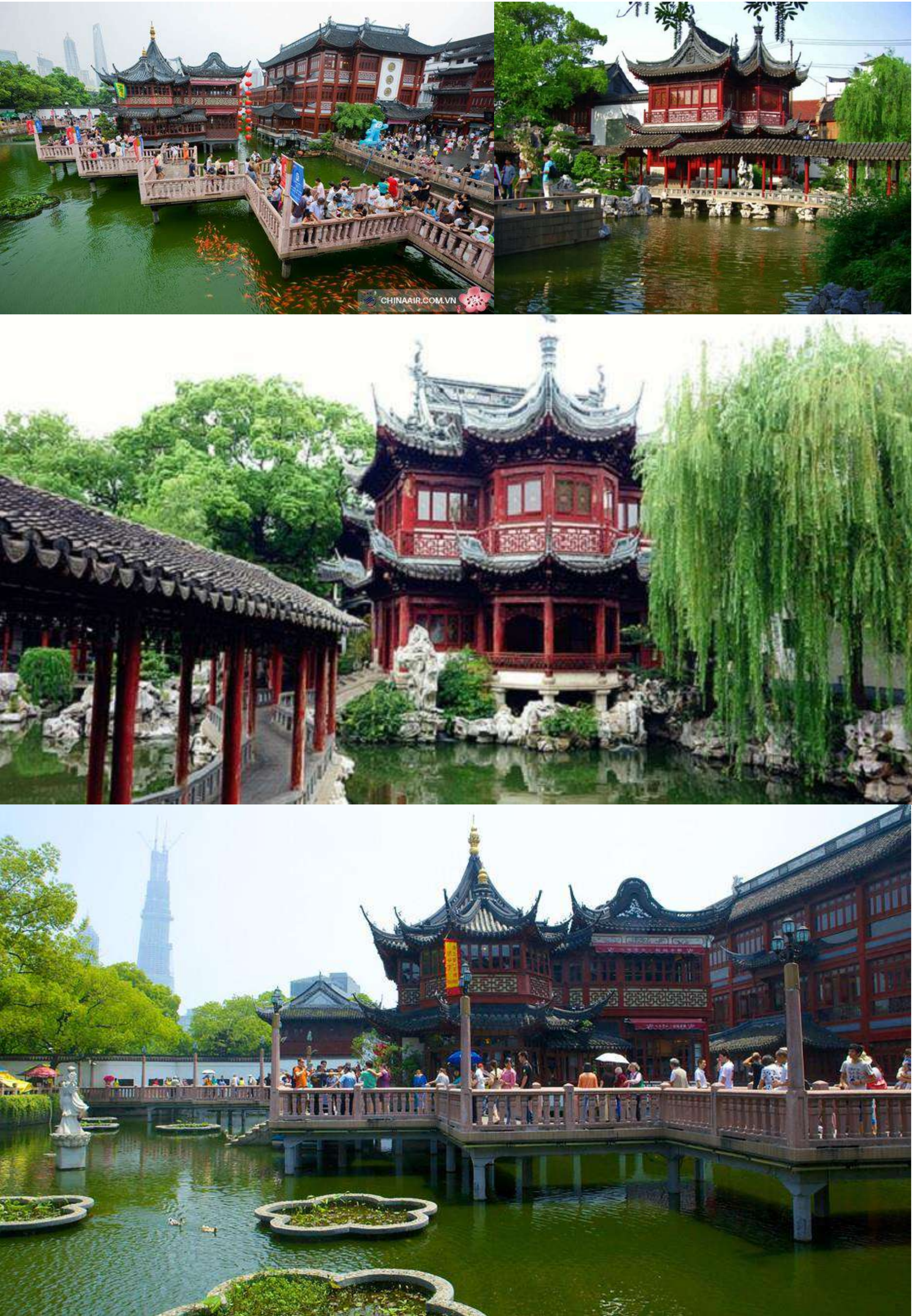} &      \includegraphics[height=0.1555\linewidth]{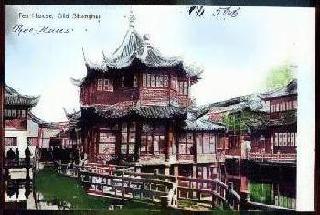}  
			& \hspace{0.02in} \includegraphics[height=0.1555\linewidth]{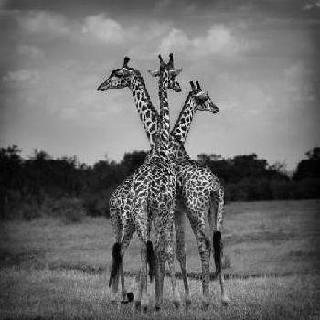}&   \includegraphics[height=0.1555\linewidth]{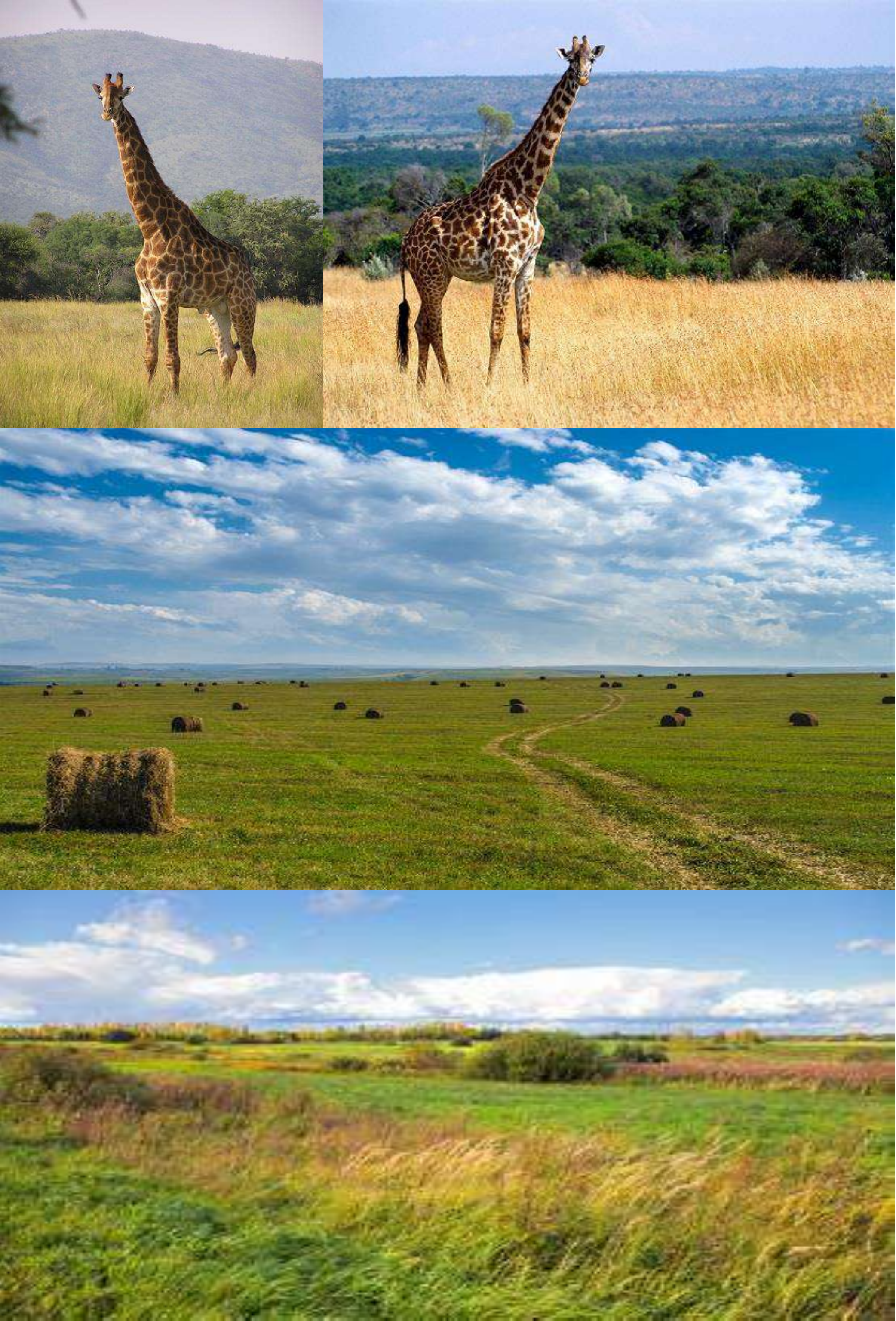} &
			\includegraphics[height=0.1555\linewidth]{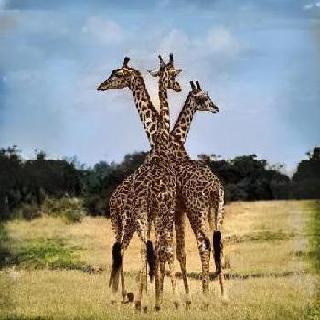}      \\
			Source & Reference & Result & \hspace{0.02in} Source & Reference & Result
		\end{tabular}   
		\caption{Colorization results. Input source images: M. Sternberg and Urszula Kozak.}  
		\label{fig:comp_set5}
	\end{figure*}

	We can also use our method for colorization of gray-scale images. We simply provide color references in the desired style in order to colorize our input gray-scale image.~\fref{fig:comp_set5} shows some colorization results.
	
	\section{Concluding Remarks}  
	In this paper, we demonstrate a new algorithm for locally transferring colors across semantically-related images. It not only handles single-reference transfer, but can also be adapted to multi-reference transfer, which avoids the often difficult task of finding a proper reference. We adapt a joint optimization of NNF search and local color transfer across CNN layers in a hierarchical manner. We have shown that our approach is widely applicable to a variety of transfer scenarios in real-world images.
	
\begin{figure}[t]
		\footnotesize
		\setlength{\tabcolsep}{0.002\linewidth}
		\begin{tabular}{ccc}      
			\includegraphics[width=0.31\linewidth,height=0.39\linewidth]{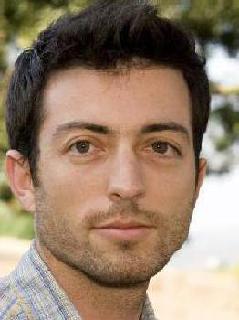}&      
			\includegraphics[width=0.31\linewidth,height=0.39\linewidth]{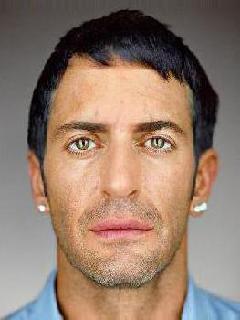}&     
			\includegraphics[width=0.31\linewidth,height=0.39\linewidth]{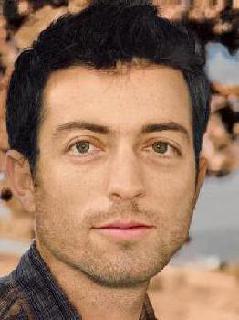} \\
			\includegraphics[width=0.31\linewidth]{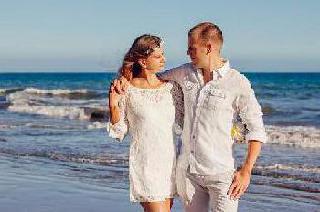}&     
			\includegraphics[height=0.26\linewidth]{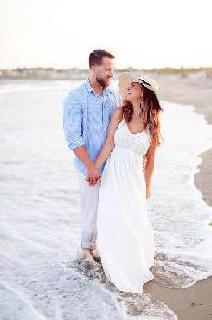}&    
			\includegraphics[width=0.31\linewidth]{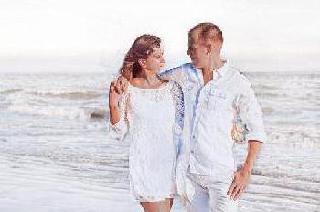}\\
			\includegraphics[width=0.31\linewidth]{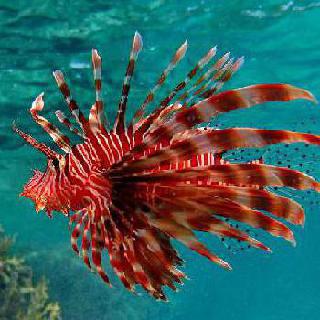}&     
			\includegraphics[width=0.31\linewidth]{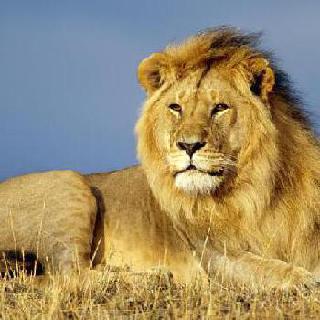}&   
			\includegraphics[width=0.31\linewidth]{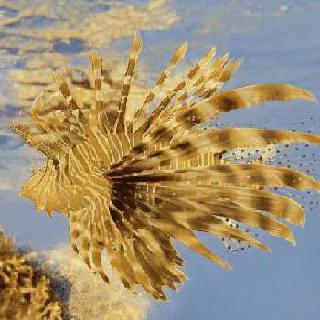}\\
			Source & Reference & Result\\
		\end{tabular} 
		\caption{Some examples of failure cases.} 
		\label{fig:limit}
	\end{figure}

	However our method still has limitations. It may mismatch some regions which exist in the source but not in the reference, and thus cause incorrect color transfer, such as the background on the left example of~\fref{fig:limit}. This often happens in the single-reference transfer and can be reduced by introducing more references. The VGG network we relied on \hmm{is not trained to} distinguish different instances with the same semantic labels, so it may cause color mixing between different instances, such as the man's shirt on the right example of~\fref{fig:limit}, which is transfered with mixed blue and white colors from two persons in the reference. A possible improvement would be to train a network on a dataset with instance labels. Moreover, directly applying our image color transfer method to video may cause some flickering artifacts ~\citep{chen2017coherent}. Addressing this would require temporal constraints, and will be considered for future work. \hmm{Our color transfer is suitable for semantically similar images. However, it may lead to some unnatural color effects for image pairs without a semantic relationship, such as the yellow water in the $3rd$ failure example. For future work, we would like to explore methods that would allow this example to automatically degenerate to a good color transfer result.}
	
	\bibliographystyle{ACM-Reference-Format}
	\bibliography{template}
	
\end{document}